\documentclass[conference]{IEEEtran}
\IEEEoverridecommandlockouts
\usepackage{cite}
\usepackage{amsmath,amssymb,amsfonts}

\usepackage{graphicx,booktabs,array}
\usepackage[normalem]{ulem}
\usepackage{textcomp}

\usepackage[usenames,dvipsnames]{xcolor}
\definecolor{commentsColor}{rgb}{0.497495, 0.497587, 0.497464}
\definecolor{keywordsColor}{rgb}{0.000000, 0.000000, 0.635294}
\definecolor{stringColor}{rgb}{0.558215, 0.000000, 0.135316}

\usepackage{physics}
\usepackage{float} 
\restylefloat{table}
\restylefloat{figure}

\usepackage{support-caption}
\usepackage{caption}
\makeatletter
\let\MYcaption\@makecaption
\makeatother
\usepackage[labelformat=simple,font=footnotesize]{subcaption}
\makeatletter
\let\@makecaption\MYcaption
\makeatother
\DeclareCaptionLabelSeparator{periodspace}{.\quad}
\usepackage{algorithm,algorithmicx,algcompatible}
\usepackage{placeins}
\usepackage{tikz}
\usepackage{adjustbox}
\usetikzlibrary{3d,calc}
\usepackage{tikz-3dplot}

\usepackage{listings}
\definecolor{codegreen}{rgb}{0,0.6,0}
\definecolor{codegray}{rgb}{0.5,0.5,0.5}
\definecolor{codepurple}{rgb}{0.58,0,0.82}
\definecolor{backcolour}{rgb}{0.95,0.95,0.92}

\def\BibTeX{{\rm B\kern-.05em{\sc i\kern-.025em b}\kern-.08em
    T\kern-.1667em\lower.7ex\hbox{E}\kern-.125emX}}

\graphicspath{{./Plots/}}
\newcommand{\plt}[1]{\includegraphics[width=0.25\textwidth]{#1}}

\makeatletter
\newcommand{\linebreakand}{%
  \end{@IEEEauthorhalign}
  \hfill\mbox{}\par
  \mbox{}\hfill\begin{@IEEEauthorhalign}
}
\makeatother

\begin{document}

\title{Generative Discrete Event Process Simulation for Hidden Markov Models to Predict Competitor Time-to-Market\\
}

\newcommand{\rt}[1]{{\color{blue}[(Reuben) #1}]}
\newcommand{\rte}[1]{{\color{blue} #1}}
\newcommand{\cut}{\text{cut}}

\definecolor{kellygreen}{rgb}{0.3, 0.73, 0.09}

\newcommand{\se}[1]{{\color{kellygreen}[(Stephan) #1}]}
\newcommand{\see}[1]{{\color{kellygreen} #1}}

\newcommand{\sfe}[1]{{\color{red}[(Sean) #1}]}
\newcommand{\sfee}[1]{{\color{red} #1}}

\author{
\IEEEauthorblockN{Nandakishore Santhi}
\IEEEauthorblockA{
\textit{Los Alamos National Laboratory}\\
Los Alamos, NM, 87545 USA \\
nsanthi@lanl.gov}

\and

\IEEEauthorblockN{Stephan Eidenbenz}
\IEEEauthorblockA{
\textit{Los Alamos National Laboratory}\\
Los Alamos, NM, 87545 USA \\
eidenben@lanl.gov}

\linebreakand

\IEEEauthorblockN{Brian Key}
\IEEEauthorblockA{
\textit{Los Alamos National Laboratory}\\
Los Alamos, NM, 87545 USA \\
bkey@lanl.gov}

\and

\IEEEauthorblockN{George Tompkins}
\IEEEauthorblockA{
\textit{Los Alamos National Laboratory}\\
Los Alamos, NM, 87545 USA \\
tompkins@lanl.gov}
}

\maketitle

\begin{abstract}
We study the challenge of predicting the time at which a competitor product, such as a novel high-capacity EV battery or a new car model, will be available to customers; as new information is obtained, this time-to-market estimate is revised. Our scenario is as follows: We assume that the product is under development at a Firm B, which is a competitor to Firm A; as they are in the same industry, Firm A has a relatively good understanding of the processes and steps required to produce the product. While Firm B tries to keep its activities hidden (think of stealth-mode for start-ups), Firm A is nevertheless able to gain periodic insights by observing what type of resources Firm B is using, such as from regulatory filing requirements, simple observation of usage of parking lots, insights from recruiting trends etc. We show how Firm A can build a model that predicts when Firm B will be ready to sell its product; the model leverages knowledge of the underlying processes and required resources to build a Parallel Discrete Simulation (PDES)-based process model that it then uses as a generative model to train a Hidden Markov Model (HMM) 

We study the question of how many resource observations Firm A requires in order to accurately assess the current state of development at Firm B. In order to gain general insights into the capabilities of this approach, we study the effect of different process graph densities, different densities of the resource-activity maps, etc., and also scaling properties as we increase the number resource counts. 

We find that in most cases, the HMM achieves a prediction accuracy of 70 to 80 percent after 20 (daily) observations of a production process that lasts 150 days on average and we characterize the effects of different problem instance densities on this prediction accuracy. Our results give insight into the level of market knowledge required for accurate and early time-to-market prediction.

\end{abstract}


\section{Introduction}
\label{sec:intro{}}
Consider the following scenario:
A product is under development at a Firm B, which is a competitor to Firm A; as they are in the same industry, Firm A has a relatively good understanding of the processes and steps required to produce the product, including likely footprint of plants to be built/re-purposed, amount of labor and skills, raw materials etc required at different stages of the production process. While Firm B tries to keep its activities hidden (think of stealth-mode for start-ups), Firm A is nevertheless able to gain periodic insights by observing what type of resources Firm B is using, such as from regulatory filing requirements, simple observation of usage of parking lots, legally obtained insights from poached employees etc. Firm A aims to build a model that predicts when Firm B will be ready to sell its product.

\begin{figure*}[t]
    \centering
 \includegraphics[width=\textwidth]{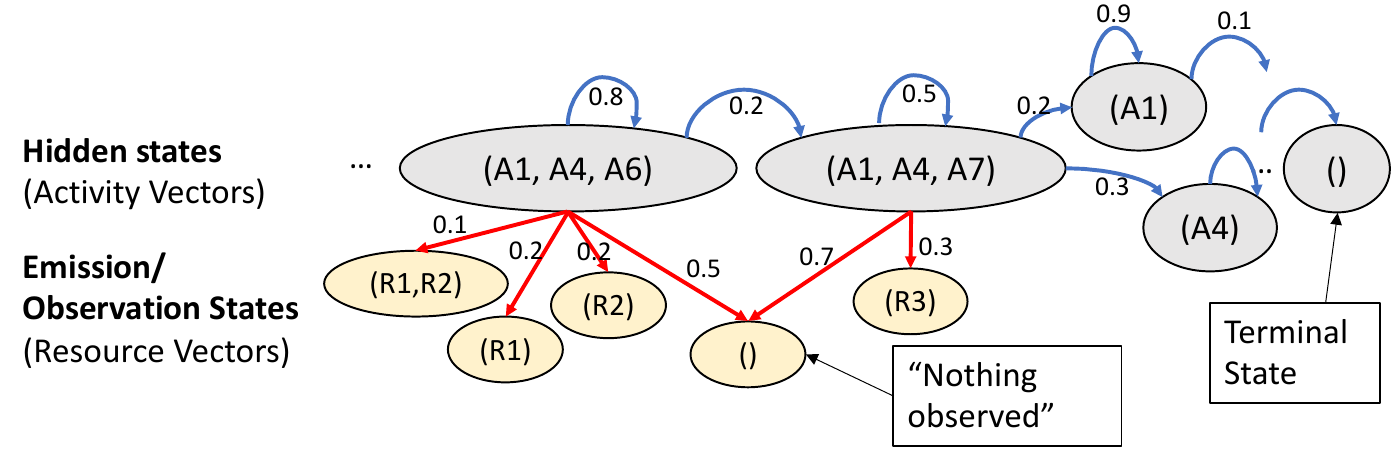}
    \caption{Illustrative example of the Hidden HMM model with Activity Sets as Hidden States and Resource Sets as Observation or Emission states
    }
    \label{fig:hmmexample}
\end{figure*}

We propose a model that assumes (i) knowledge of the required directed process graph (where we call individual vertices activities; each activity can only be started once its parent activities have been completed), and (ii) knowledge of what resources each activity requires, e.g. a bipartite resource-activity mapping. As Firm B works through its production process with a set of current activities at any given point in time, Firm A is able to periodically (let's assume daily for ease of discussion) observe the resources being used at that time with some uncertainty, e.g, with some probability a resource is being used, but Firm A is unable to observe it. 

Our approach has four phases: 
\begin{enumerate}
\item 
We build a discrete-event simulation-based process model, implemented in the Simian framework \cite{simian} that encodes our knowledge of the directed process graph and the resource-activity map. 
\item 
This process model is then used in generative fashion to train a Hidden Markov Model (HMM), where a hidden state corresponds to set of currently active activities and the observed state is a subset of the resources required by those activities. 
\item 
The HMM model is then given an actual sequence of real-life observations (say ending with the observation of resources from today), from which it predicts the most likely sequence of hidden states, and thus the activities that have led to this observation sequence. 
\item 
We return to the process model and start it from the state of active activities (in the process graph) to see how many more days the process will take to complete, thus predicting the time to market of the new product. 
\end{enumerate}

We study the question of how many days of observation Firm A requires in order to accurately assess the current state of development at Firm B and thus predict time to market. 

While such a model will by its very nature be very specific to the product under development, we perform an extensive parameter study to gain more general insights. In particular, we study the effect of different process graph densities, different densities of the resource-activity maps, noise-levels on the probability of observation, etc., and also scaling properties as we increase number of activities and resource counts. 

Our main findings are the following: We find that in most cases, the HMM achieves a prediction accuracy of 70 to 80 percent after 20 (daily) observations of a production process that lasts 150 days on average. The prediction accuracy increases with higher resource counts and with higher process graph densities; interestingly, for low density process graphs, the prediction accuracy decreases with increasing resource-activity map density, wheras -- for higher-density process graphs -- it achieves a peak for medium density resource-activity maps before falling off at very high density. Our results give insight into the level of market knowledge required for accurate and early time-to-market prediction.

In terms of related work, the concept of using a PDES-based process model in a generative fashion to feed an HMM model has been theoretically investigated in the context of tensor factorization \cite{skau2024generating}; our work uses some of the same underlying methods, but our main focus is on a broad numerical study across a large set of possible process models. We also refer to \cite{skau2024generating} for a survey on existing techniques in process modeling and HMM modeling.

As for the structure of this report, Section \ref{sec:procmodel} contains the detailed description of our process model, Section \ref{sec:hmm} describes our Hidden Markov Model approach with Section \ref{sec:simulation} detailing our simulation study before concluding with Section \ref{sec:conclusion}.

\begin{figure*}[t]\sffamily
\begin{tabular*}{\textwidth}{c@{}c@{}c@{}c@{}}
\toprule
    \plt{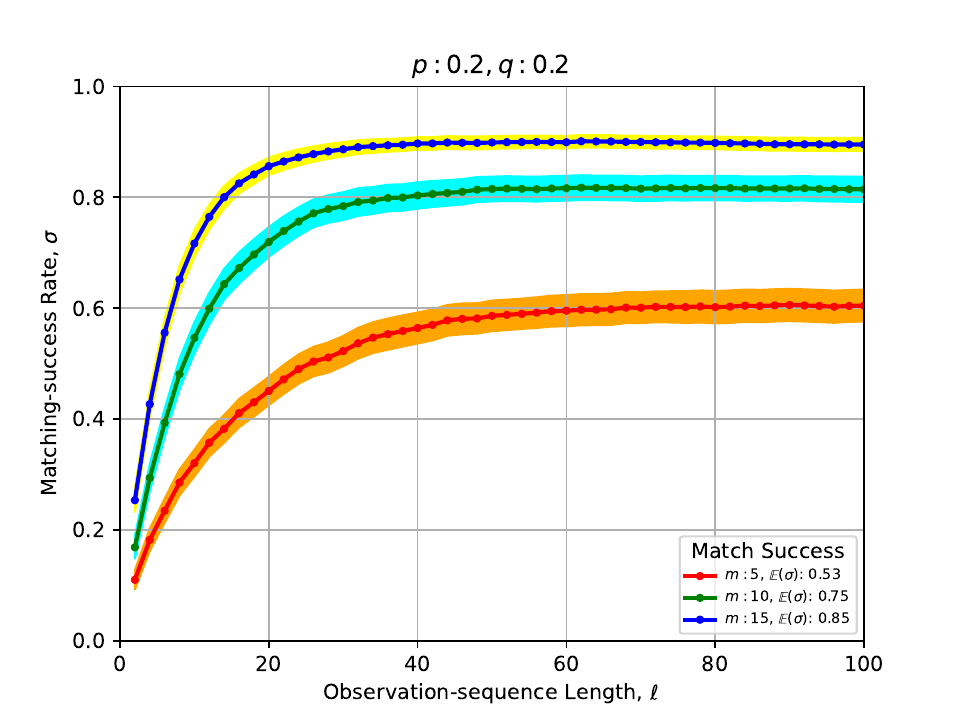} & \plt{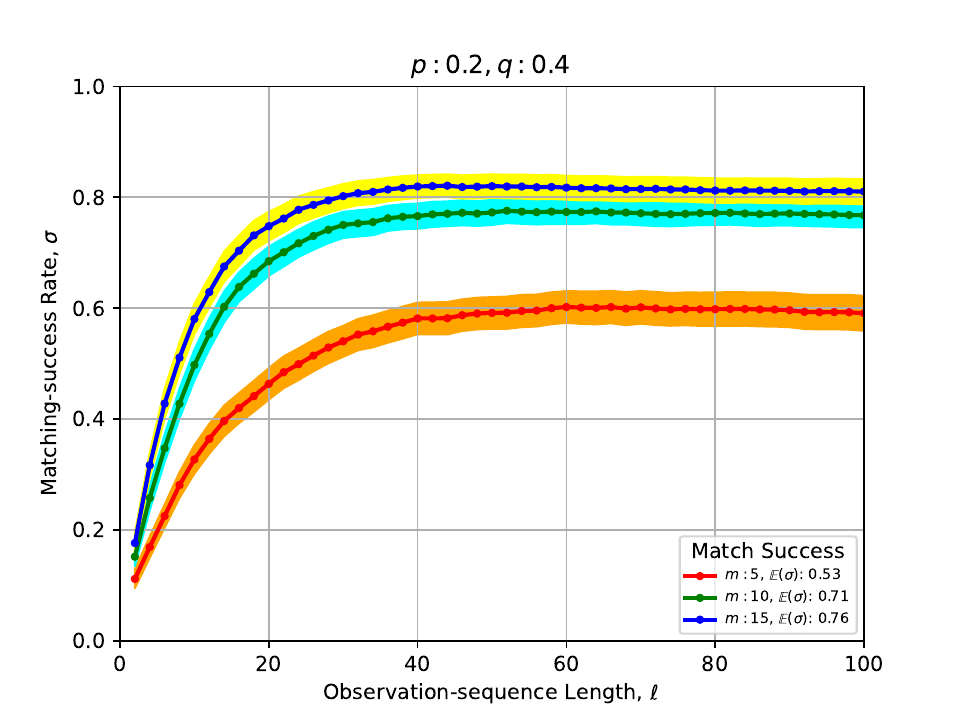} &
    \plt{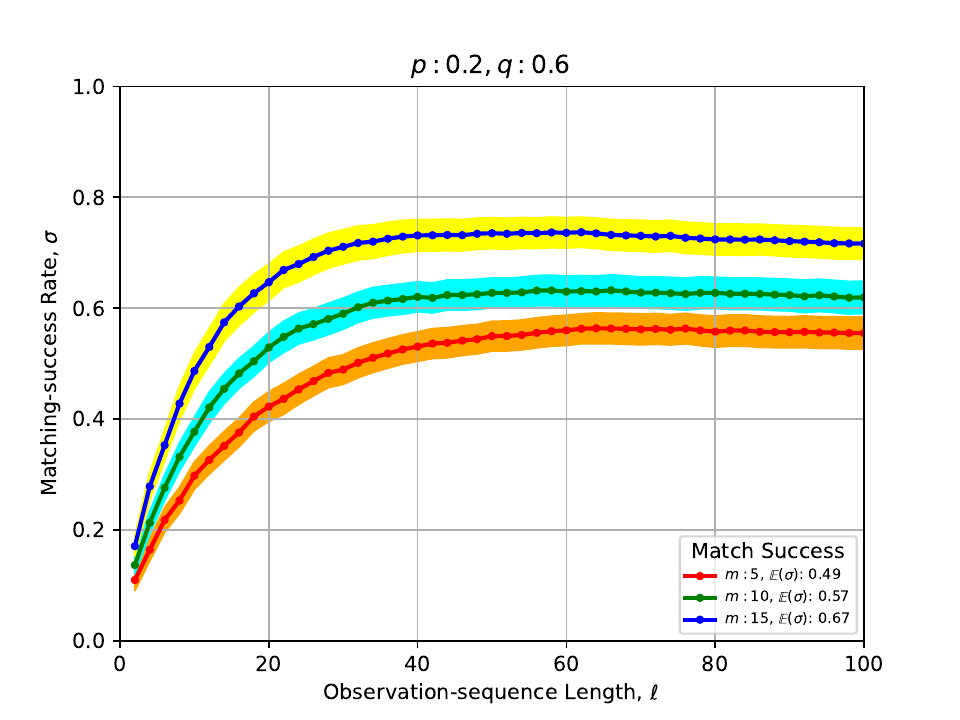} &
    \plt{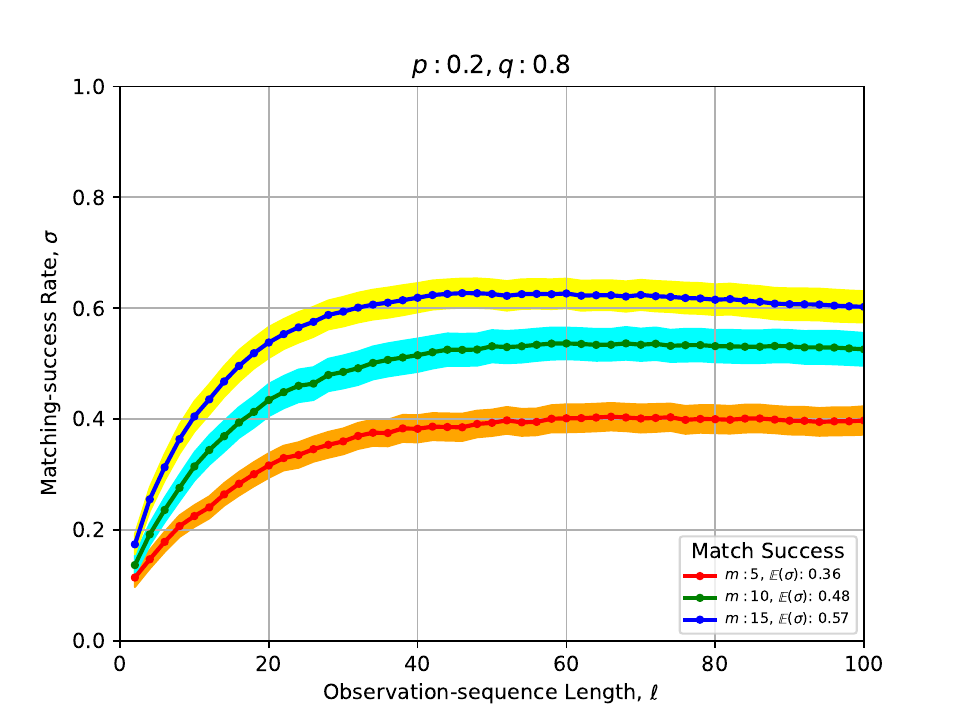} \\
    \plt{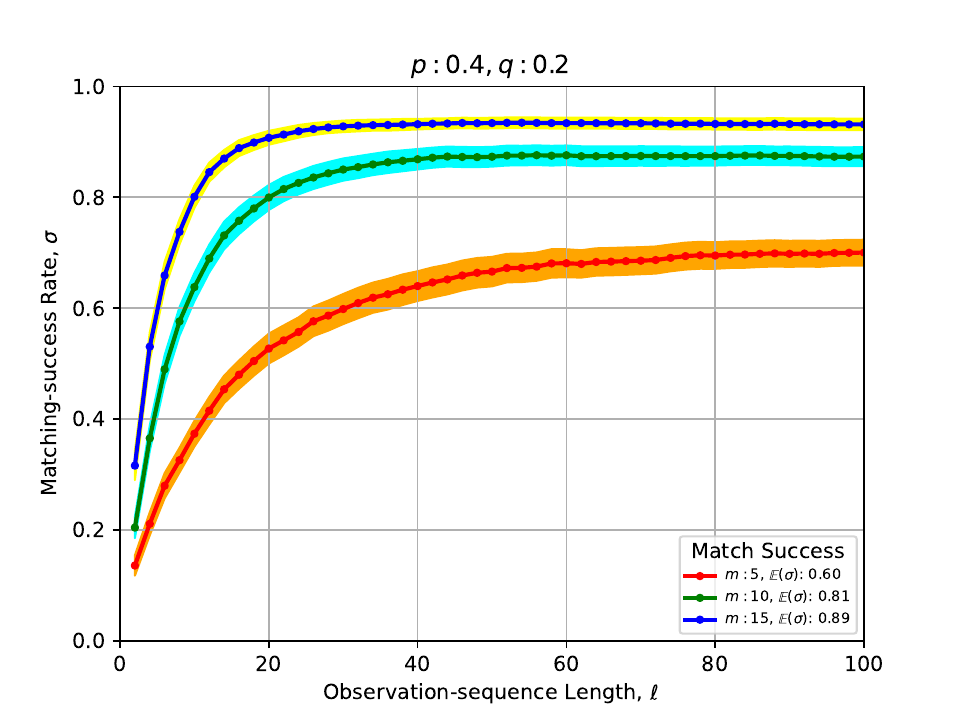} & \plt{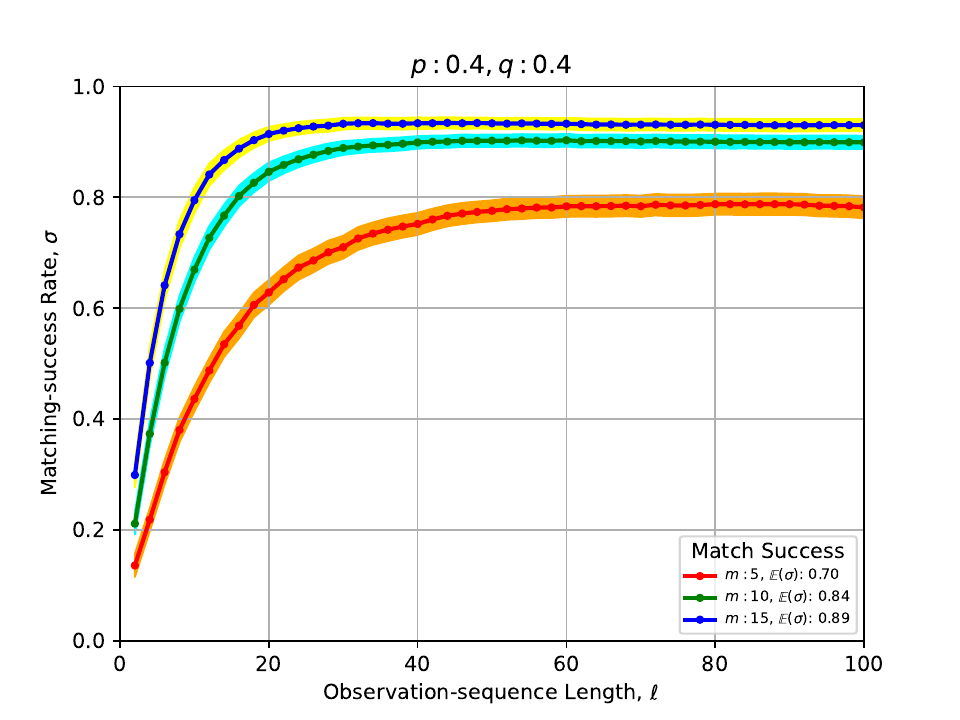} &
    \plt{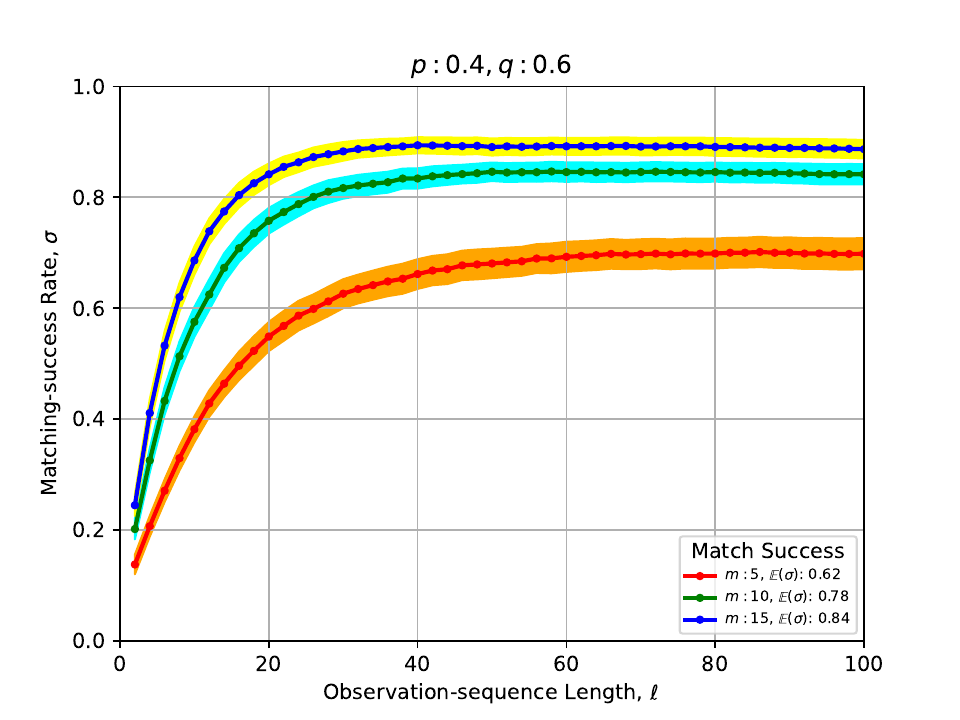} &
    \plt{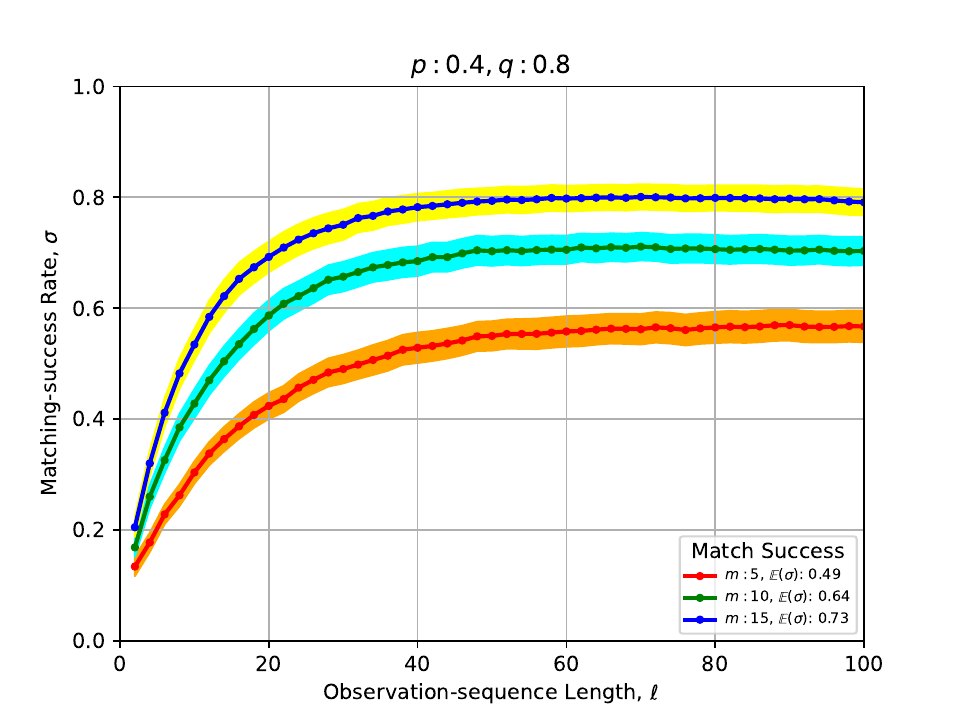} \\
    \plt{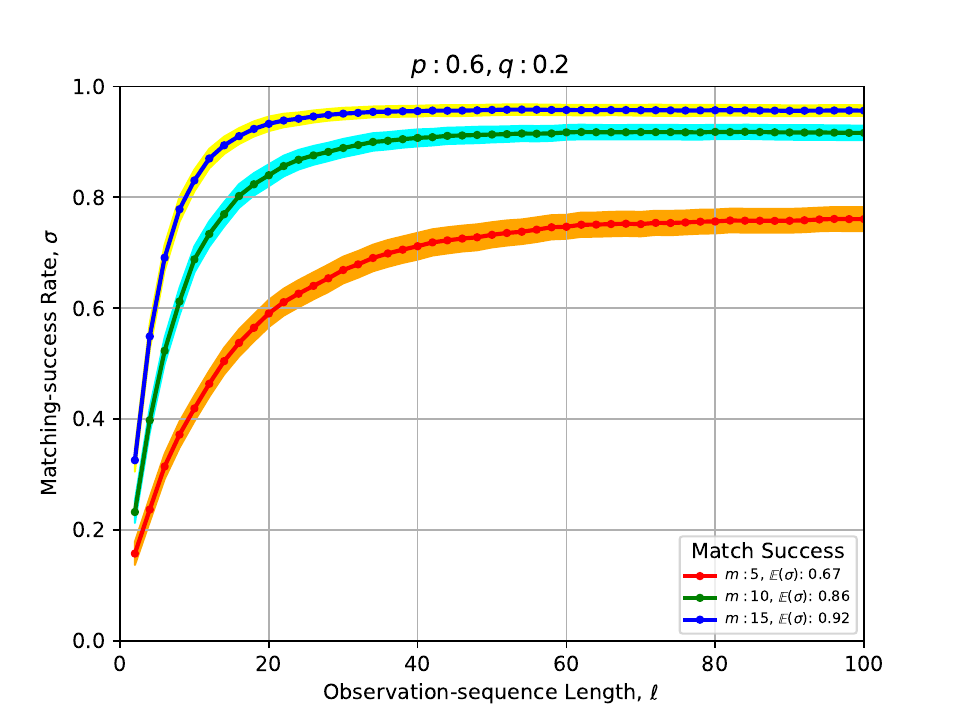} & \plt{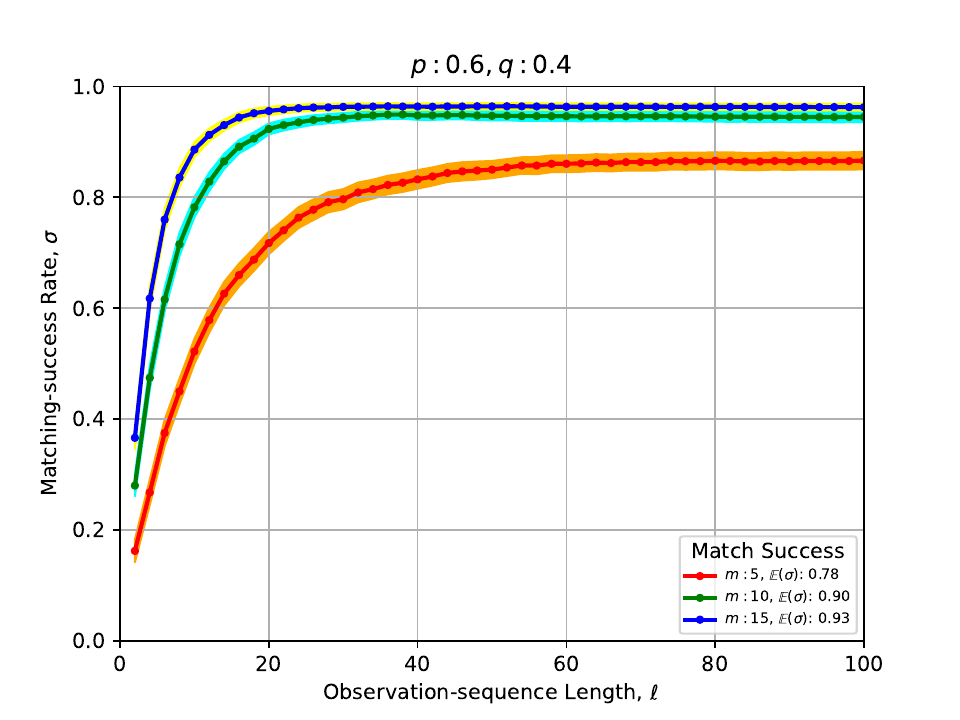} &
    \plt{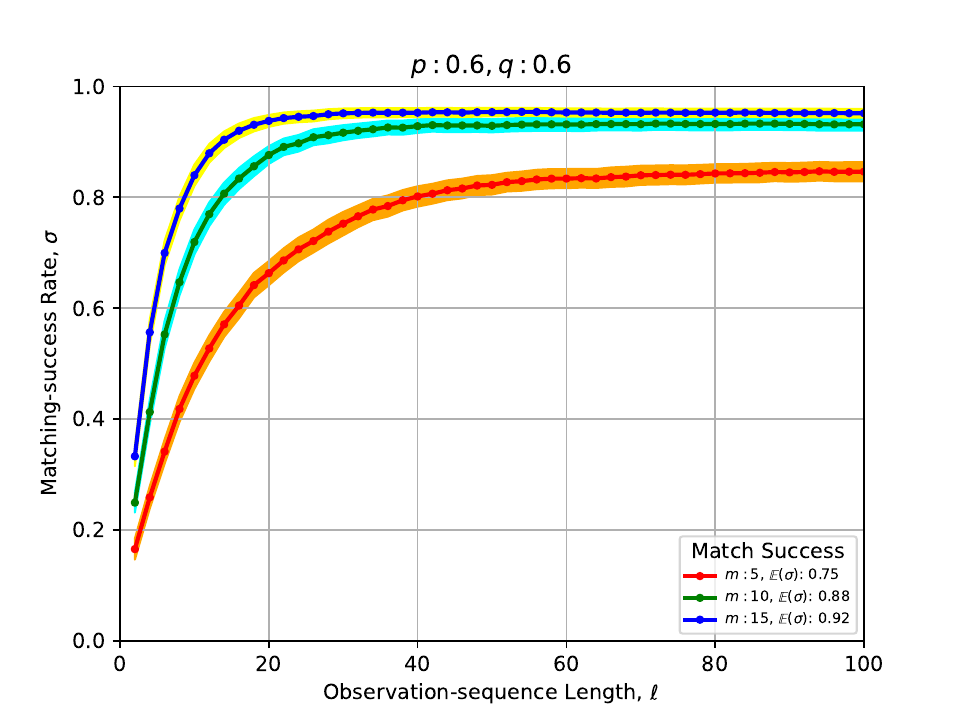} &
    \plt{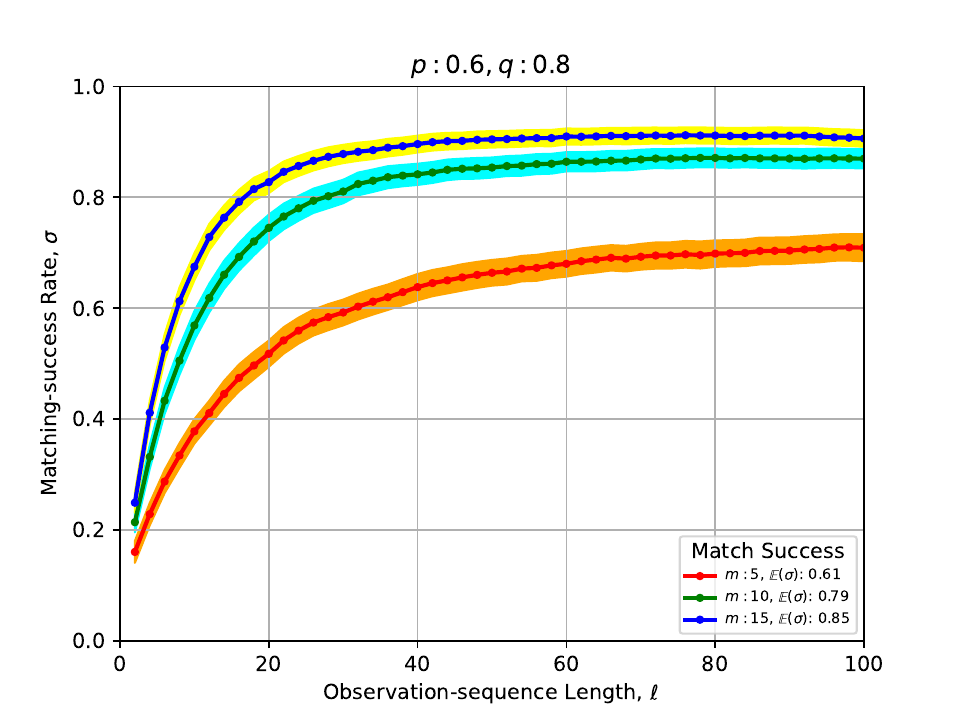} \\
    \plt{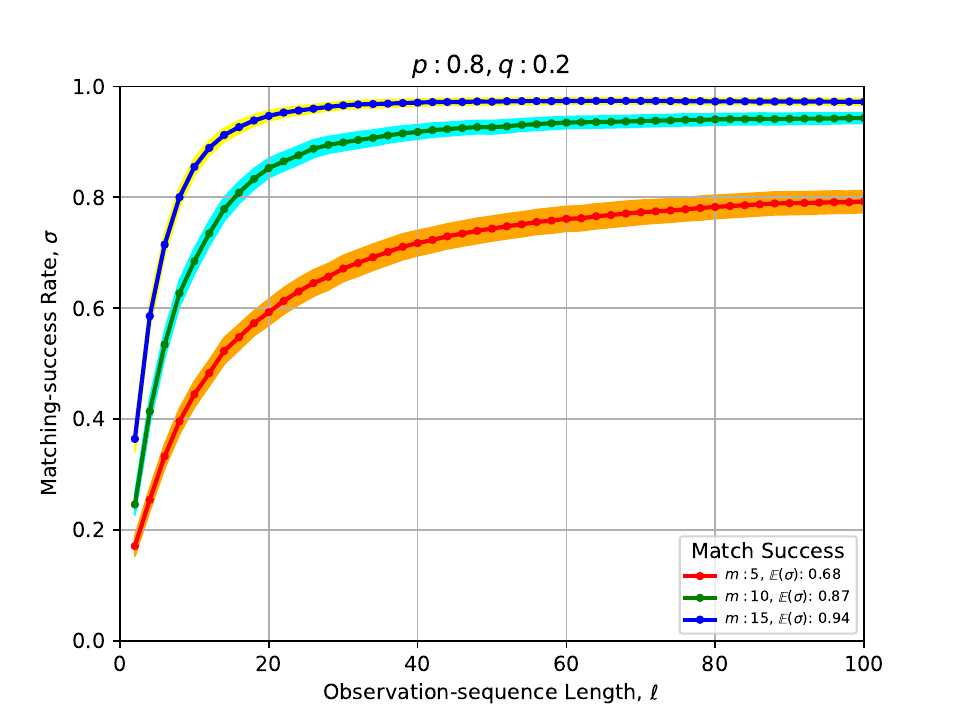} & \plt{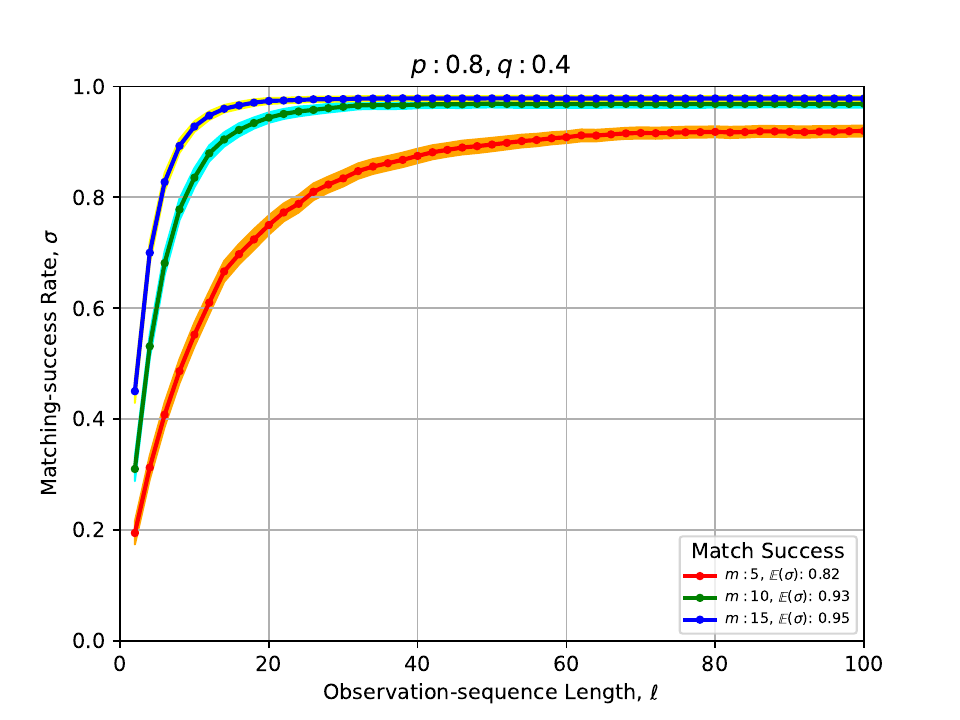} &
    \plt{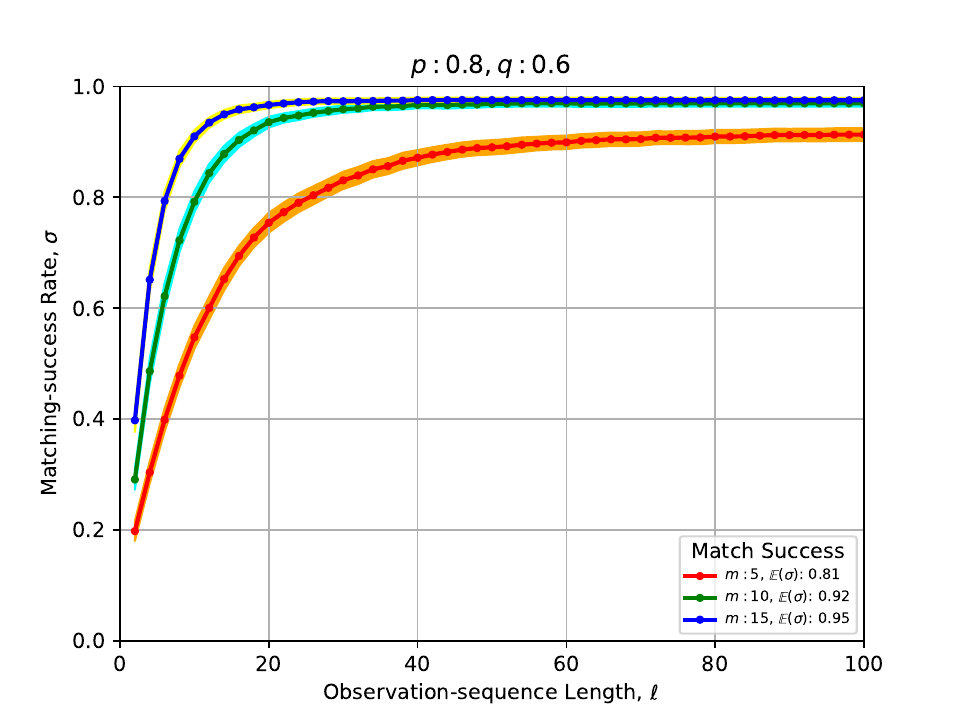} &
    \plt{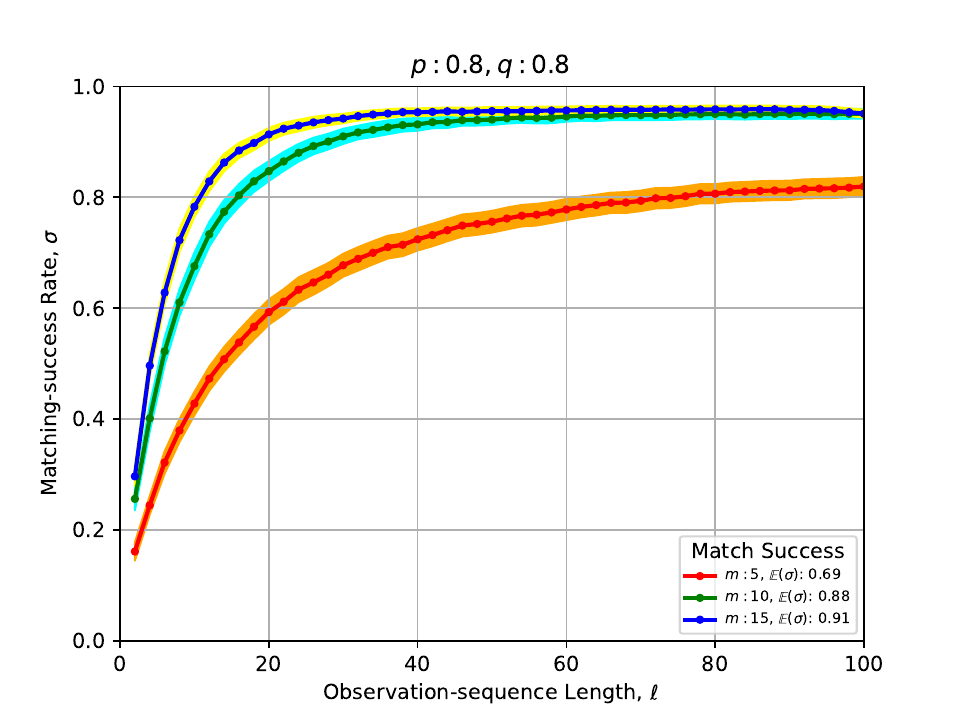} \\
\bottomrule 
\end{tabular*}
\caption{The figures show how matching-success rates change with observation sequence lengths when activity-graph density $p$ and resource-map graph density $q$ are varied across each figure in the array. Each figure shows several individual plots for different settings of number of resources $m$. Also shown are the $95\%$ confidence-bands around the sample means of the matching-success rates.}
\label{RI_1}
\end{figure*}

\begin{figure*}[t]\sffamily
\begin{tabular*}{\textwidth}{c@{}c@{}c@{}c@{}}
\toprule
    \plt{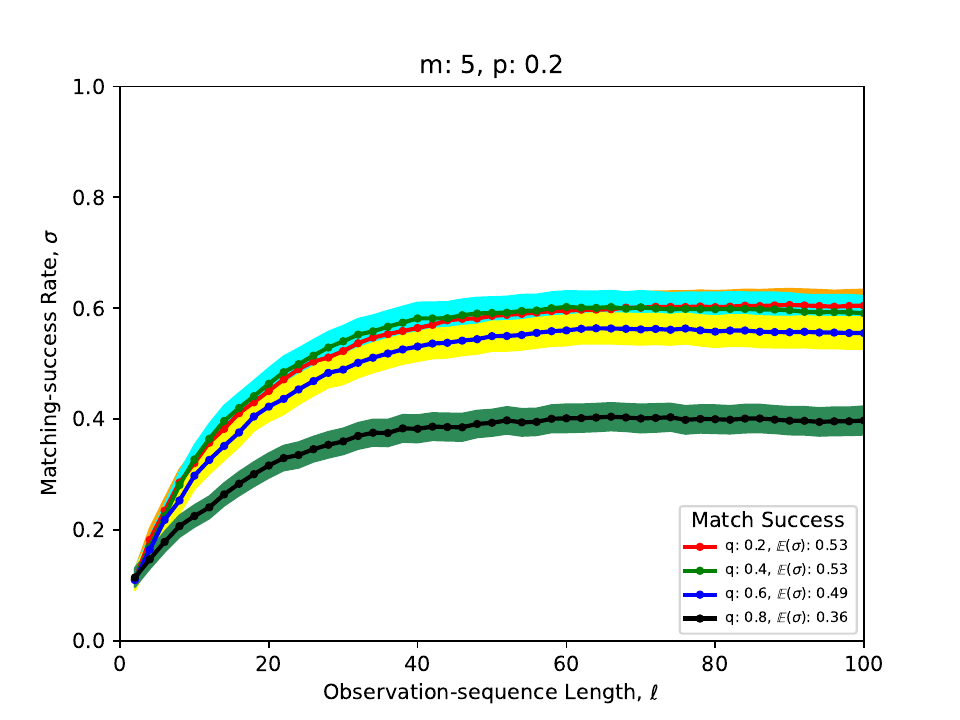} & \plt{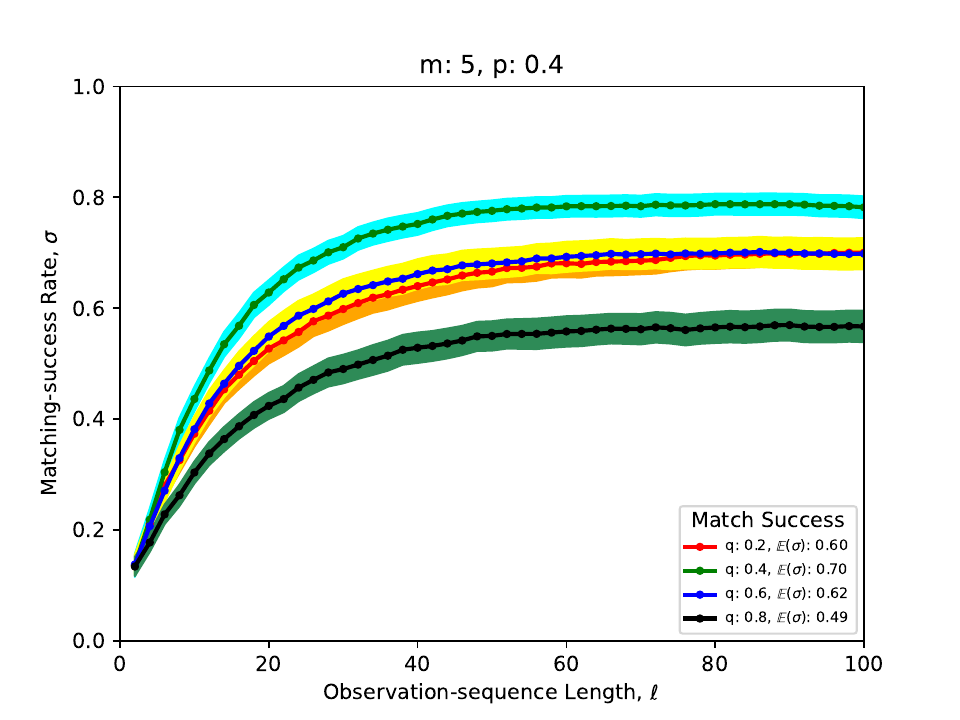} &
    \plt{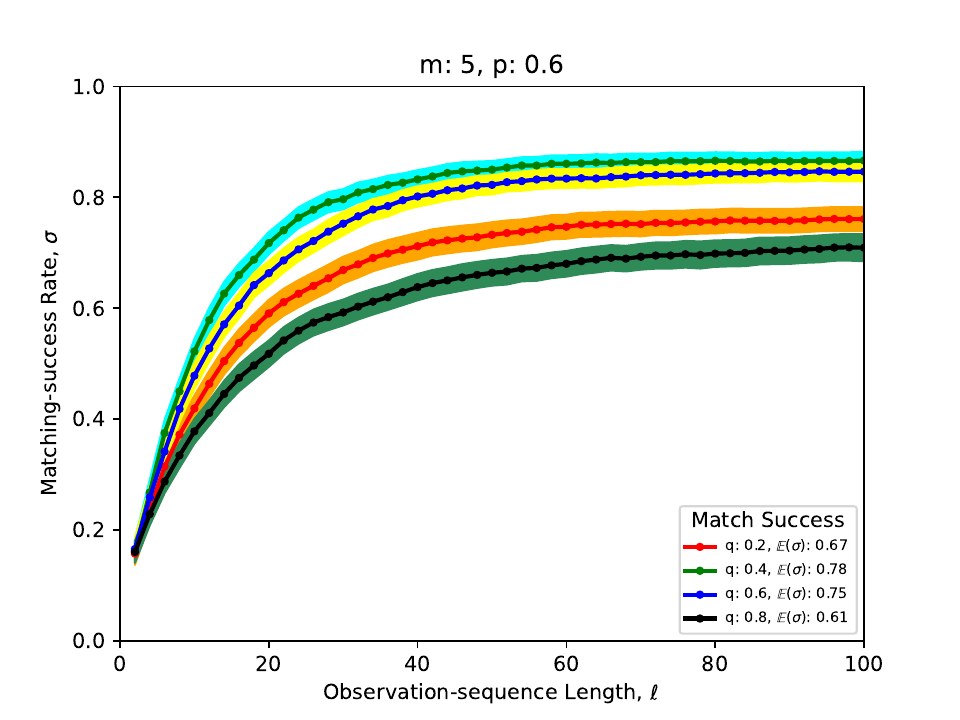} &
    \plt{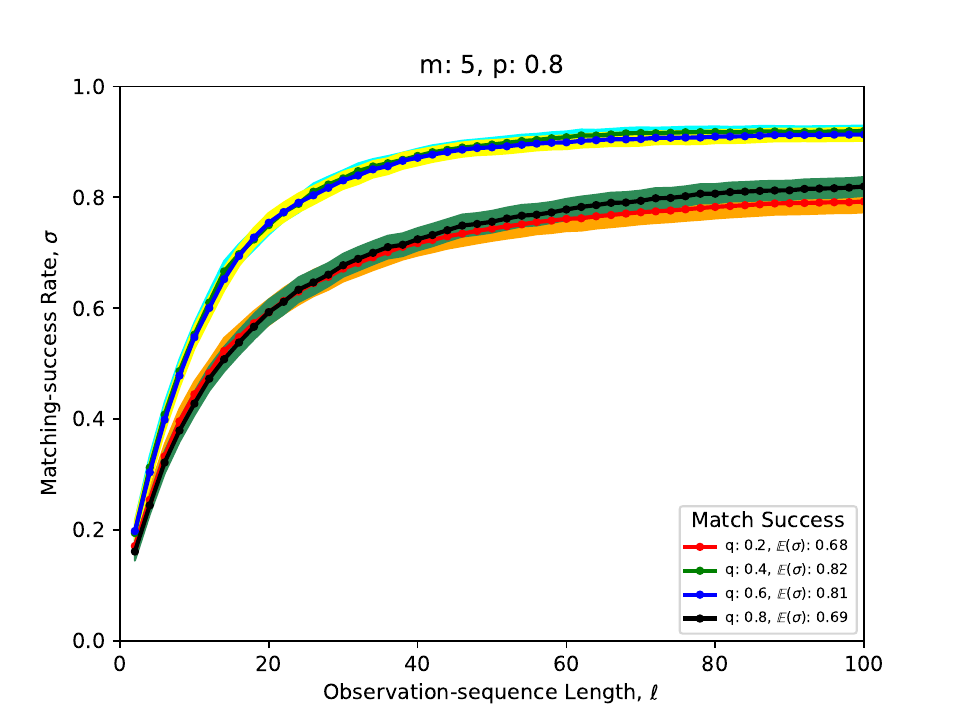} \\
    \plt{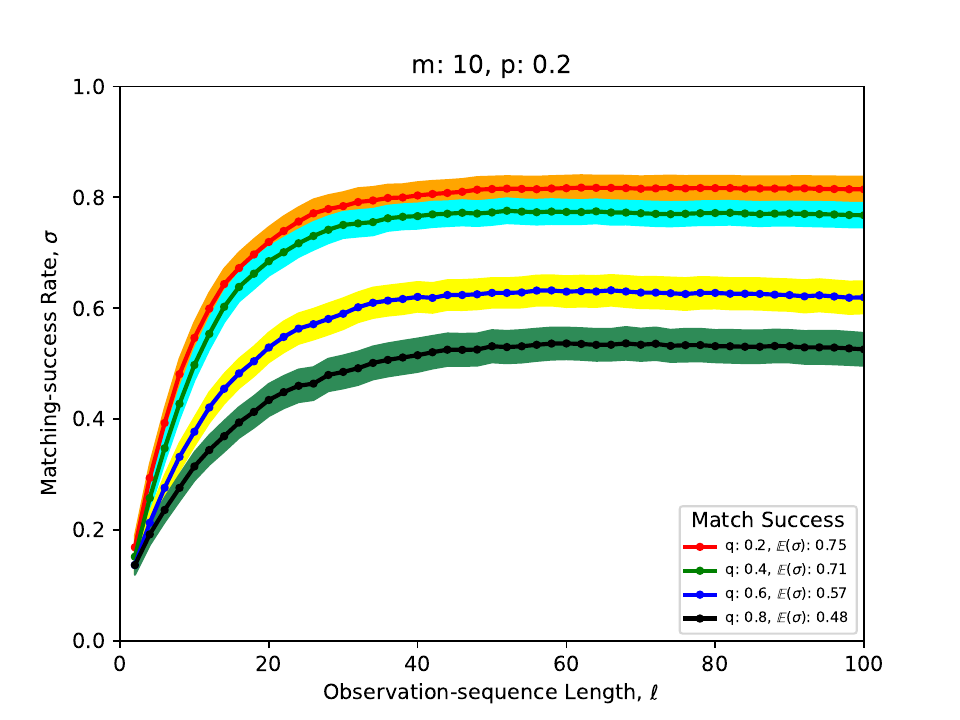} & \plt{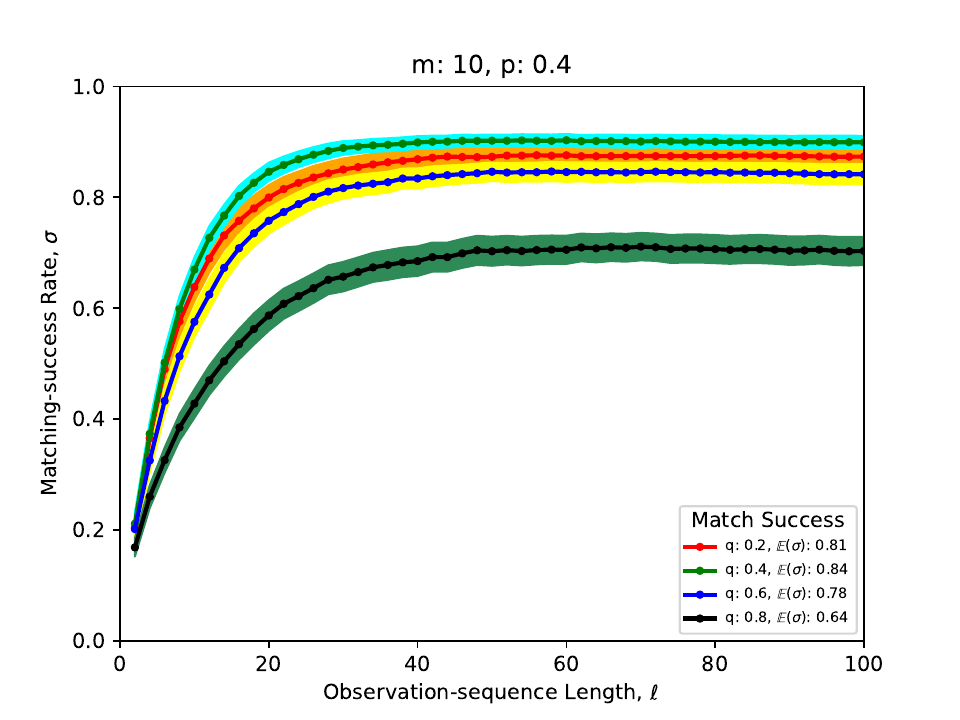} &
    \plt{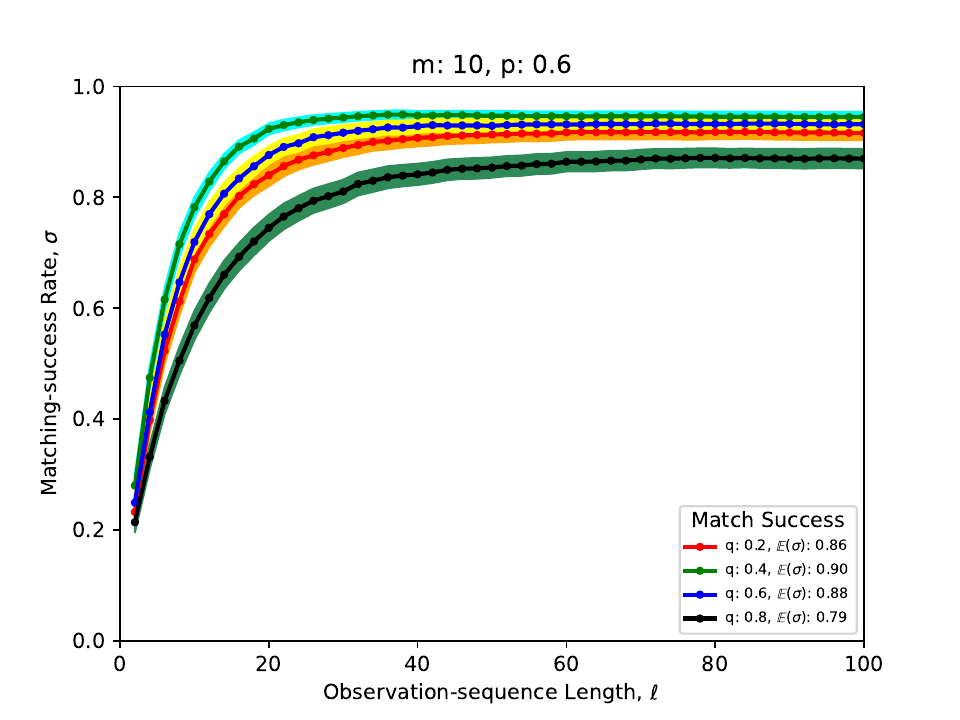} &
    \plt{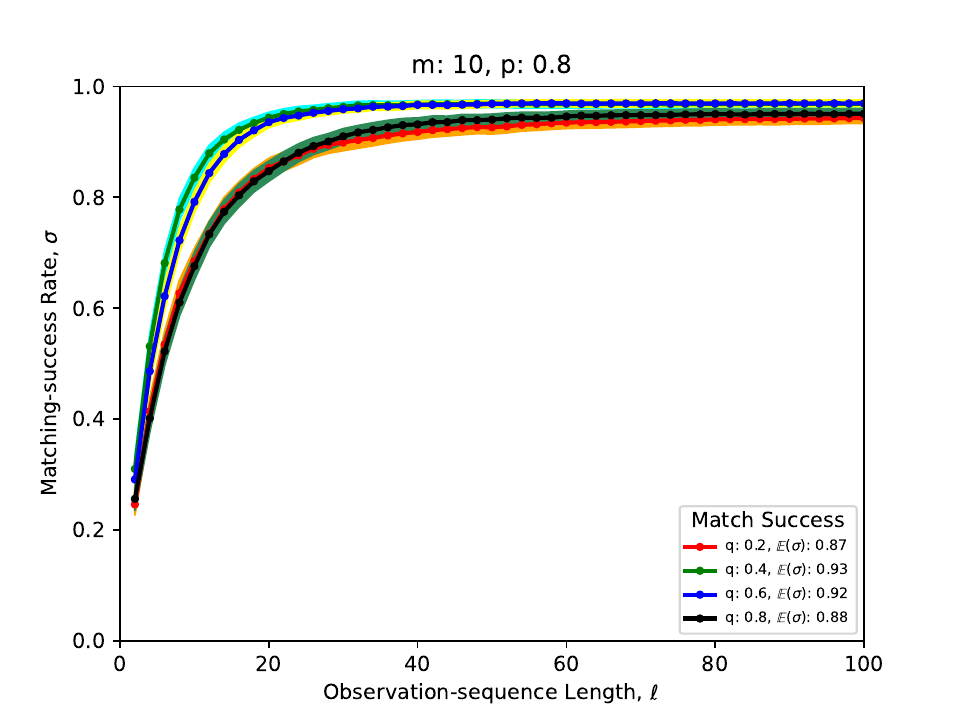} \\
    \plt{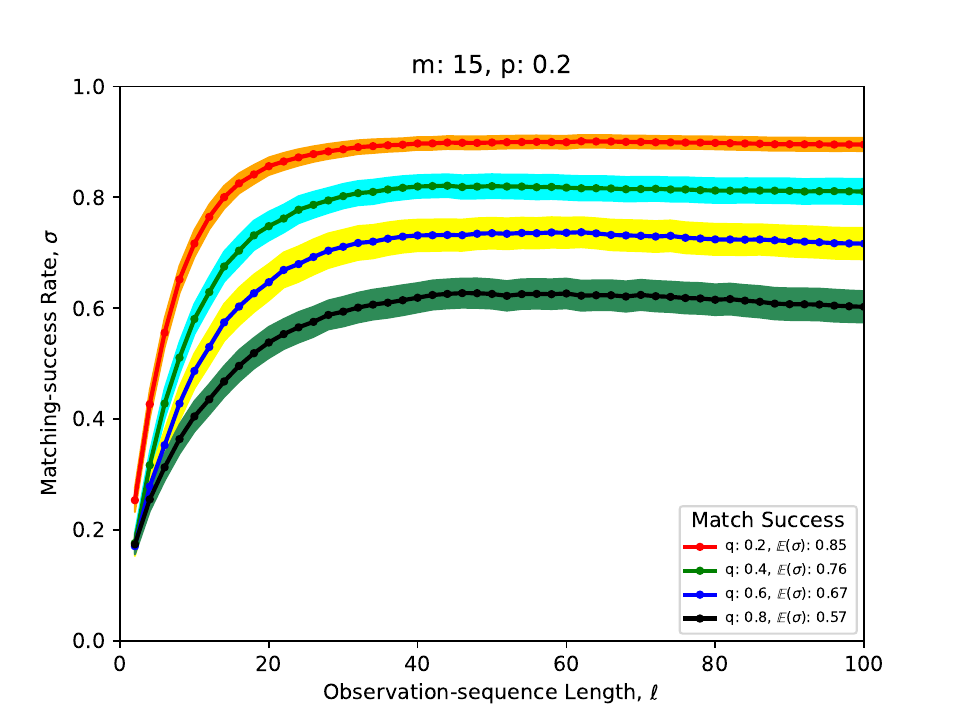} & \plt{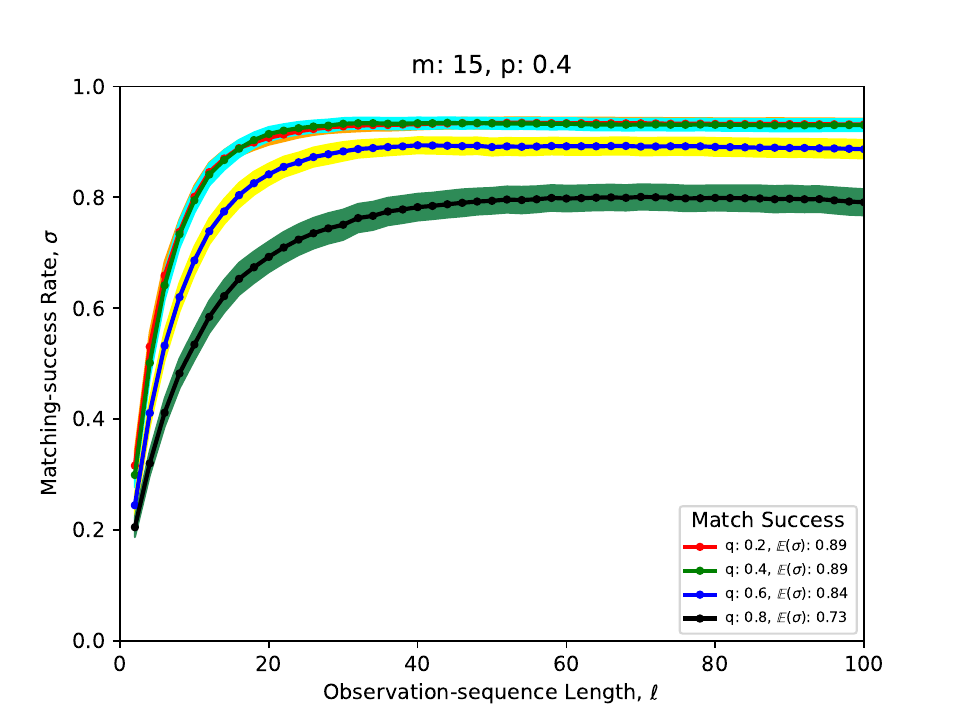} &
    \plt{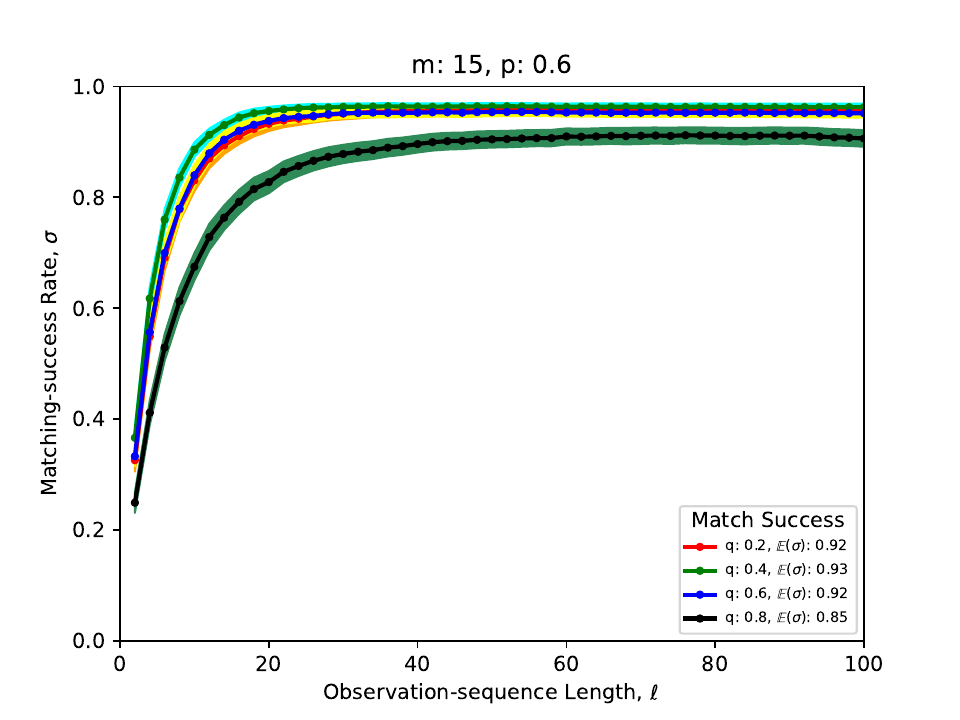} &
    \plt{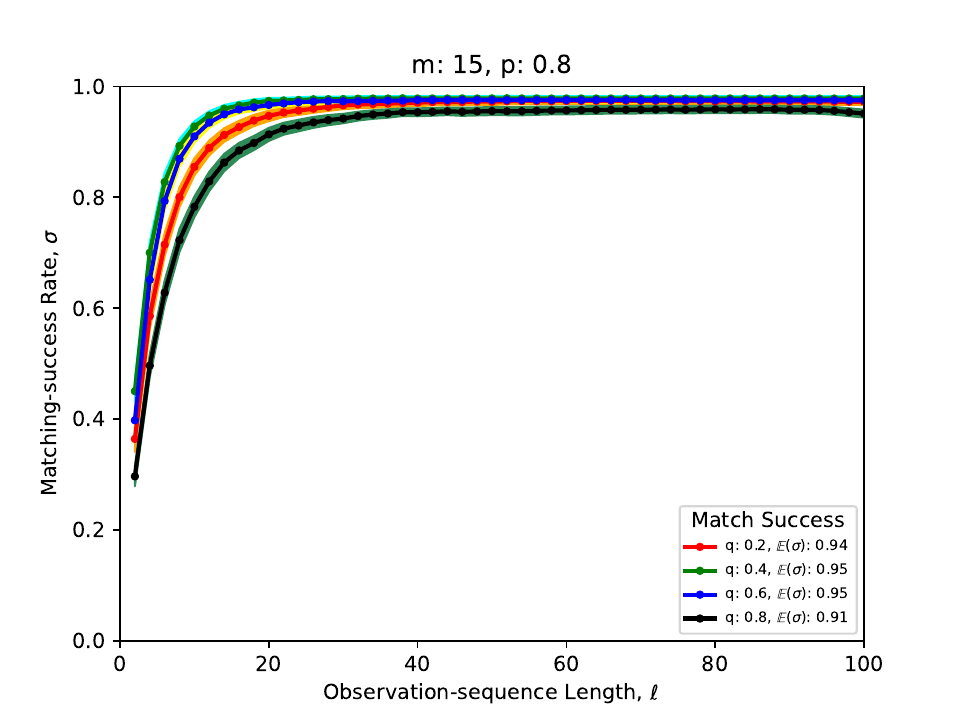} \\
\bottomrule 
\end{tabular*}
\caption{The figures show how matching-success rates change with observation sequence lengths when the number of resources $m$ and activity-graph density $p$ are varied across each figure in the array. Each figure shows several individual plots for different settings of resource-map graph densities $q$. Also shown are the $95\%$ confidence-bands around the sample means of the matching-success rates.}
\label{RI_2}
\end{figure*}

\begin{figure*}[t]\sffamily
\begin{tabular*}{\textwidth}{c@{}c@{}c@{}c@{}}
\toprule
    \plt{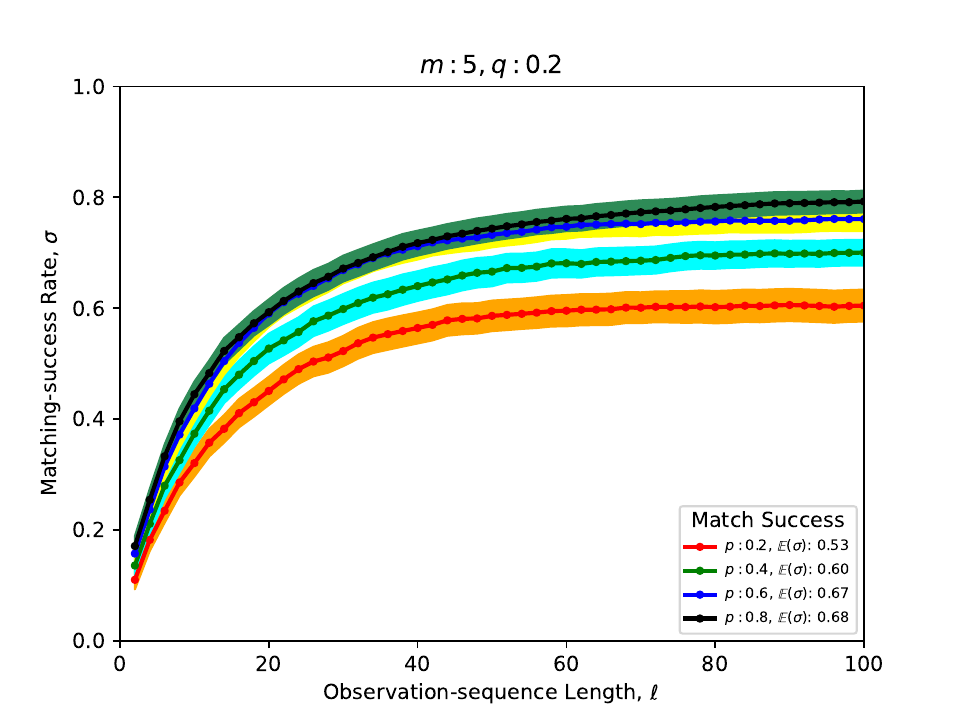} & \plt{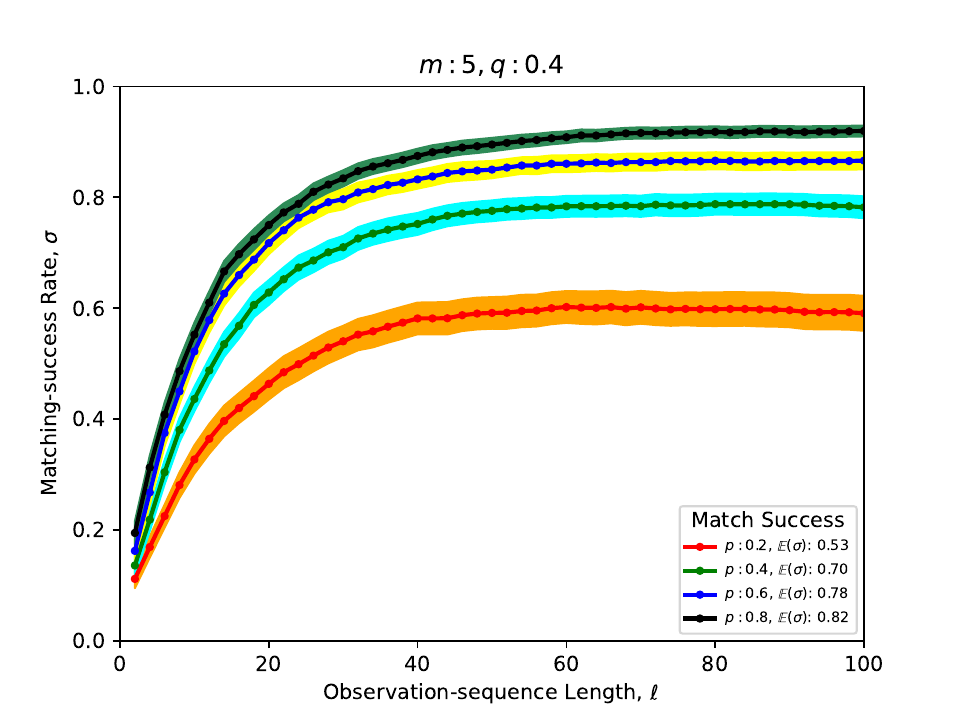} &
    \plt{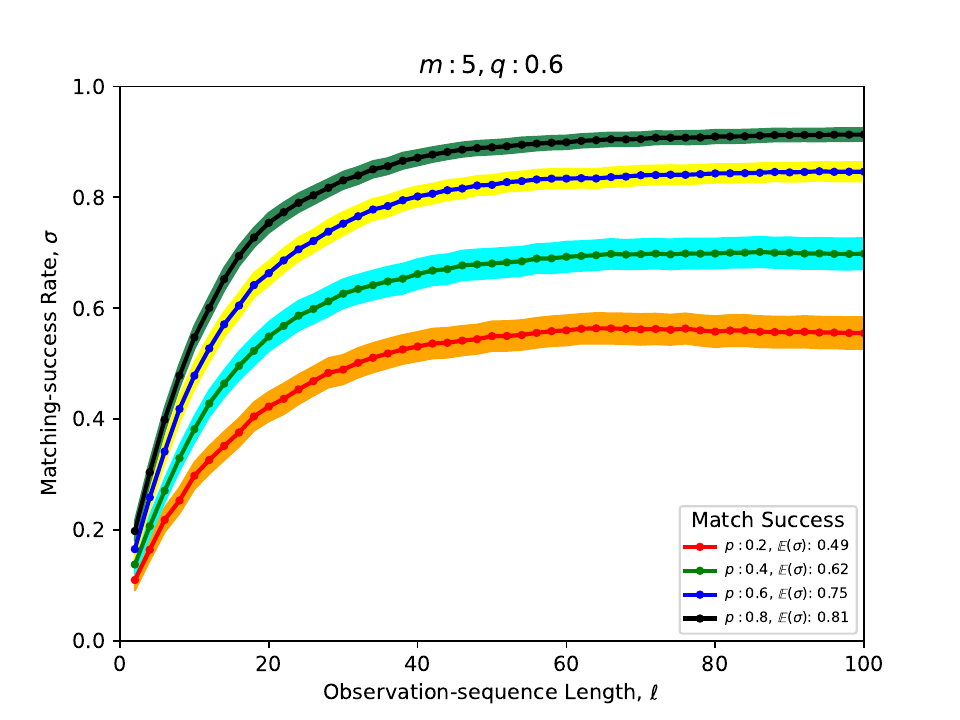} &
    \plt{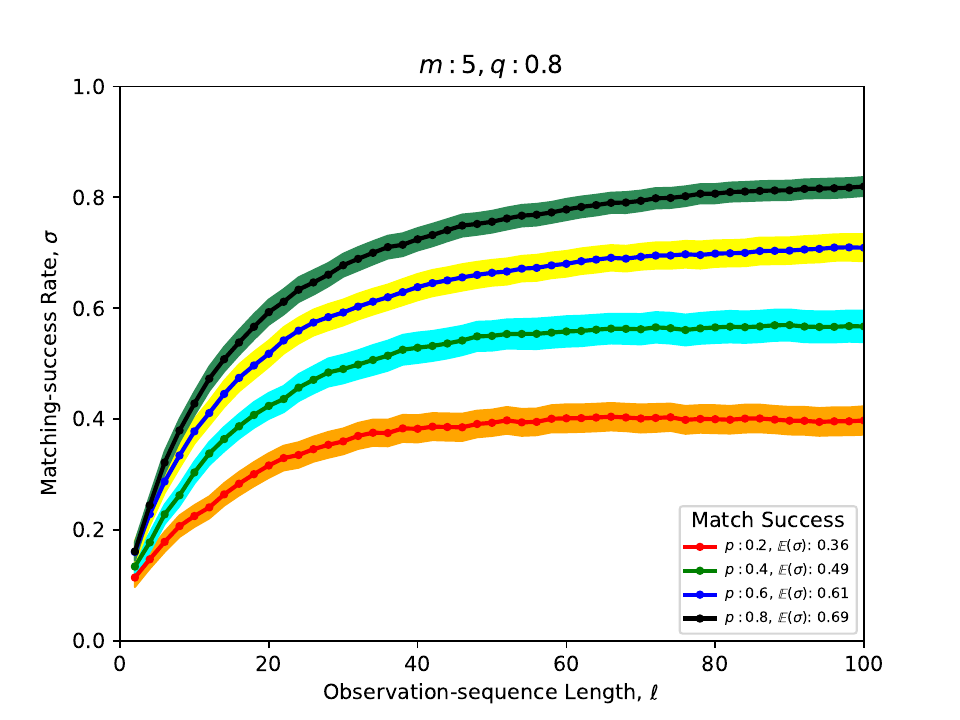} \\
    \plt{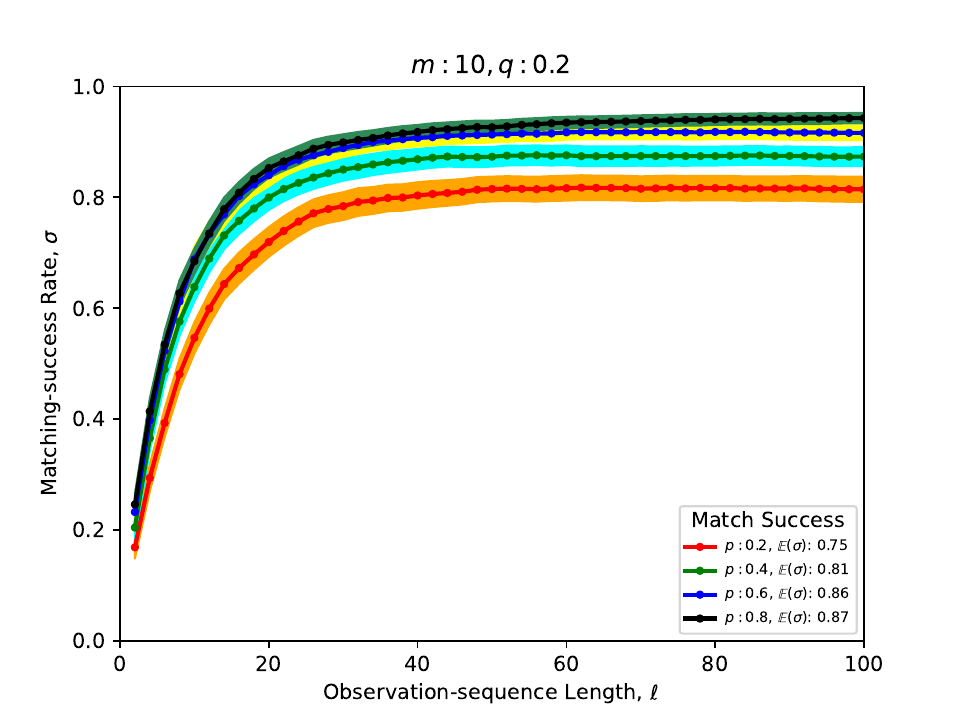} & \plt{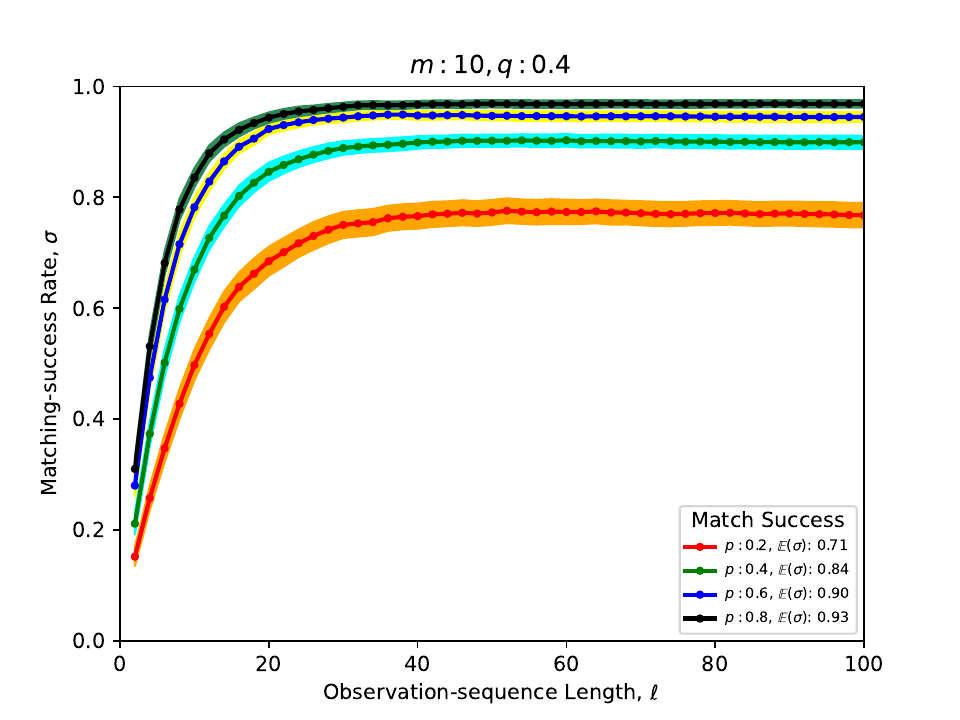} &
    \plt{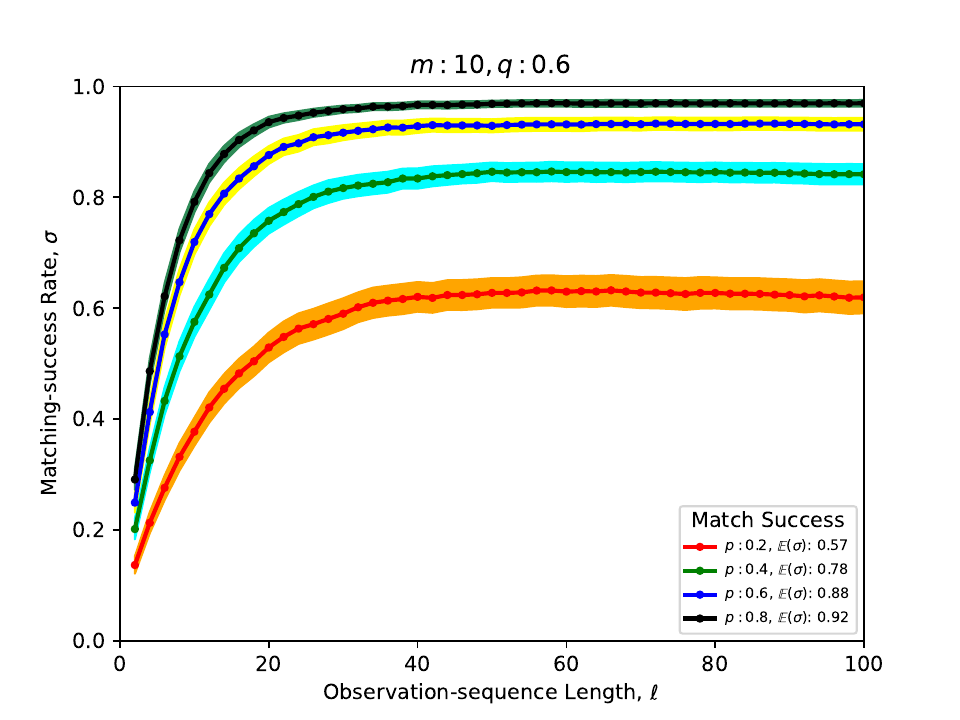} &
    \plt{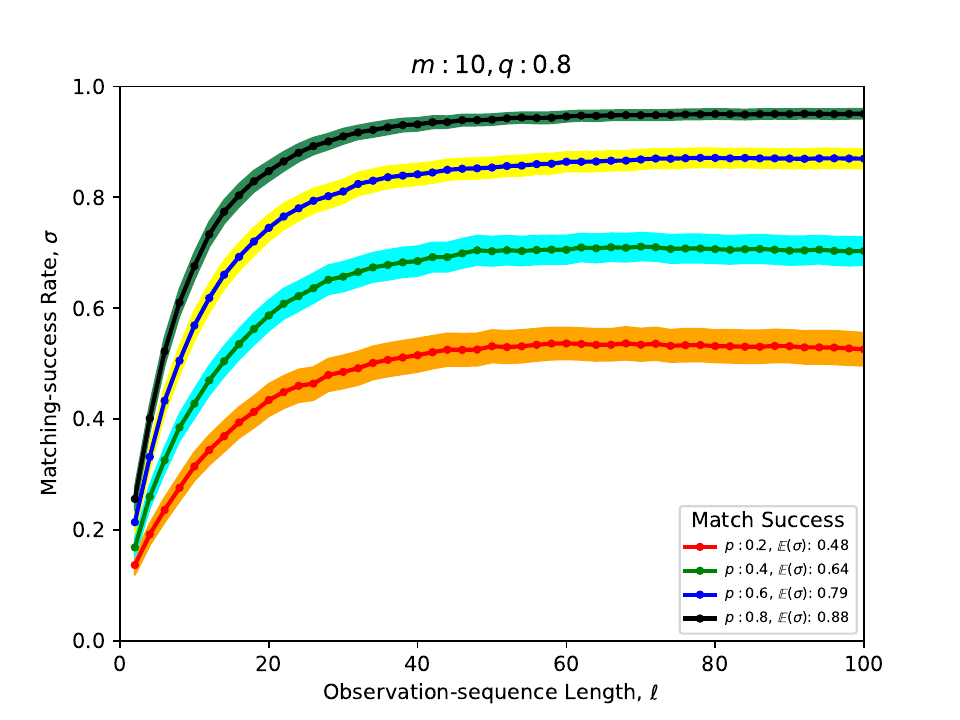} \\
    \plt{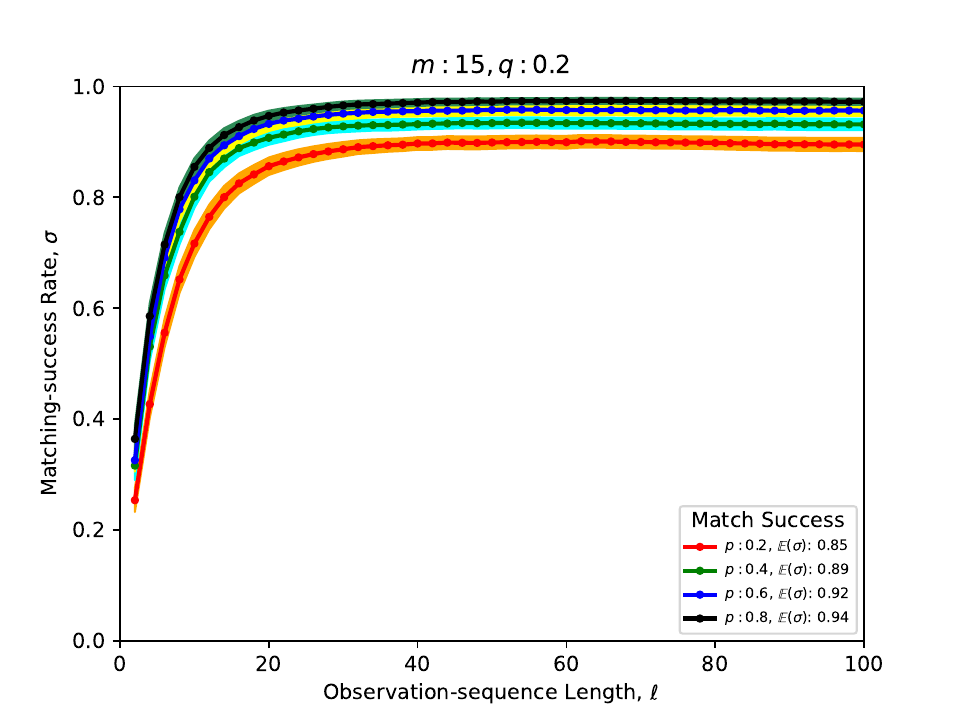} & \plt{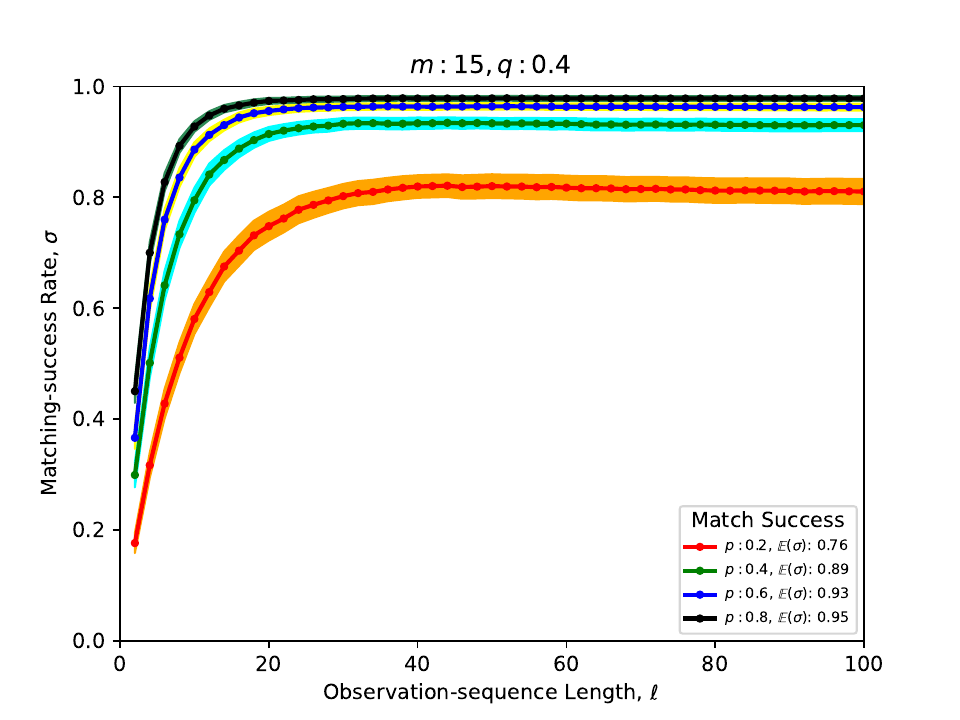} &
    \plt{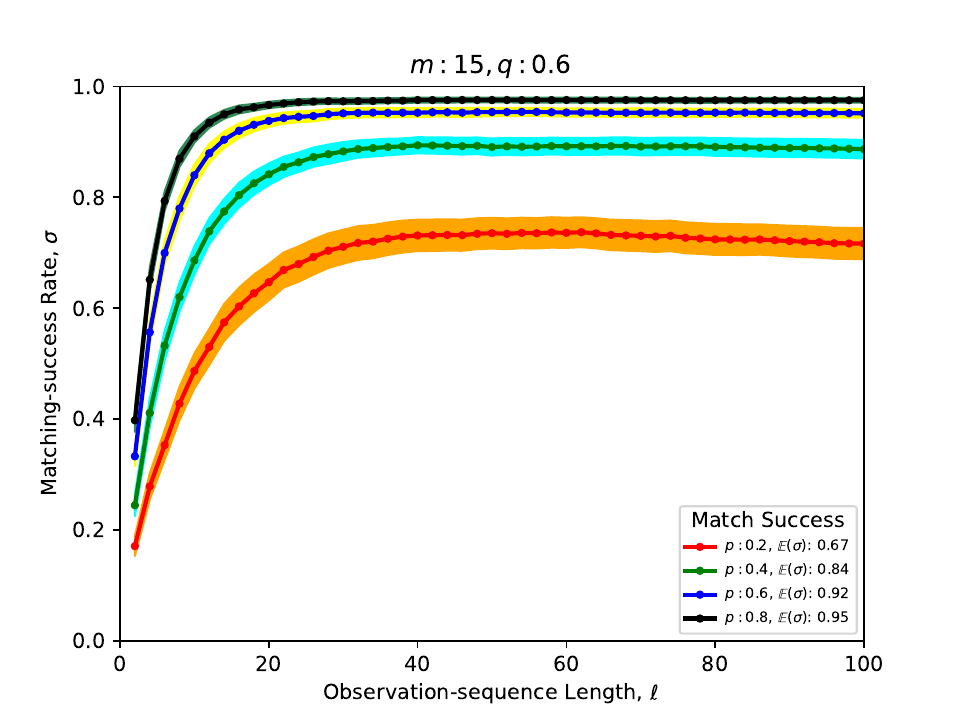} &
    \plt{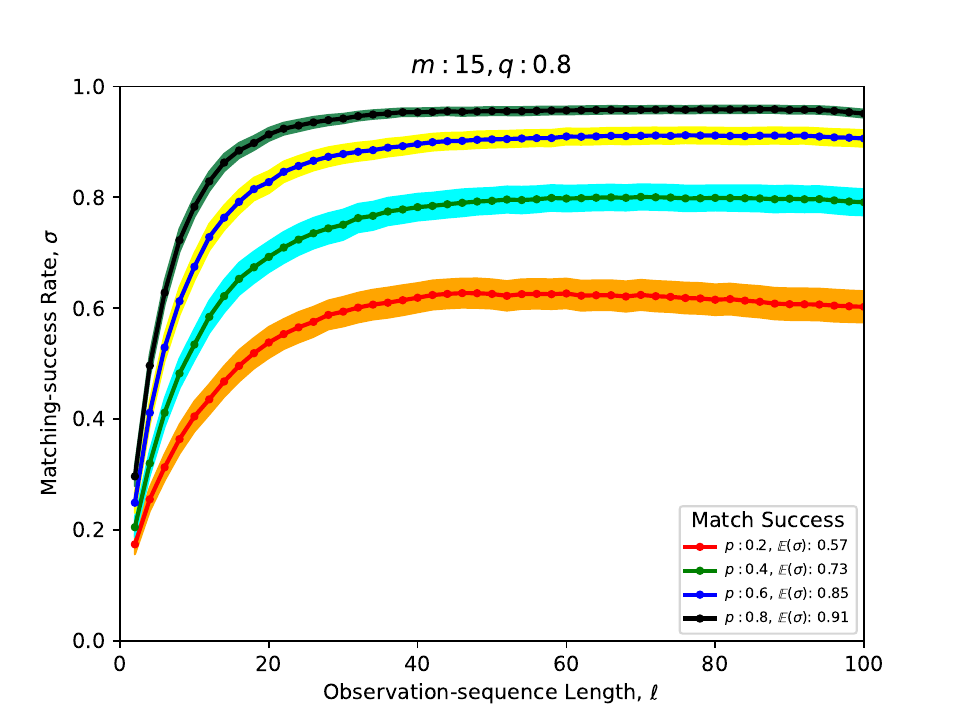} \\
\bottomrule 
\end{tabular*}
\caption{The figures show how matching-success rates change with observation sequence lengths when the number of resources $m$ and resource-map graph density $q$ are varied across each figure in the array. Each figure shows several individual plots for different settings of activity-graph densities $p$. Also shown are the $95\%$ confidence-bands around the sample means of the matching-success rates.}
\label{RI_3}
\end{figure*}

\begin{figure*}[t]\sffamily
\begin{tabular*}{\textwidth}{c@{}c@{}c@{}c@{}}
\toprule
    \plt{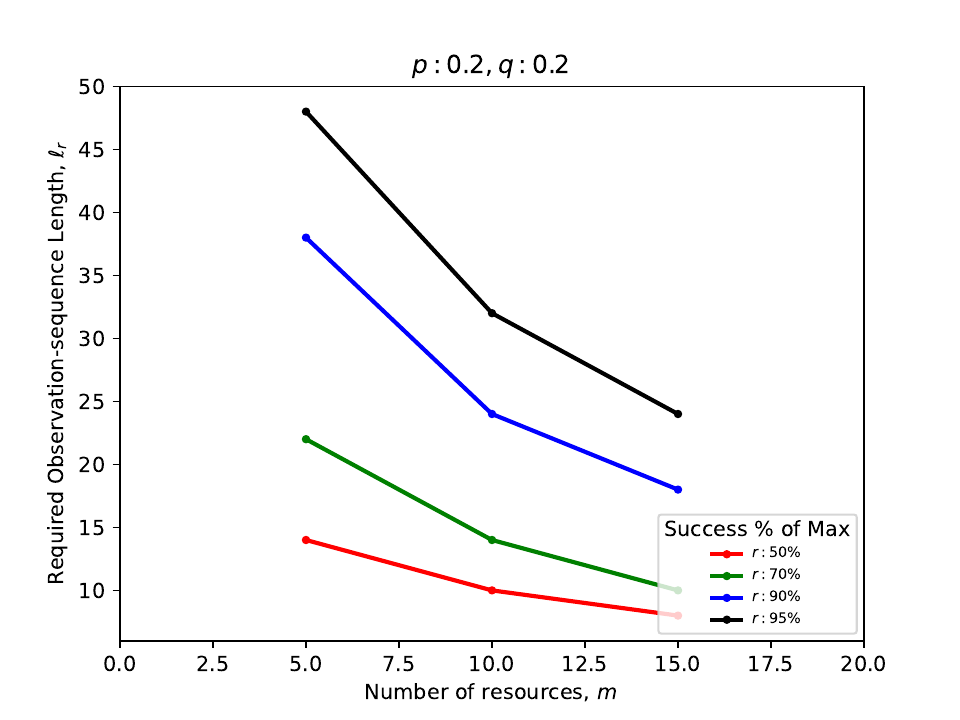} & \plt{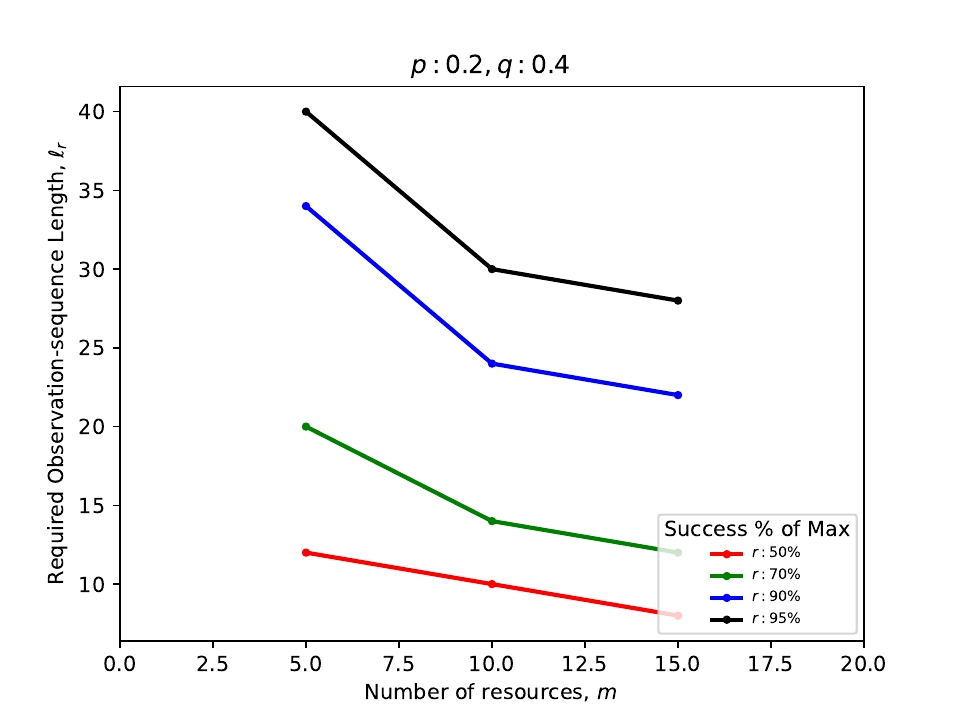} &
    \plt{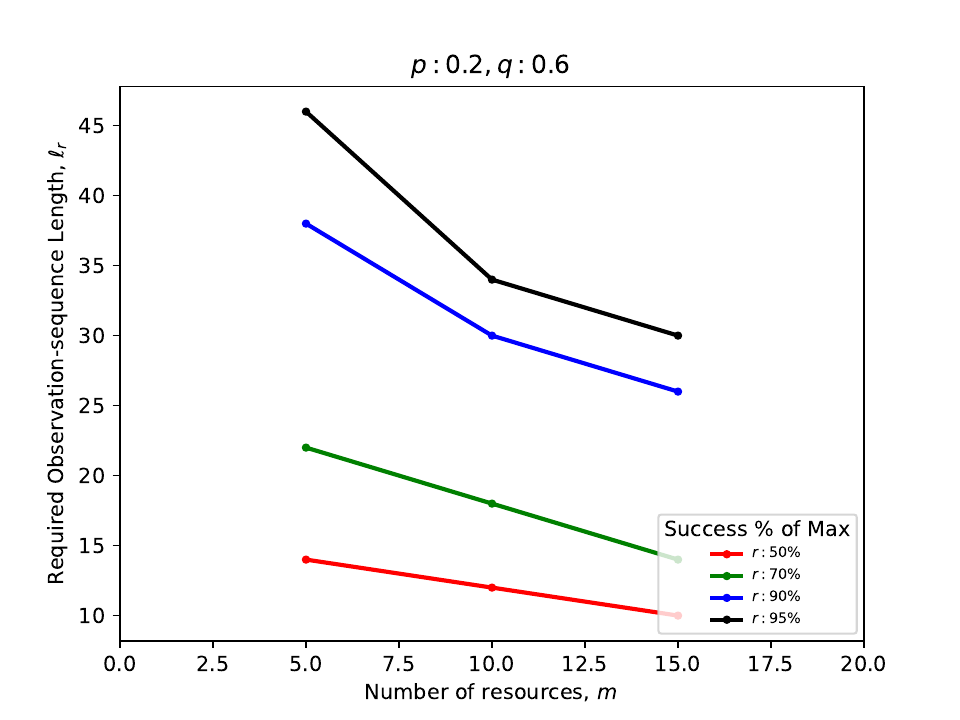} &
    \plt{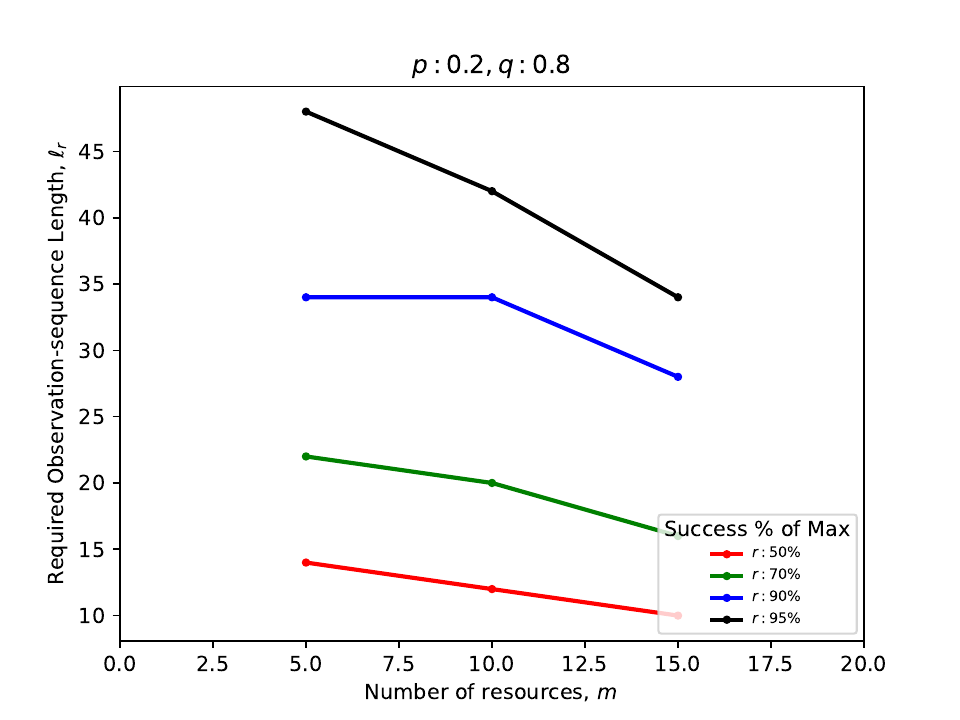} \\
    \plt{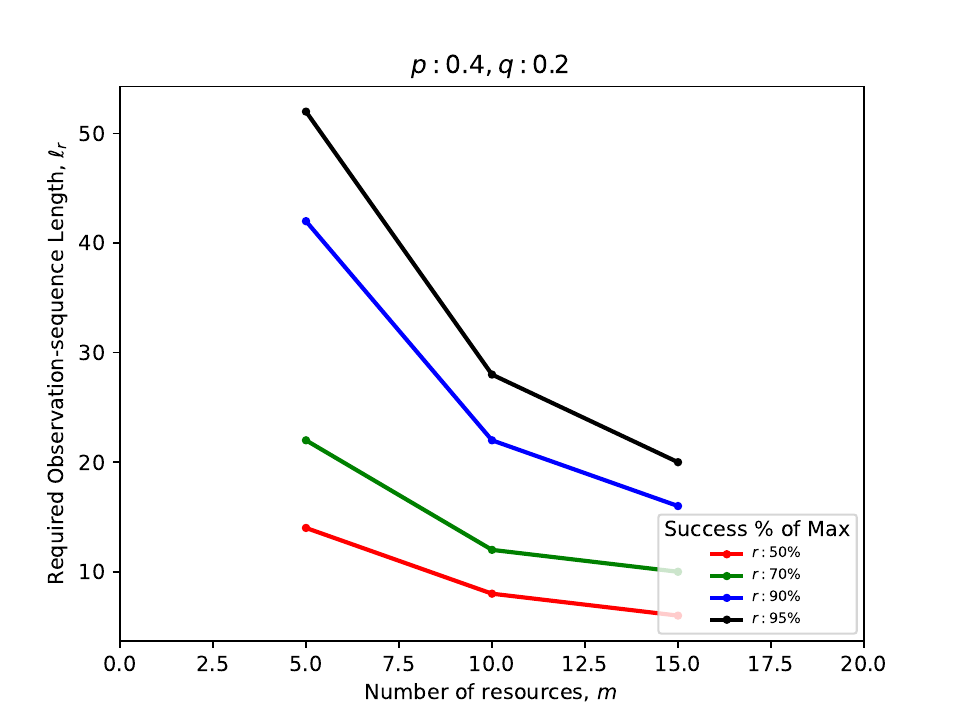} & \plt{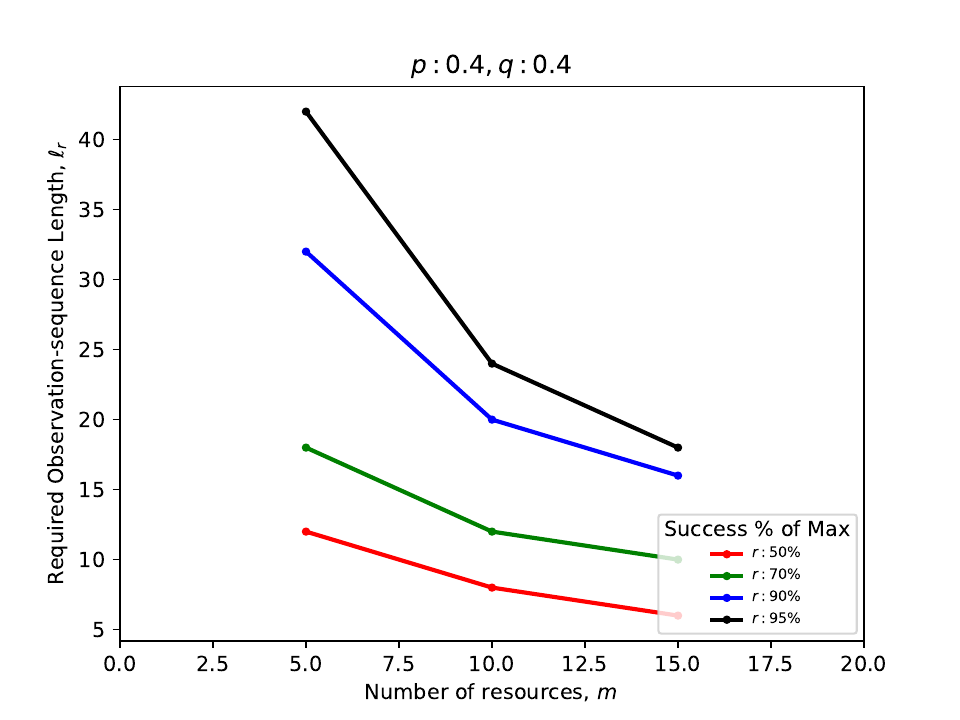} &
    \plt{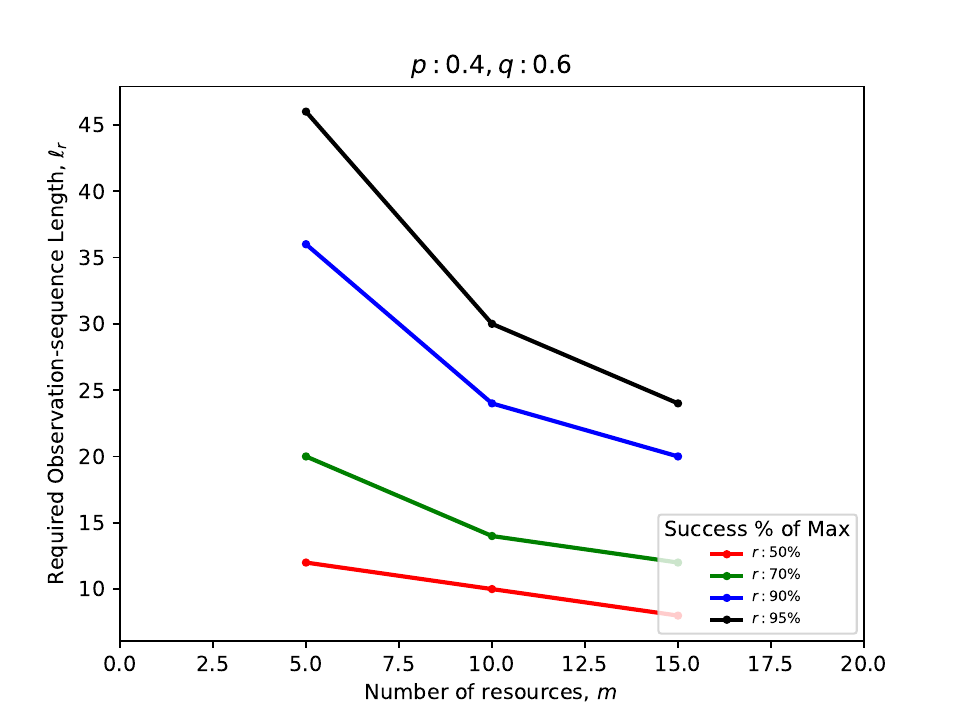} &
    \plt{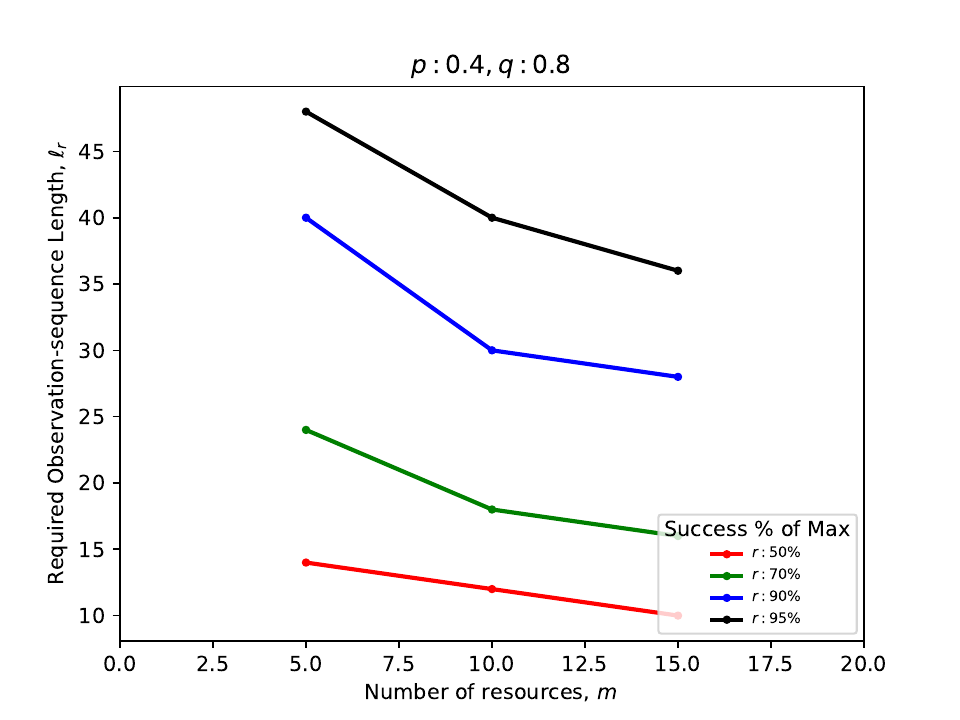} \\
    \plt{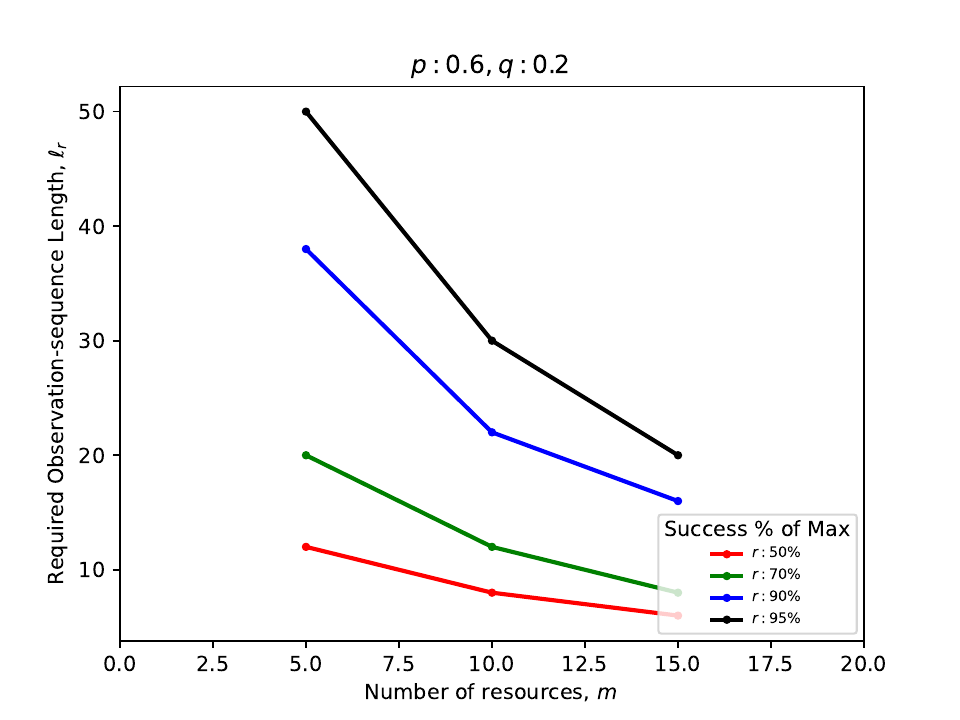} & \plt{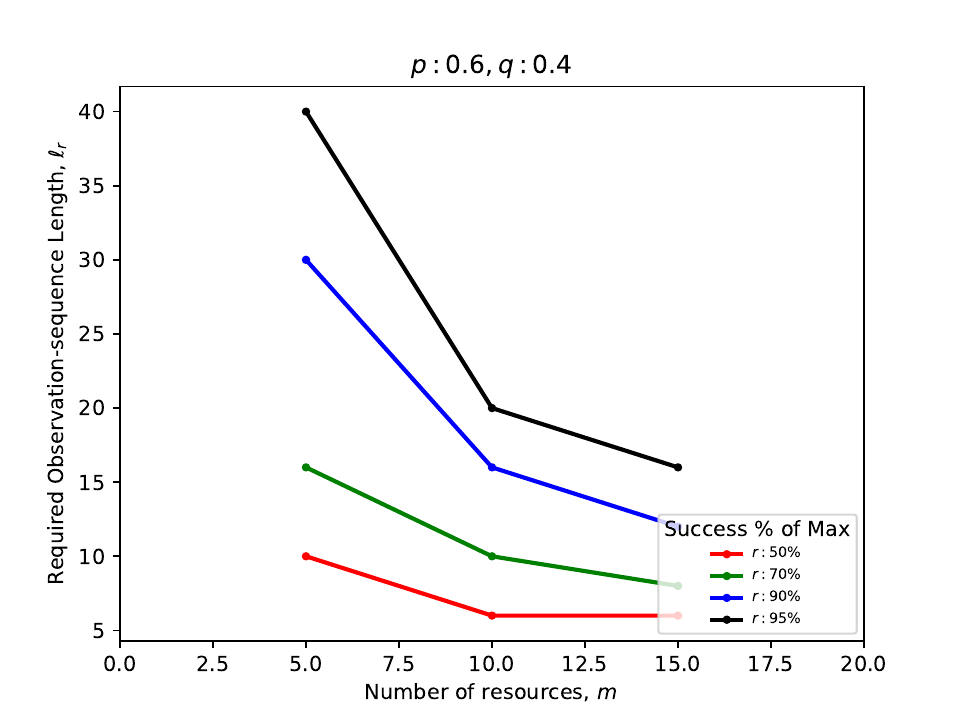} &
    \plt{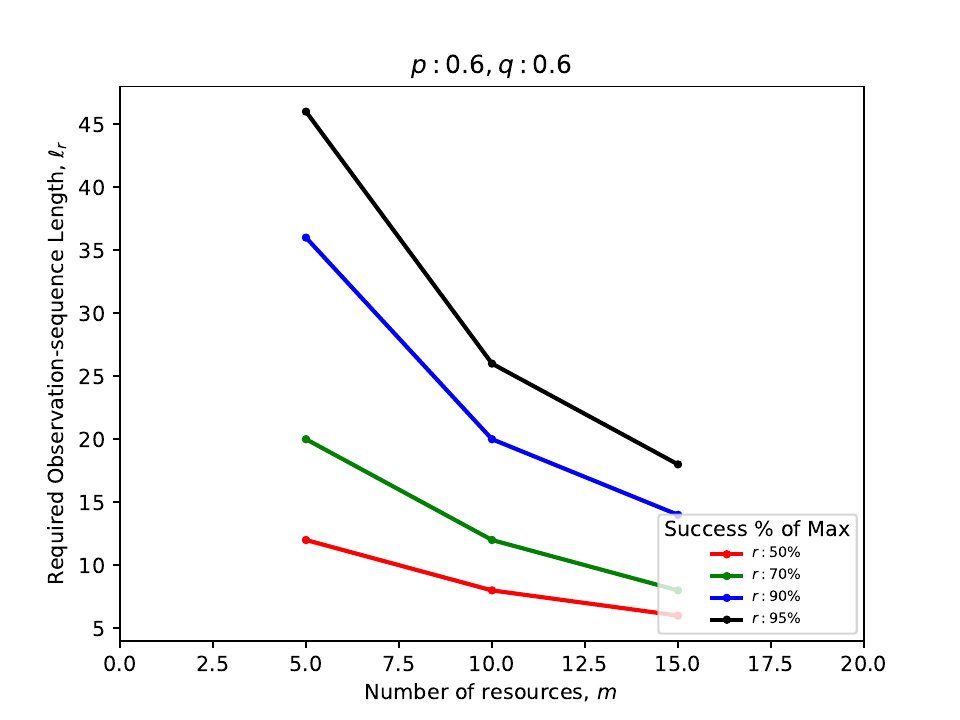} &
    \plt{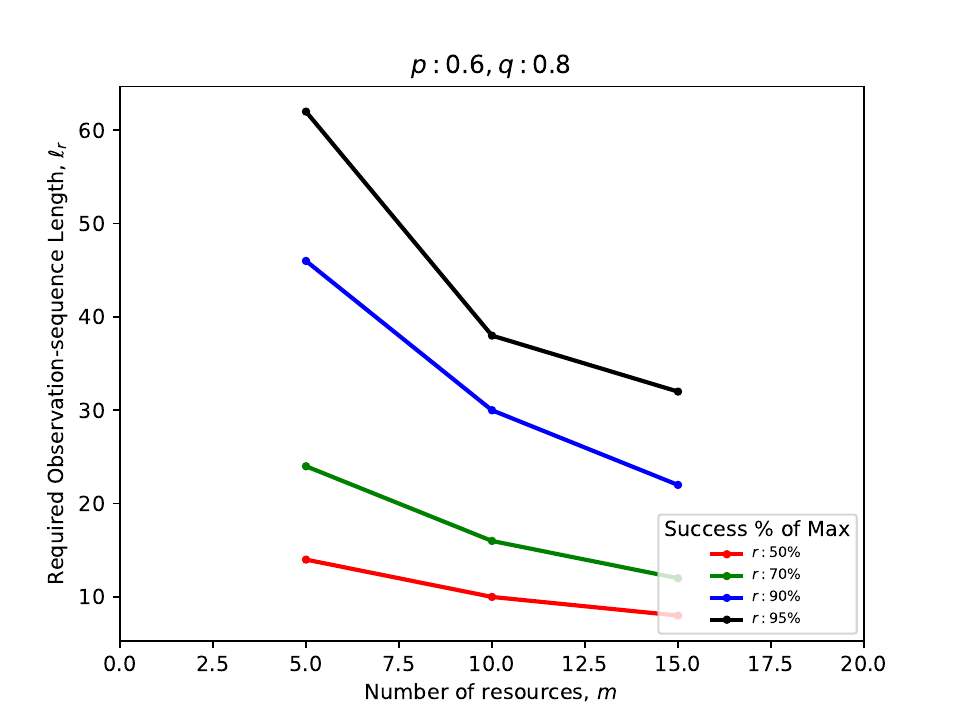} \\
    \plt{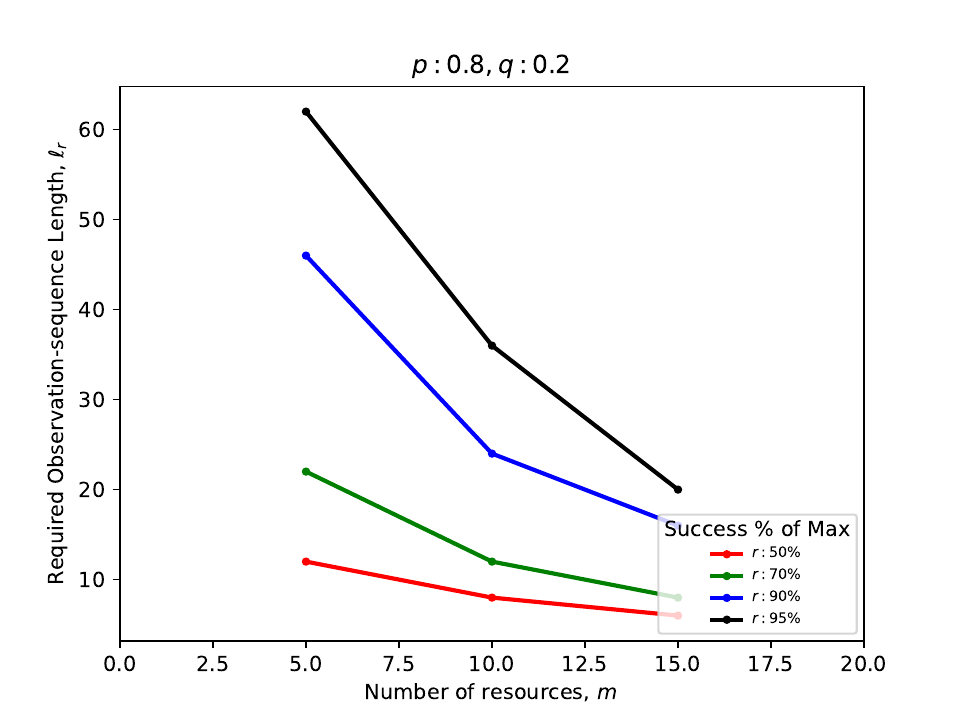} & \plt{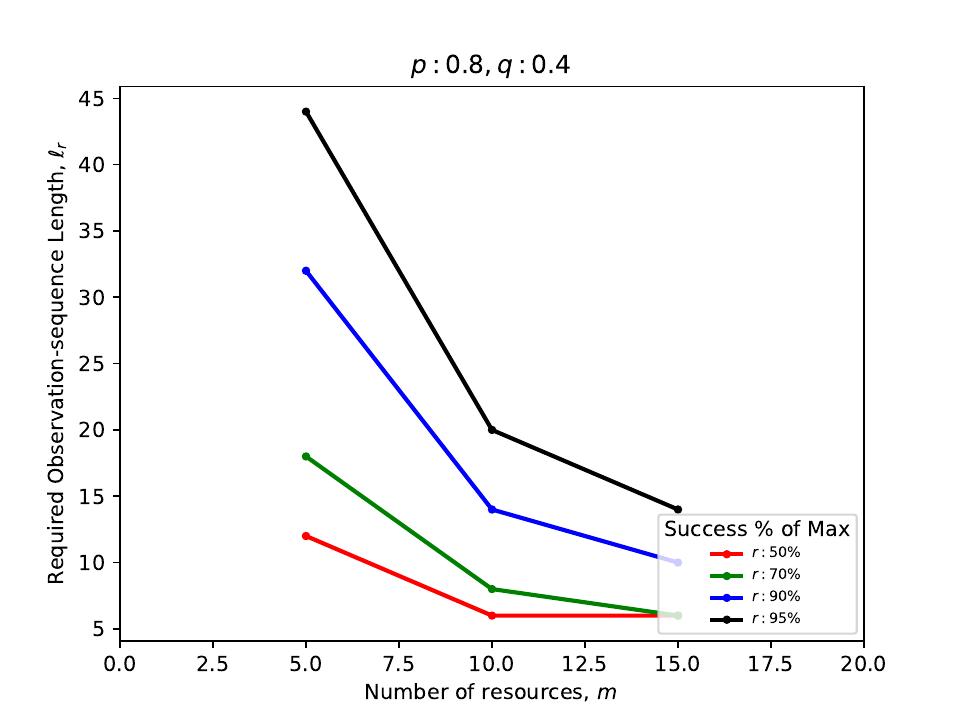} &
    \plt{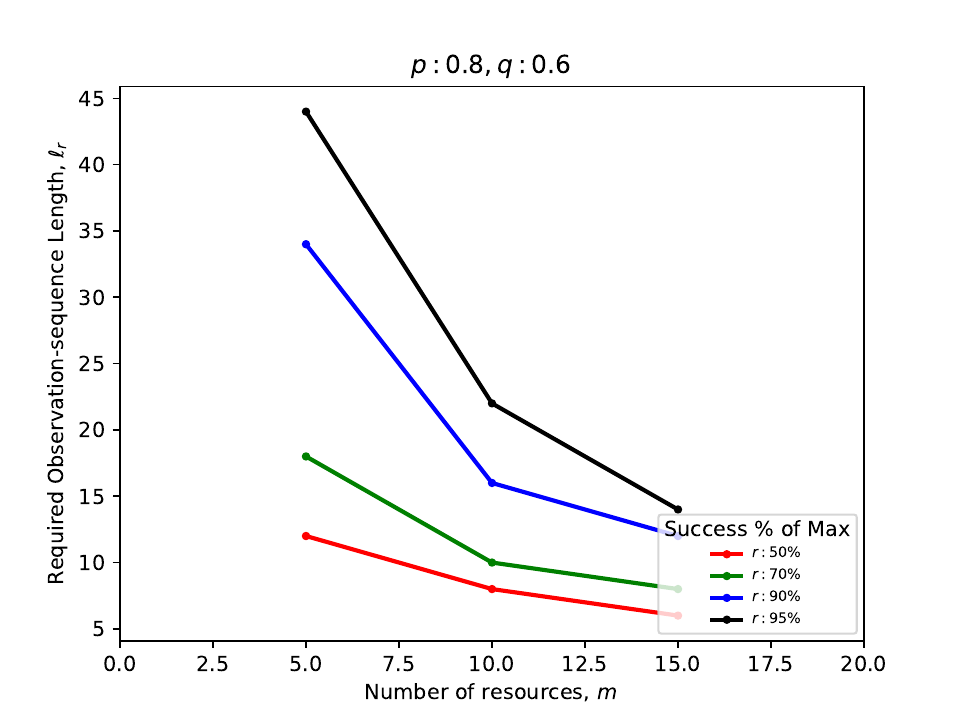} &
    \plt{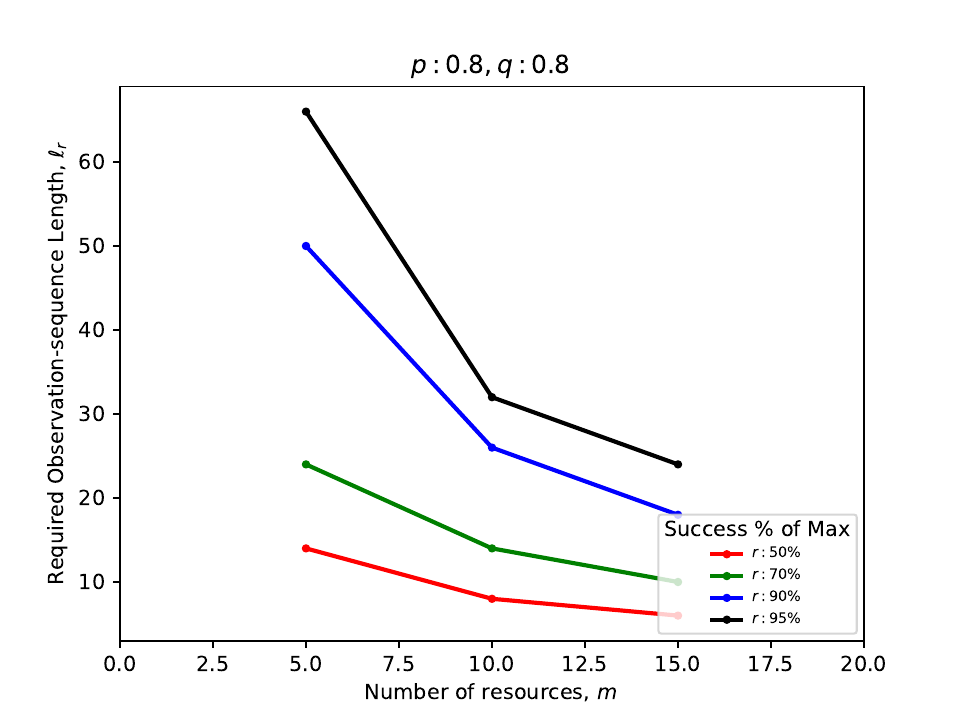} \\
\bottomrule 
\end{tabular*}
\caption{The figures show how required observation-sequence lengths $\ell_r$ change with number of resources $m$ when activity-graph density $p$ and resource-map density $q$ are varied across each figure in the array. Each figure shows several individual plots for different settings of number of resources $m$.}
\label{CL_1}
\end{figure*}

\begin{figure*}[t]\sffamily
\begin{tabular*}{\textwidth}{c@{}c@{}c@{}c@{}}
\toprule
    \plt{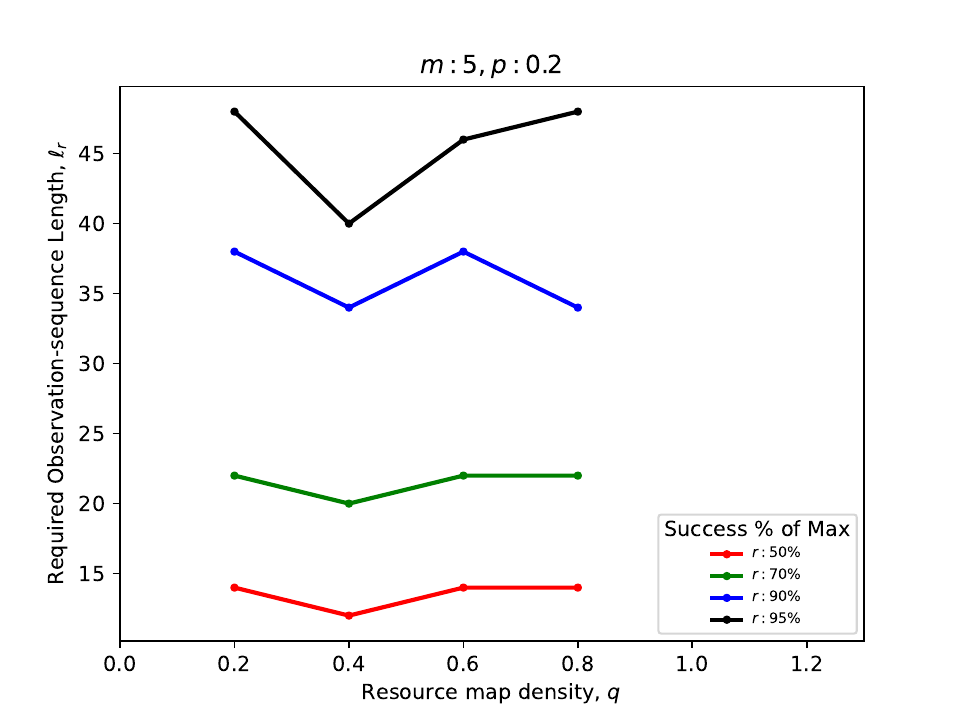} & \plt{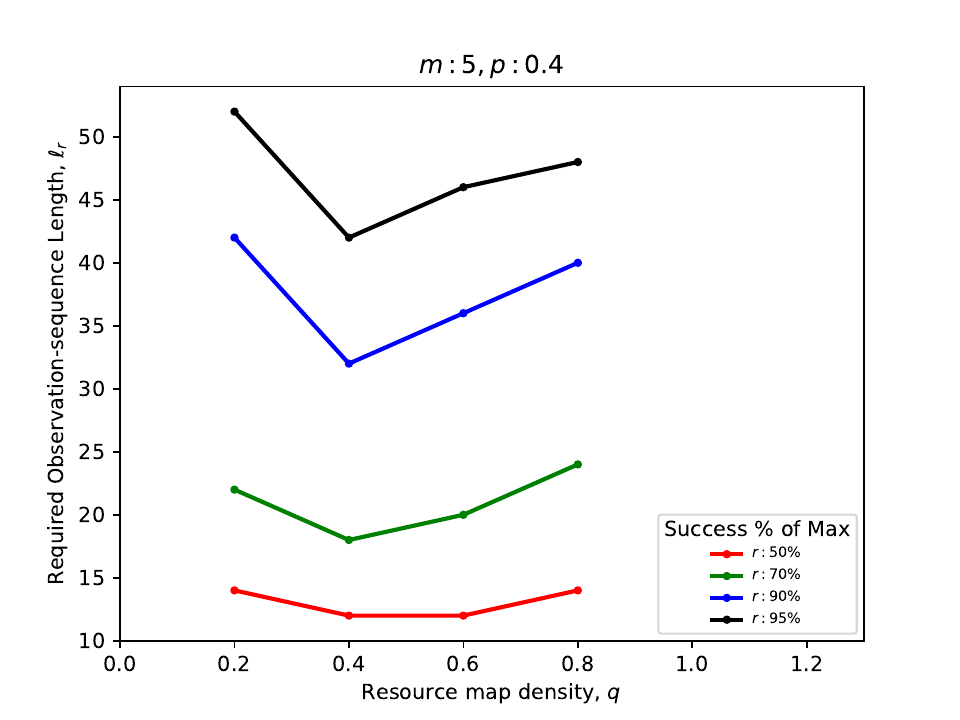} &
    \plt{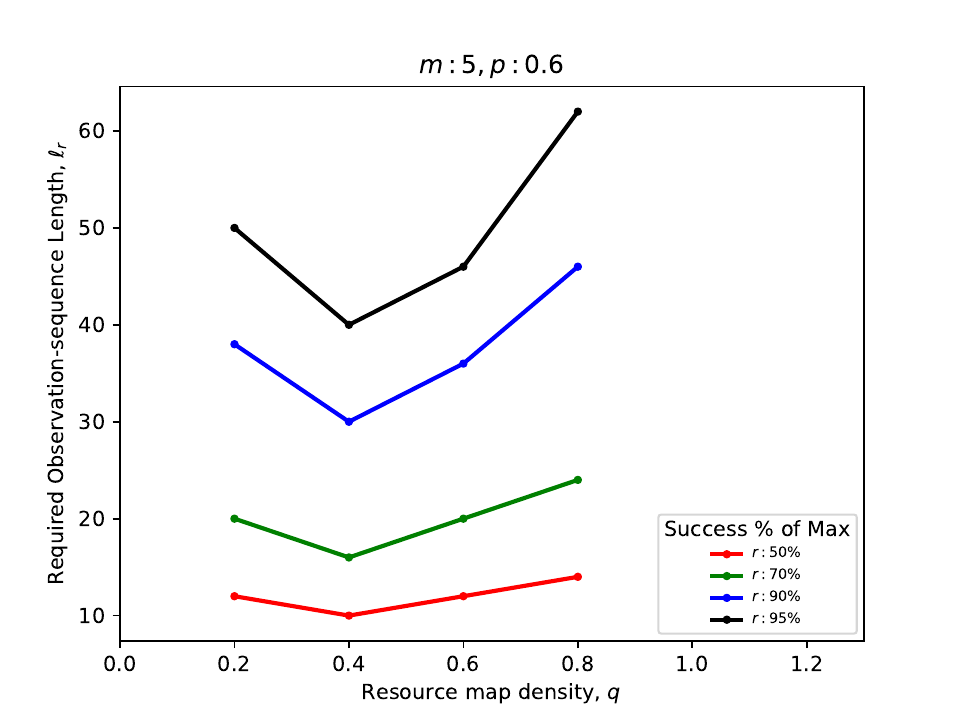} &
    \plt{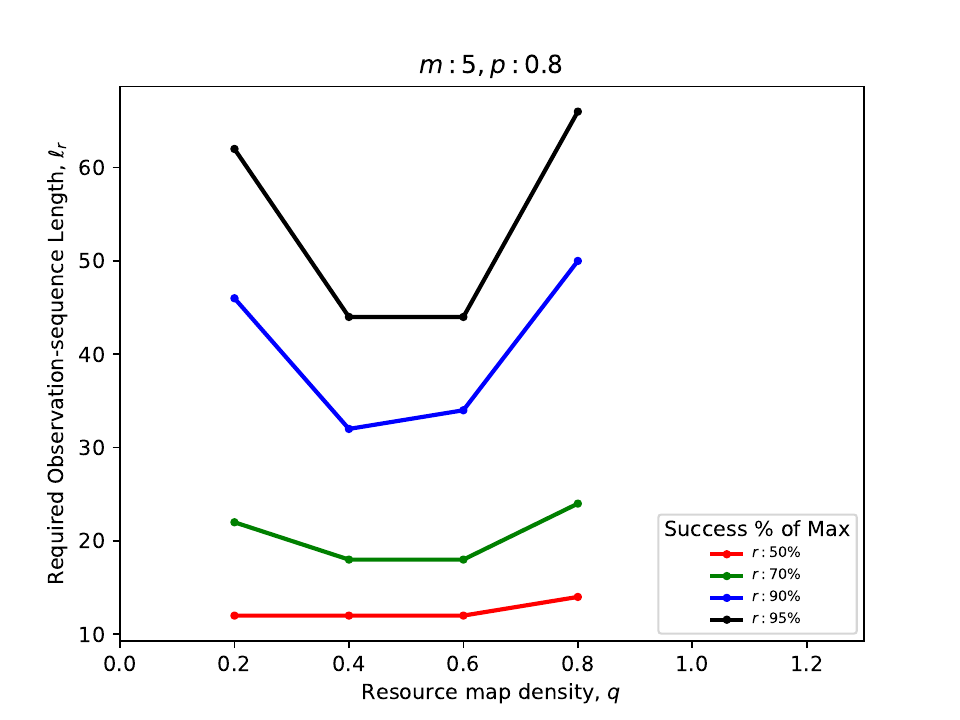} \\
    \plt{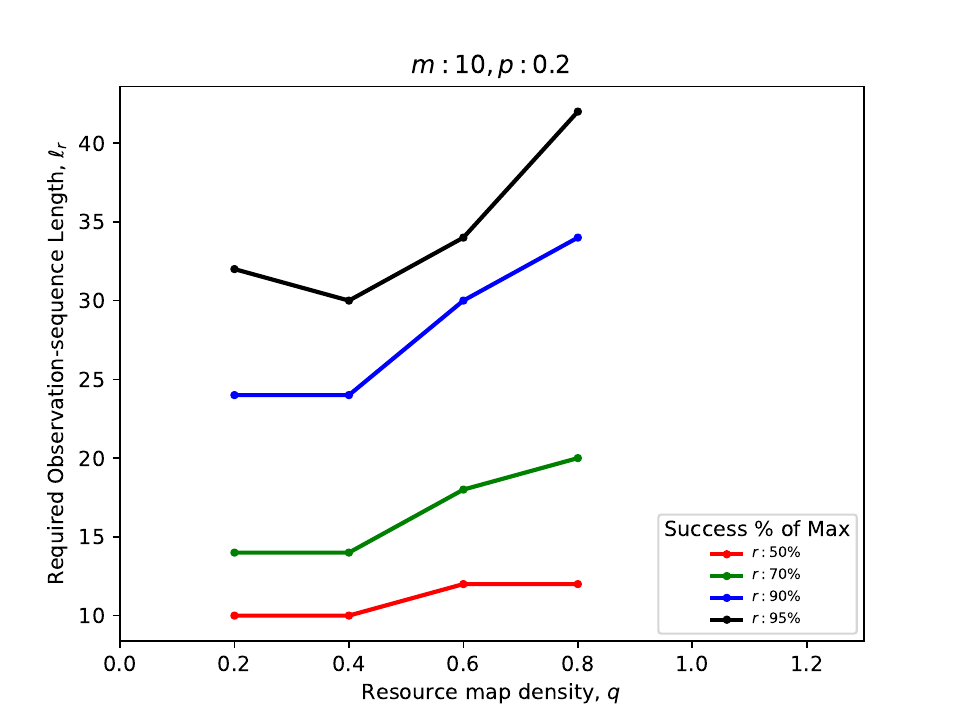} & \plt{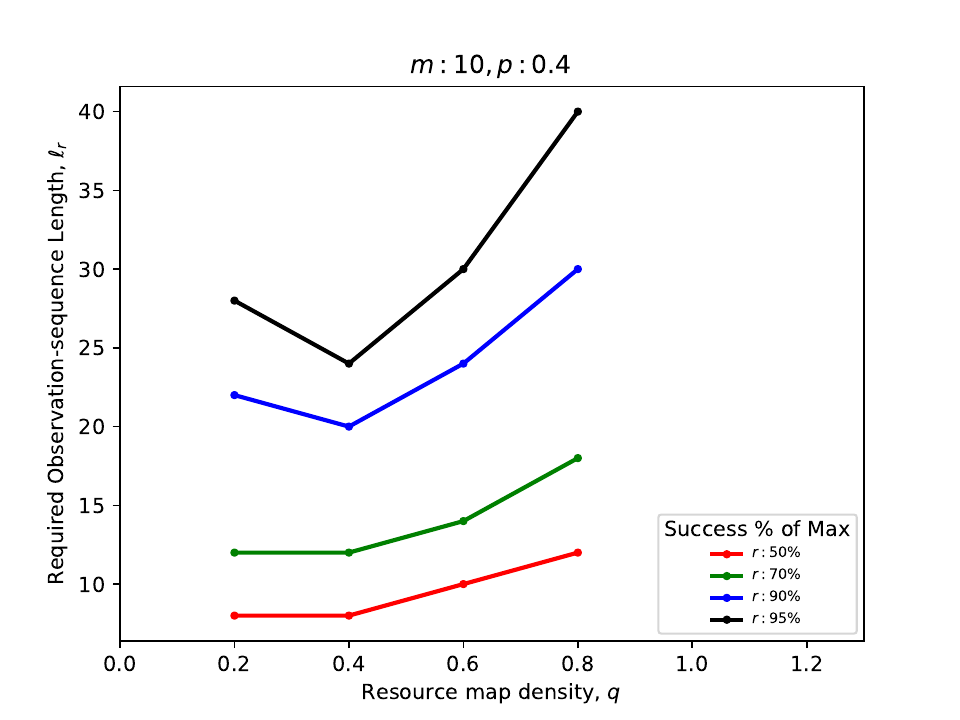} &
    \plt{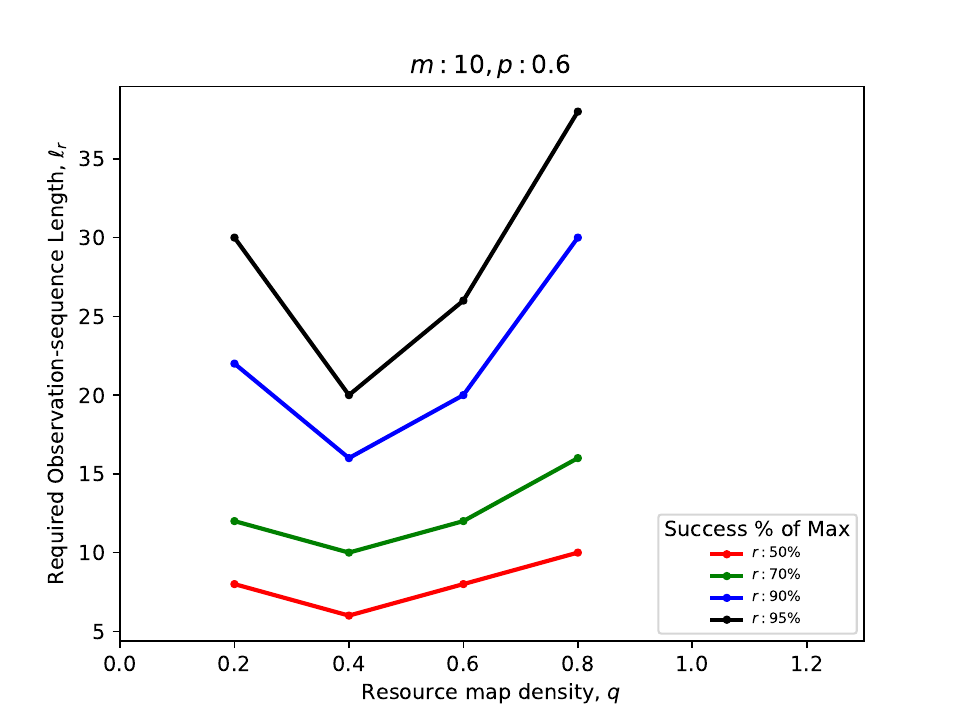} &
    \plt{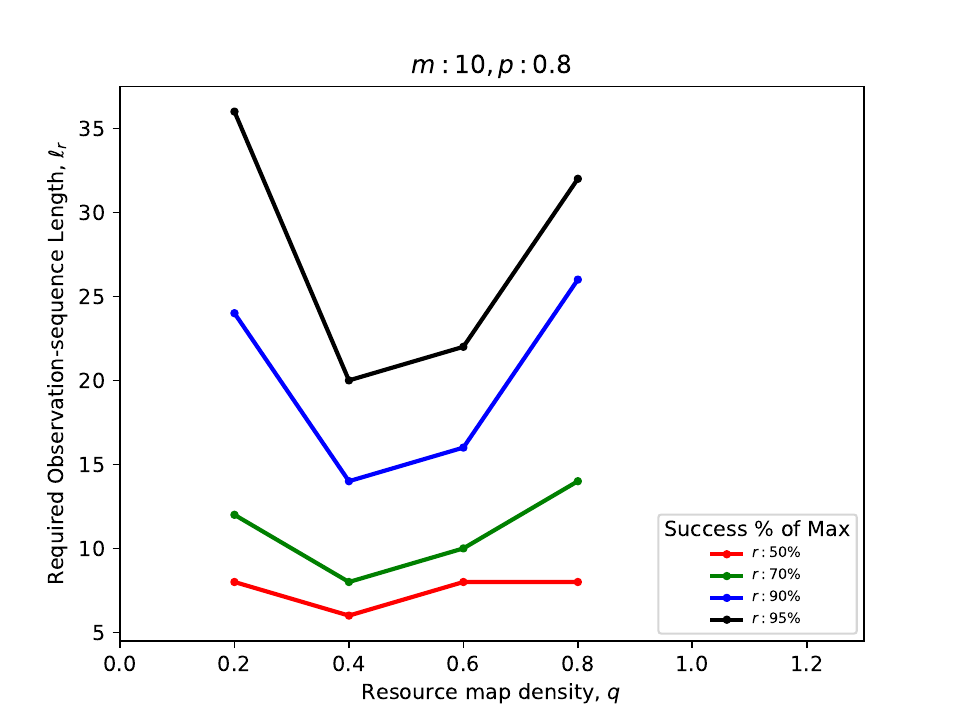} \\
    \plt{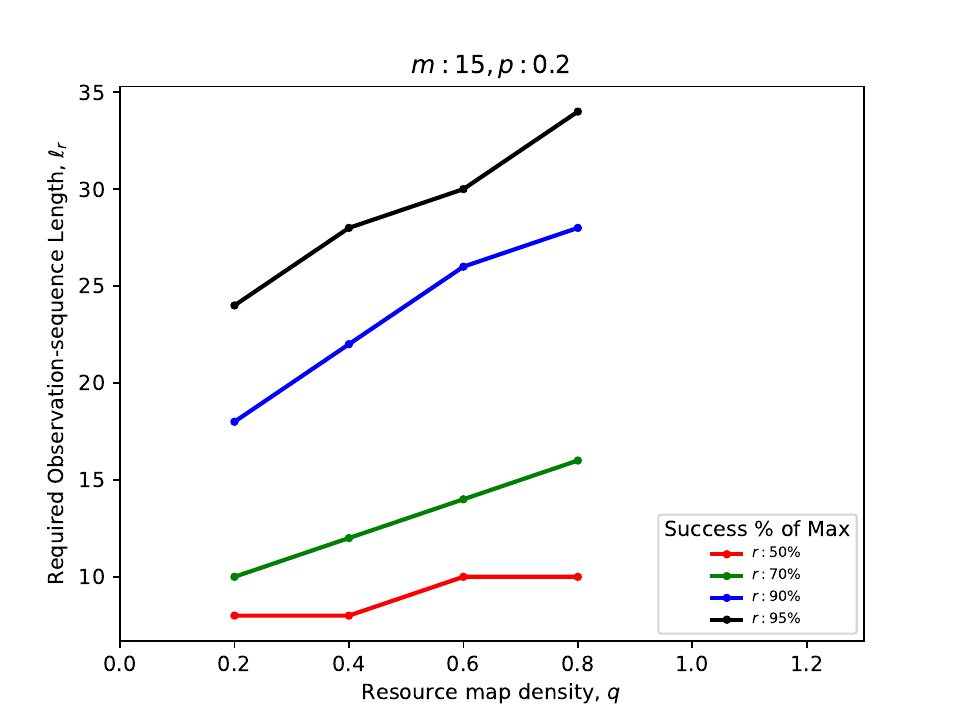} & \plt{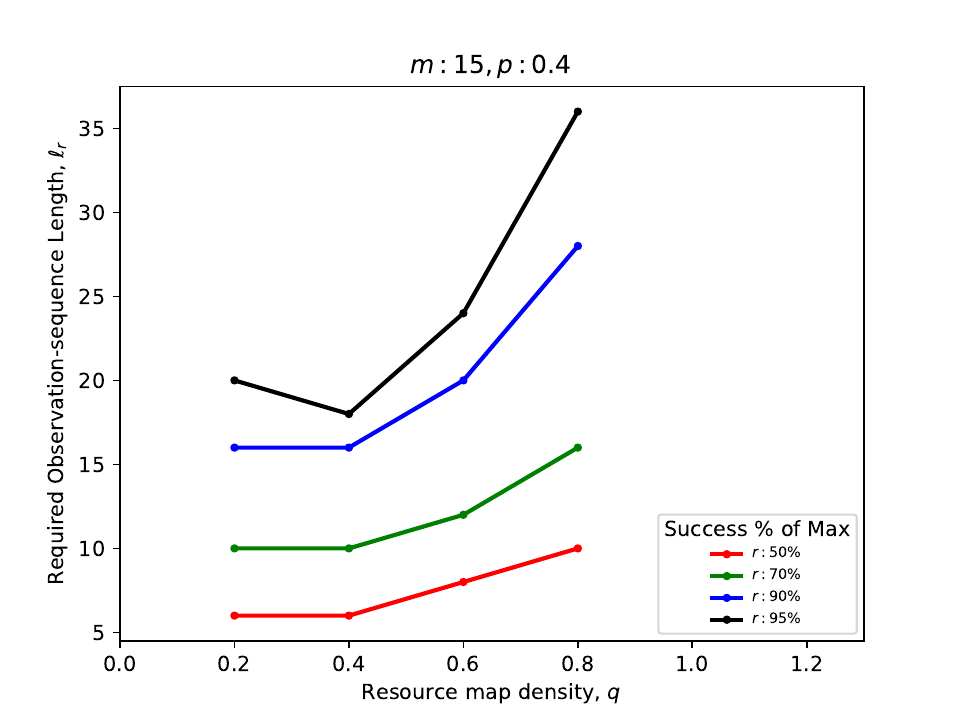} &
    \plt{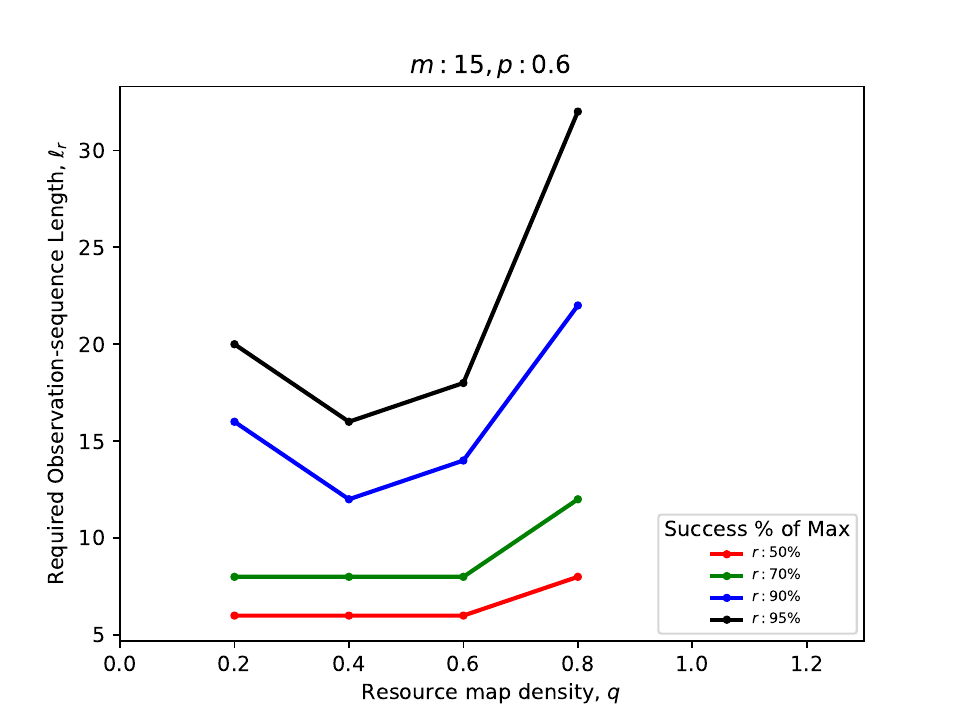} &
    \plt{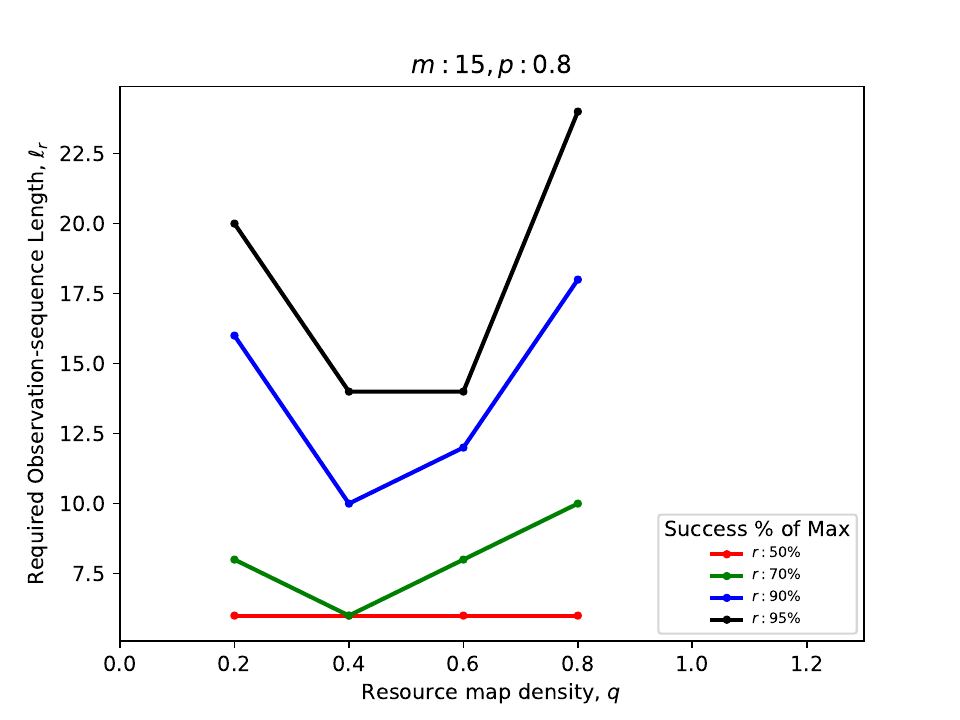} \\
\bottomrule 
\end{tabular*}
\caption{The figures show how required observation-sequence lengths $\ell_r$ change with resource-map density $q$ when the number of resources $m$ and activity-graph density $p$ are varied across each figure in the array. Each figure shows several individual plots for different settings of resource-map graph densities $q$.}
\label{CL_2}
\end{figure*}

\begin{figure*}[t]\sffamily
\begin{tabular*}{\textwidth}{c@{}c@{}c@{}c@{}}
\toprule
    \plt{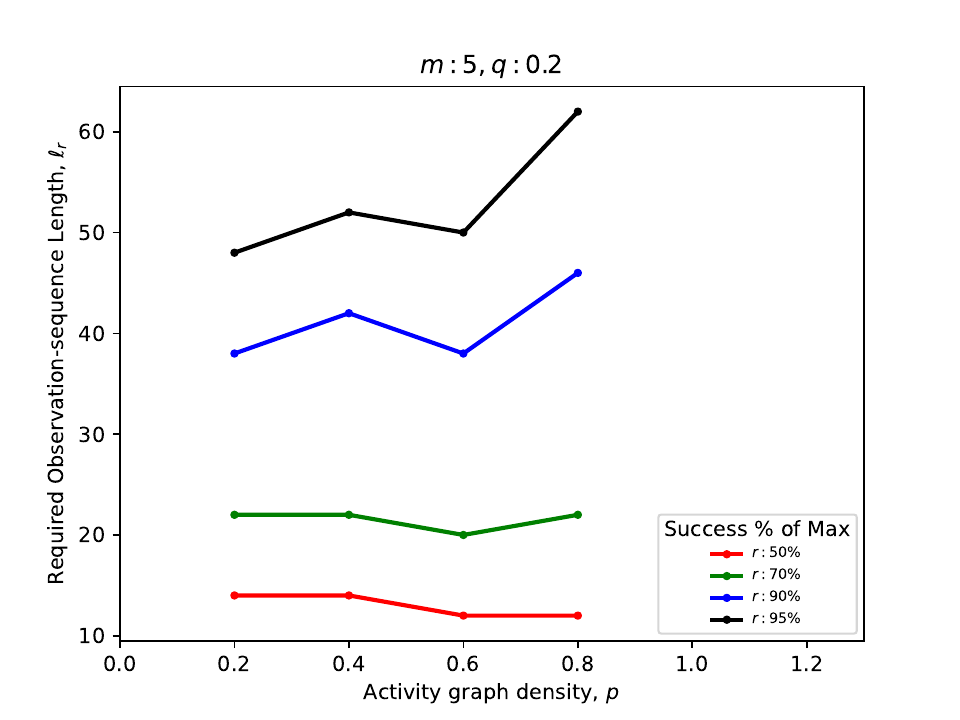} & \plt{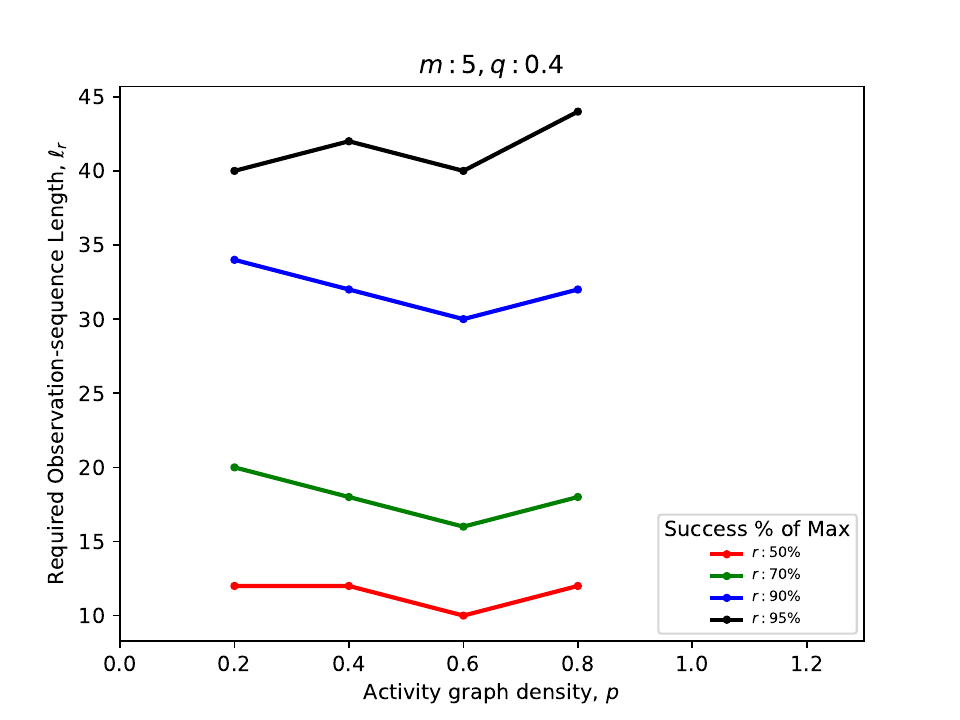} &
    \plt{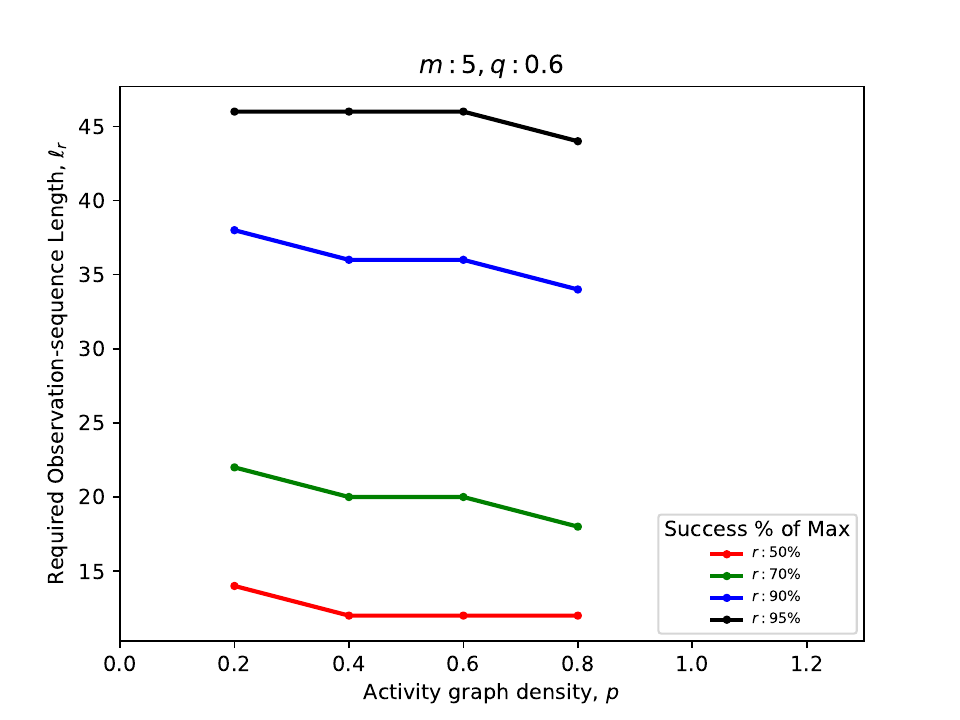} &
    \plt{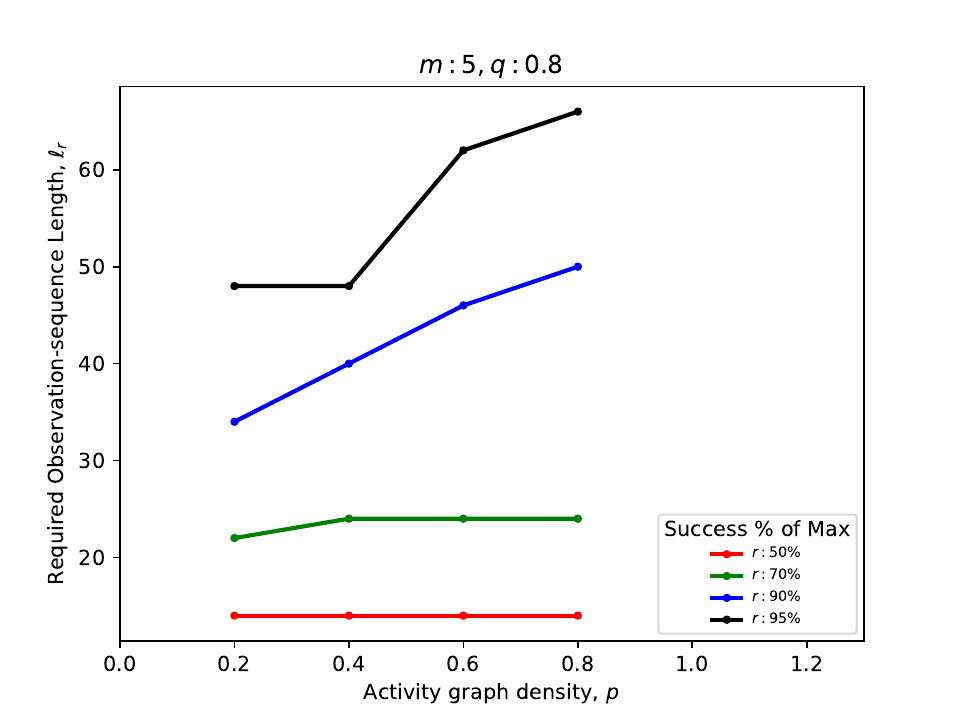} \\
    \plt{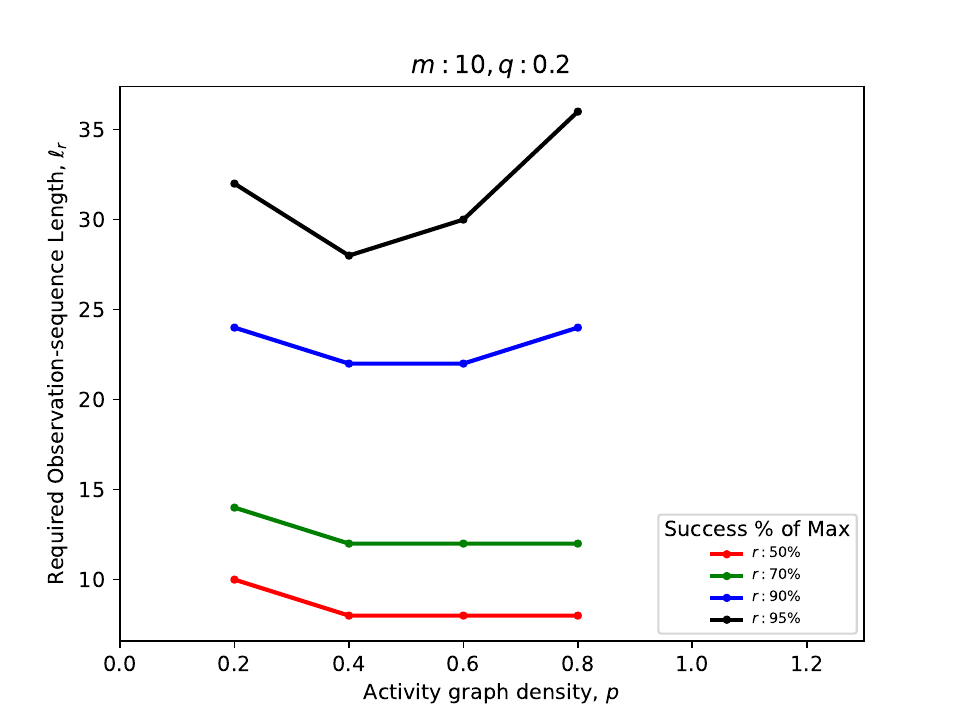} & \plt{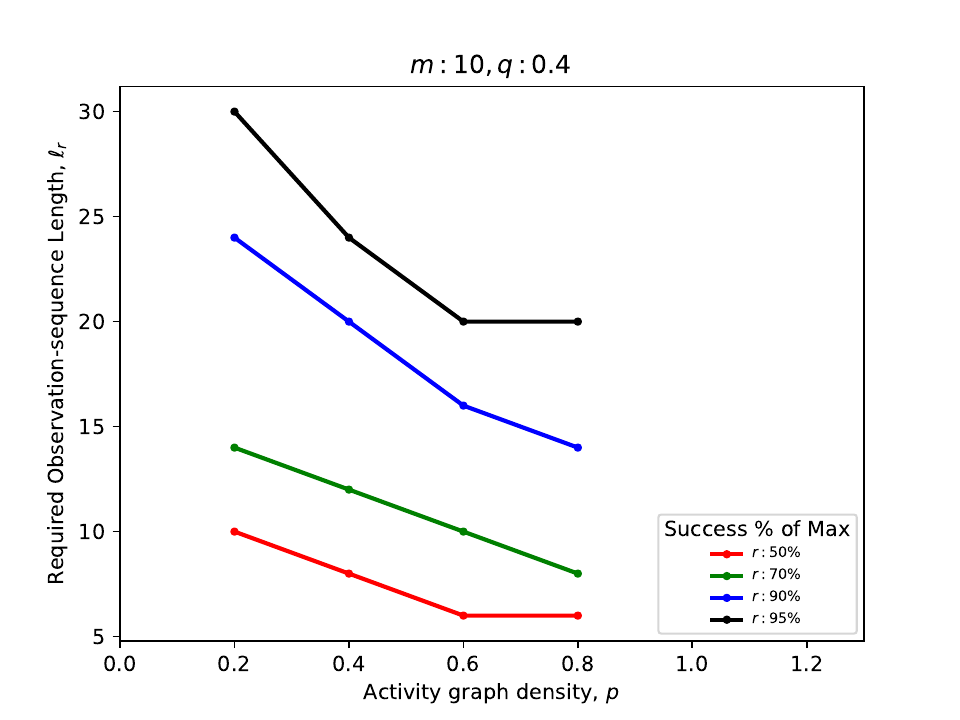} &
    \plt{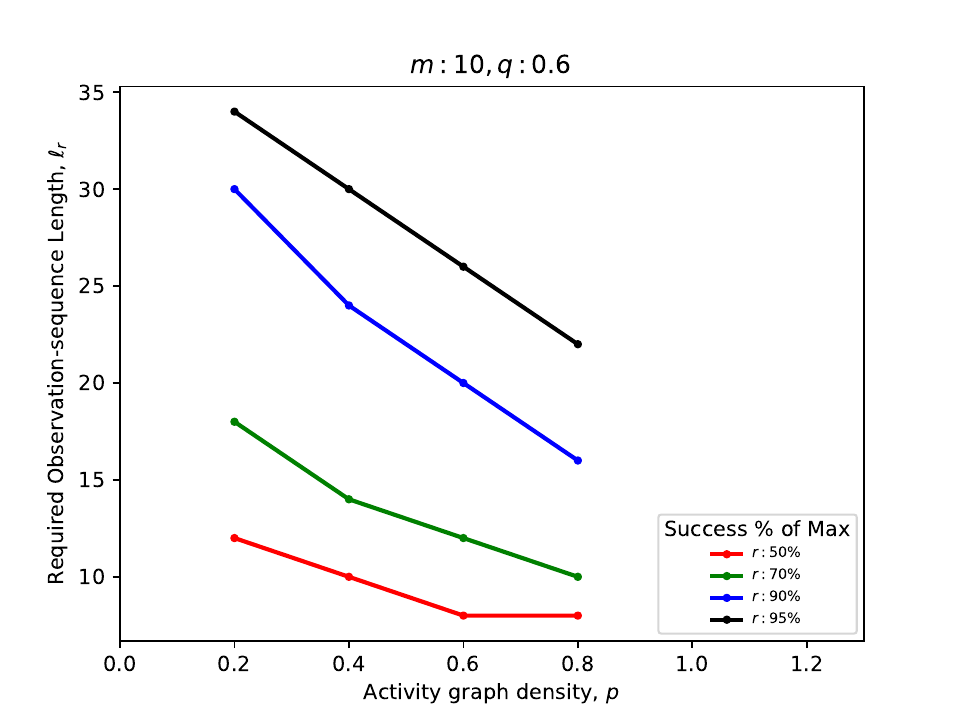} &
    \plt{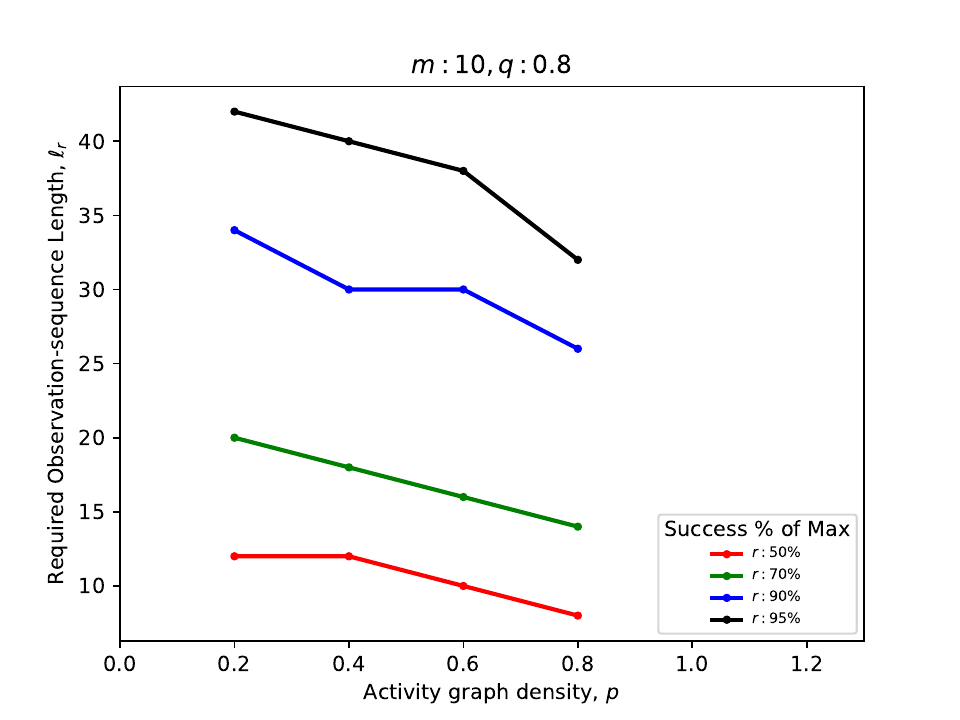} \\
    \plt{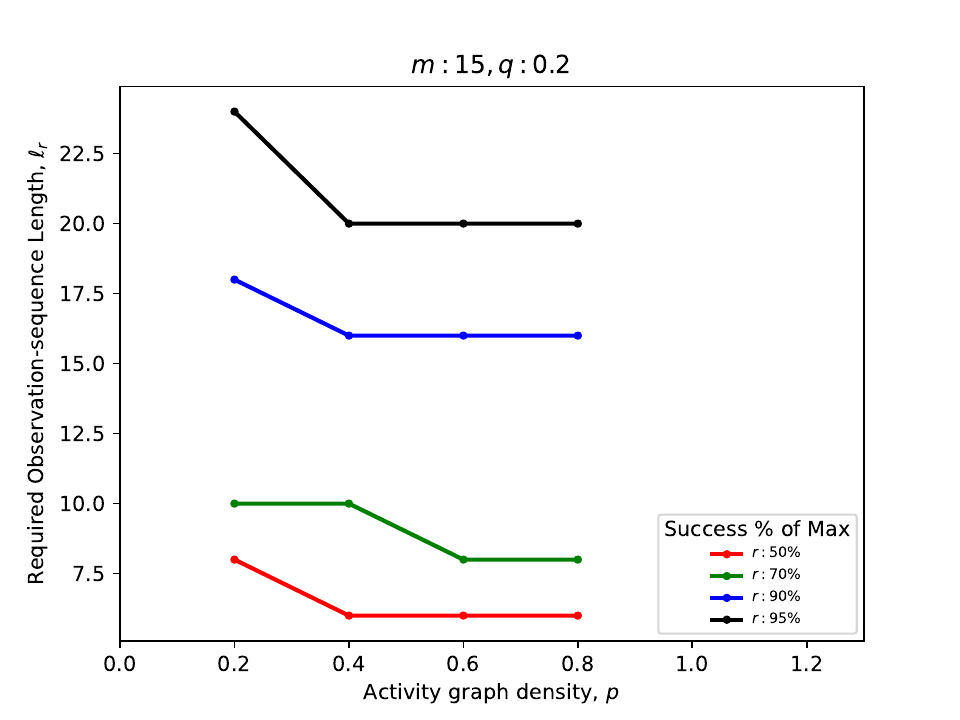} & \plt{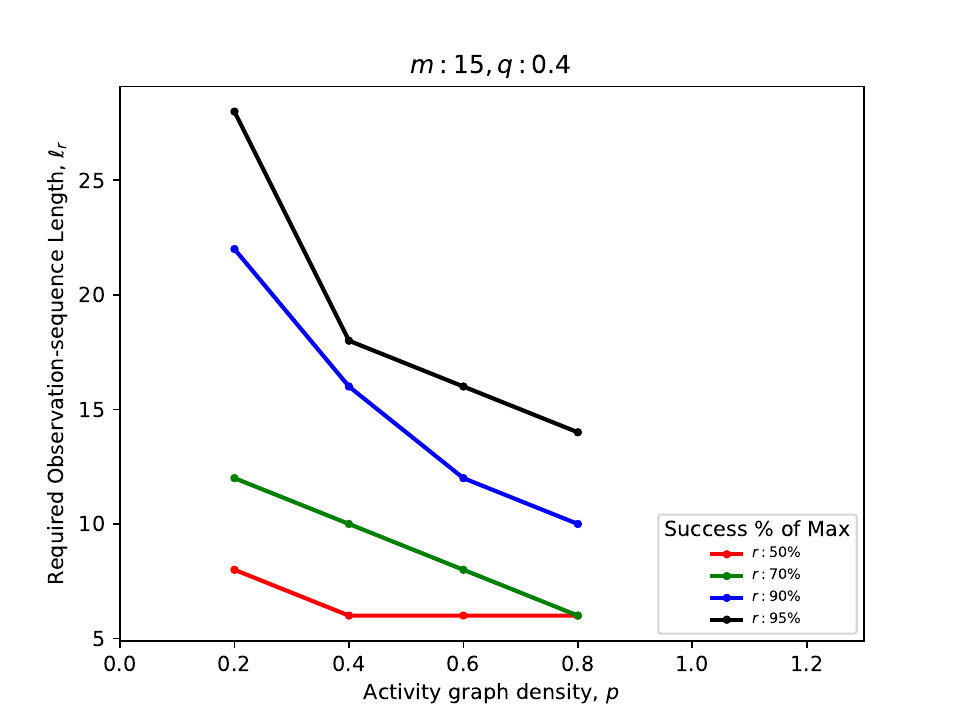} &
    \plt{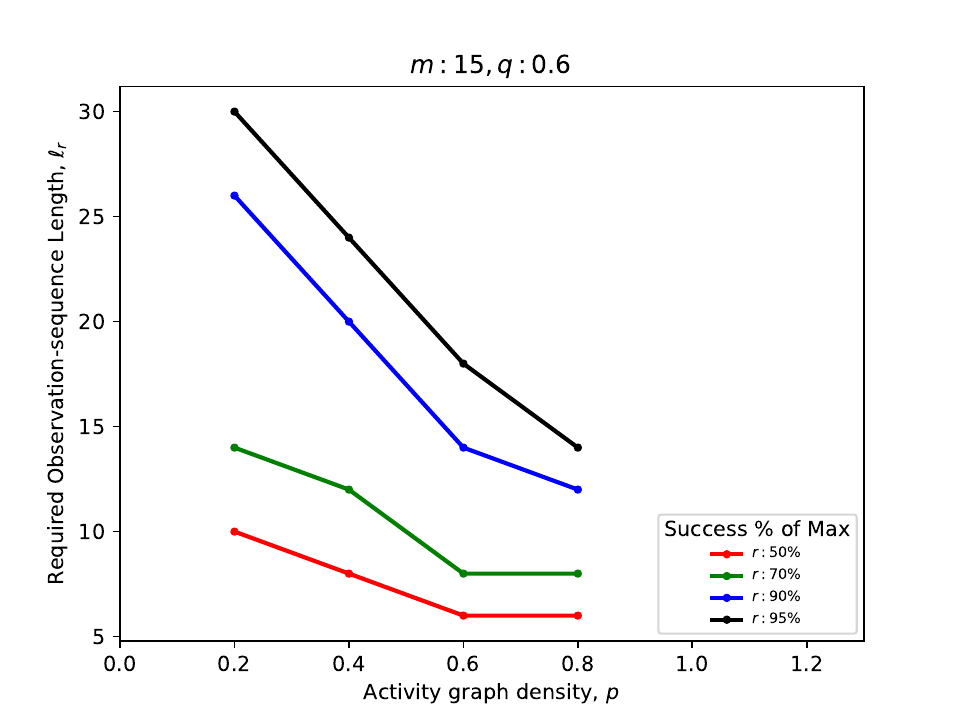} &
    \plt{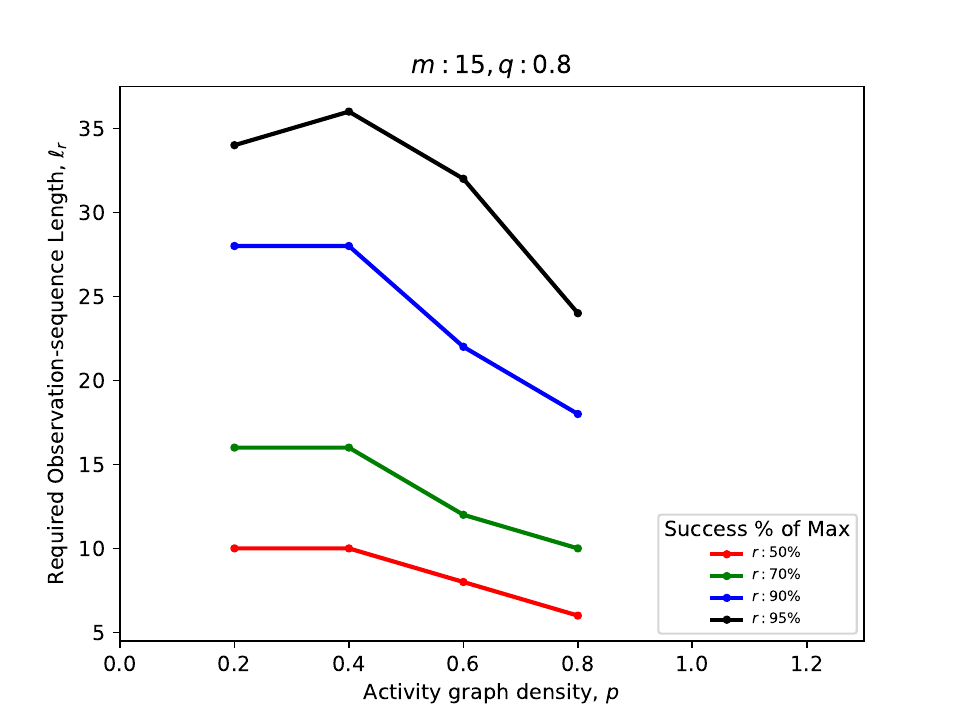} \\
\bottomrule 
\end{tabular*}
\caption{The figures show how required observation-sequence lengths $\ell_r$ change with activity-graph density $p$ when the number of resources $m$ and resource-map graph density $q$ are varied across each figure in the array. Each figure shows several individual plots for different settings of activity-graph densities $p$.}
\label{CL_3}
\end{figure*}

\begin{figure*}[t]\sffamily
\begin{tabular*}{\textwidth}{c@{}c@{}c@{}c@{}}
\toprule
    \plt{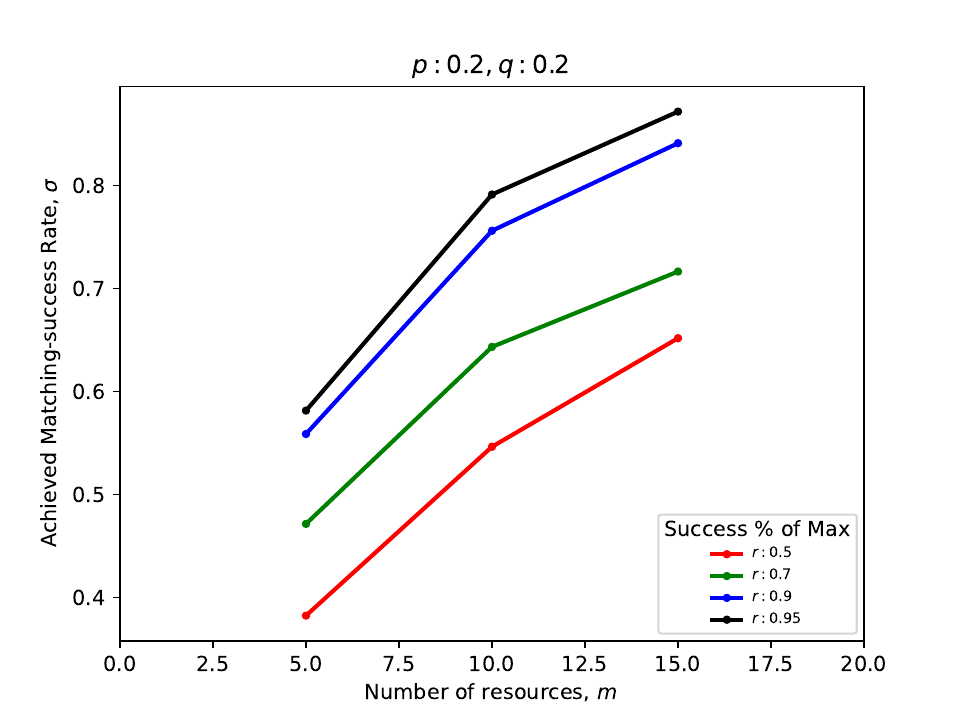} & \plt{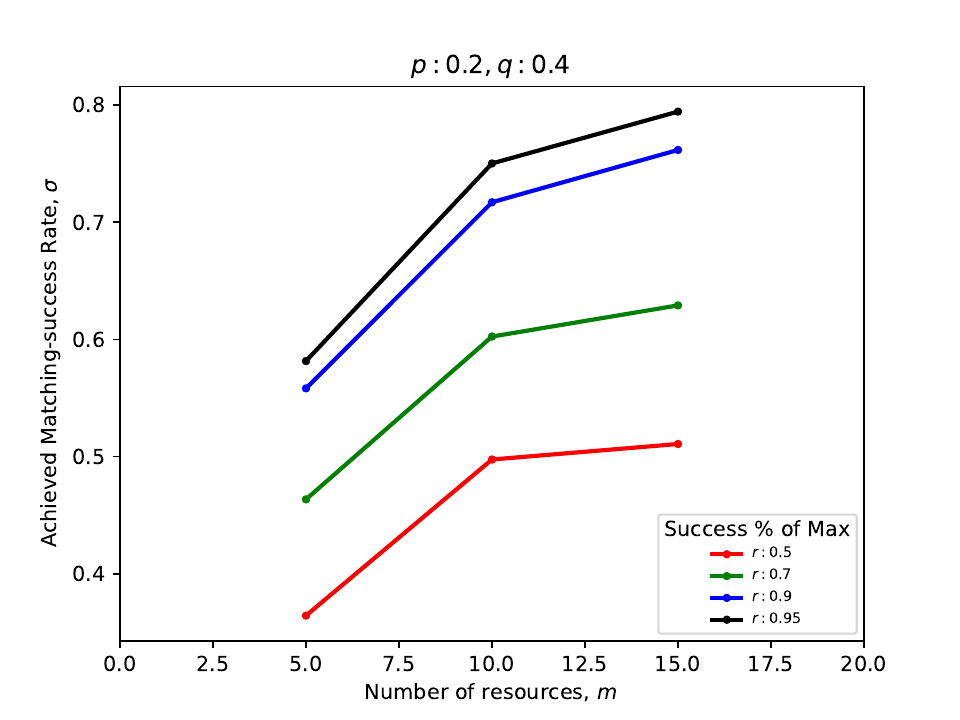} &
    \plt{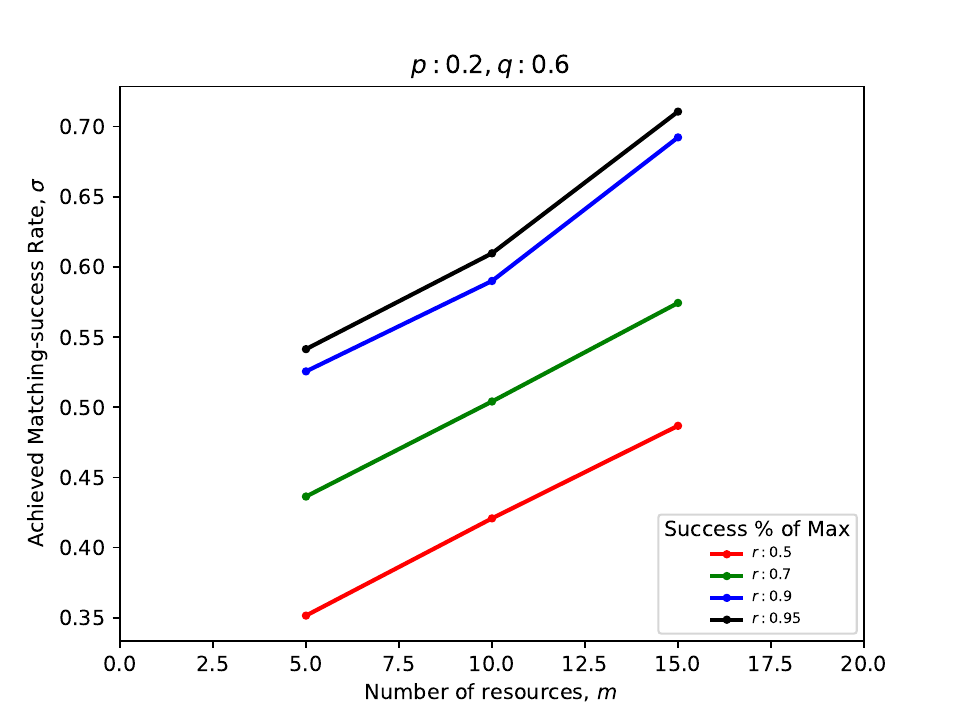} &
    \plt{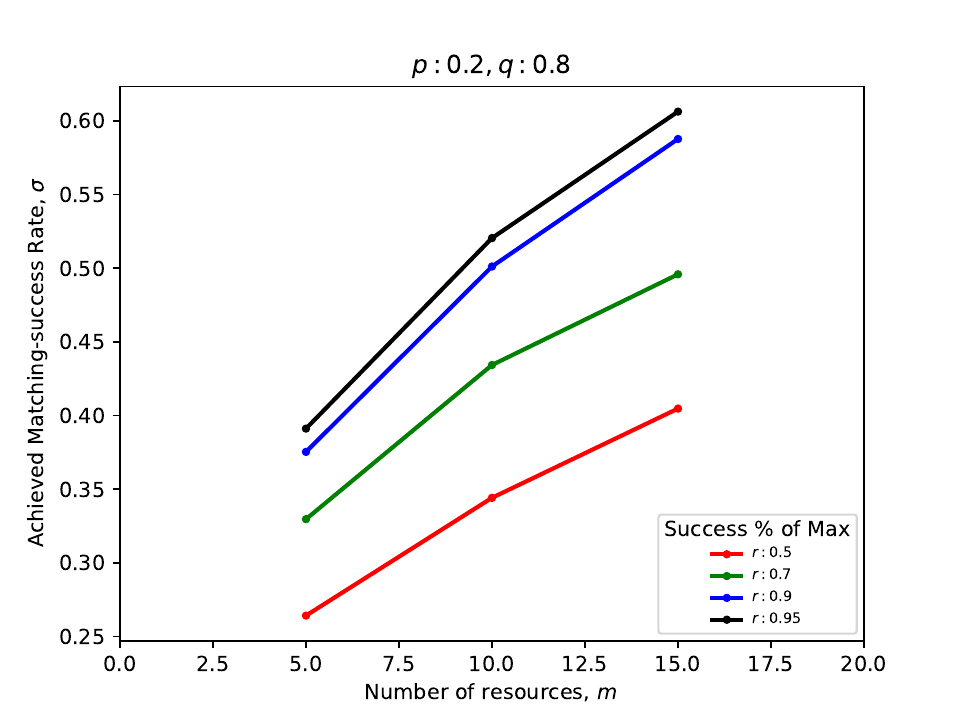} \\
    \plt{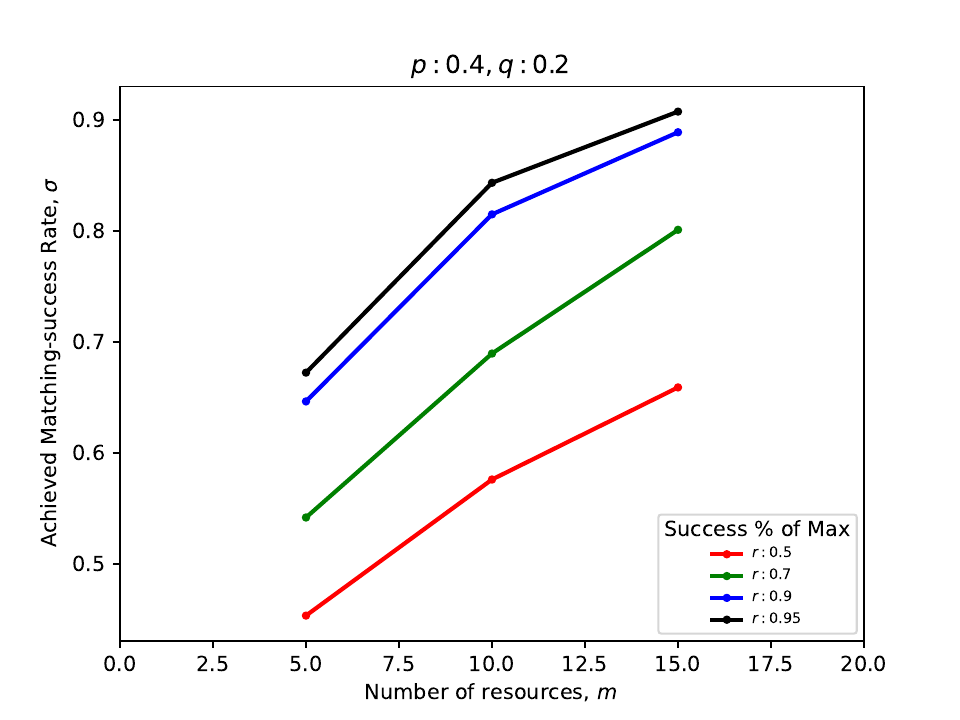} & \plt{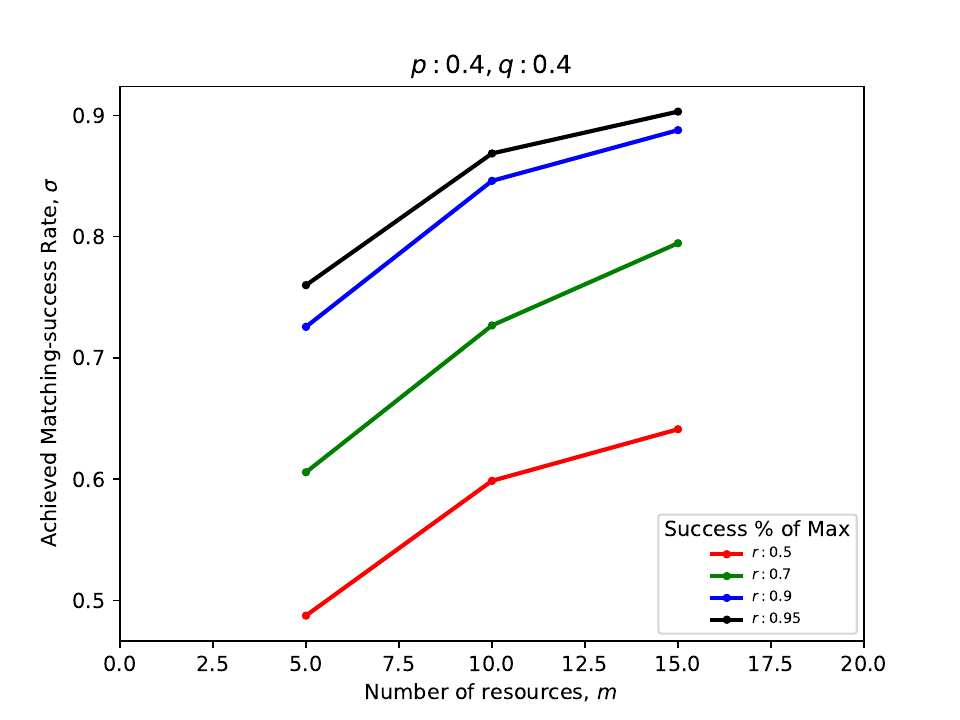} &
    \plt{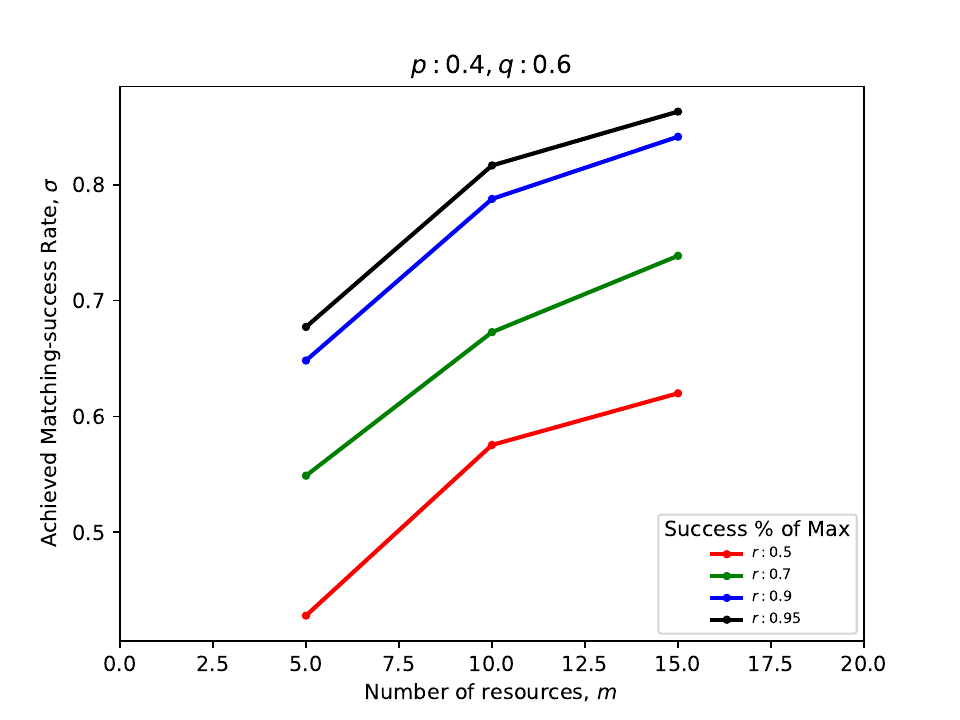} &
    \plt{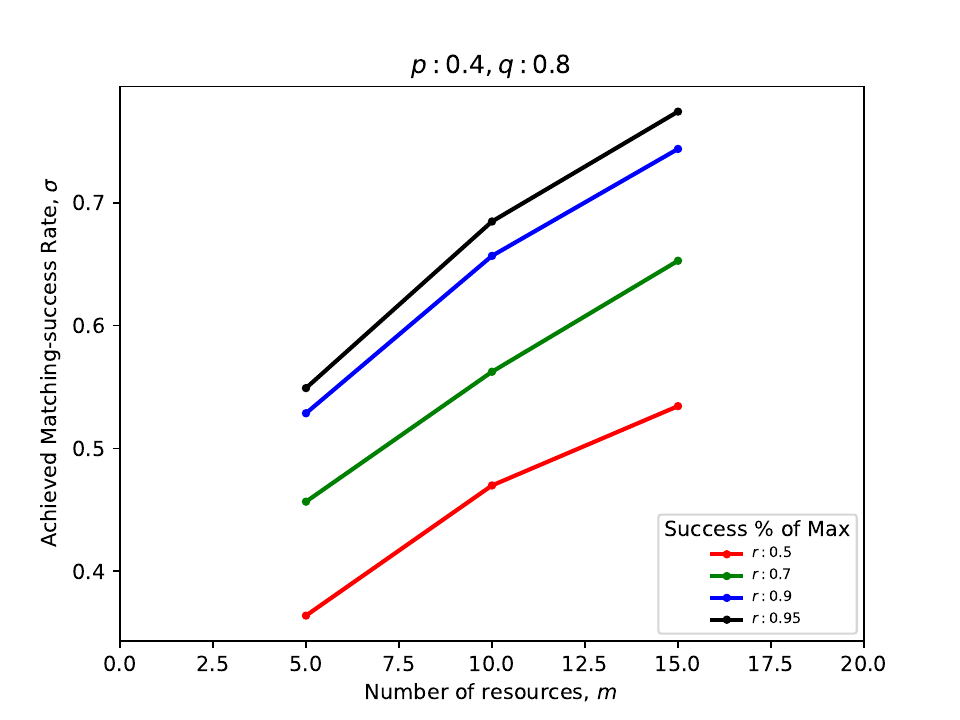} \\
    \plt{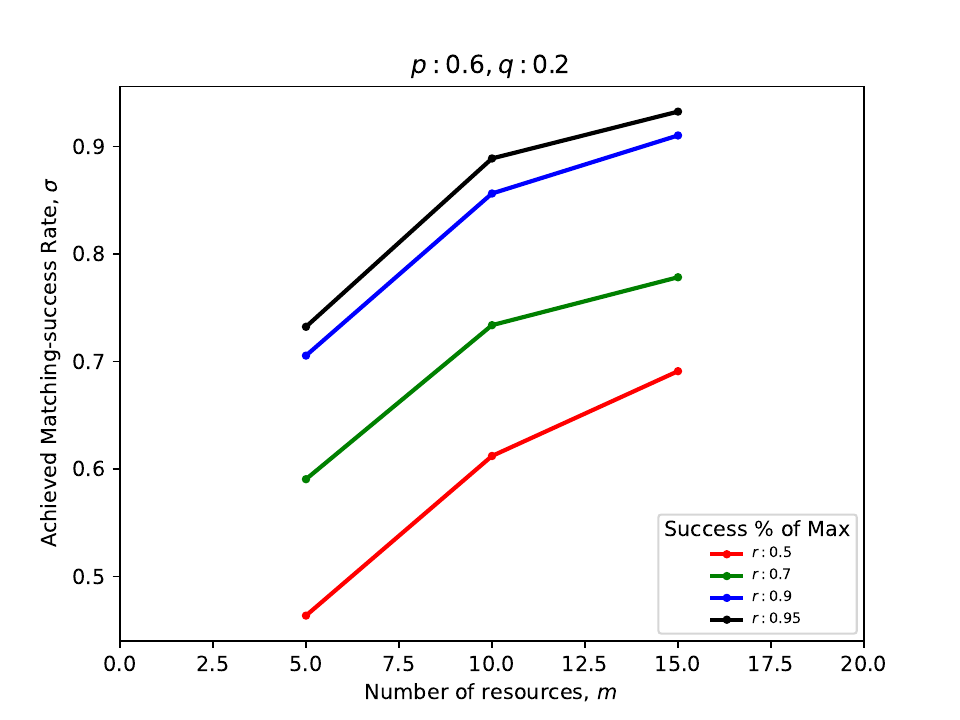} & \plt{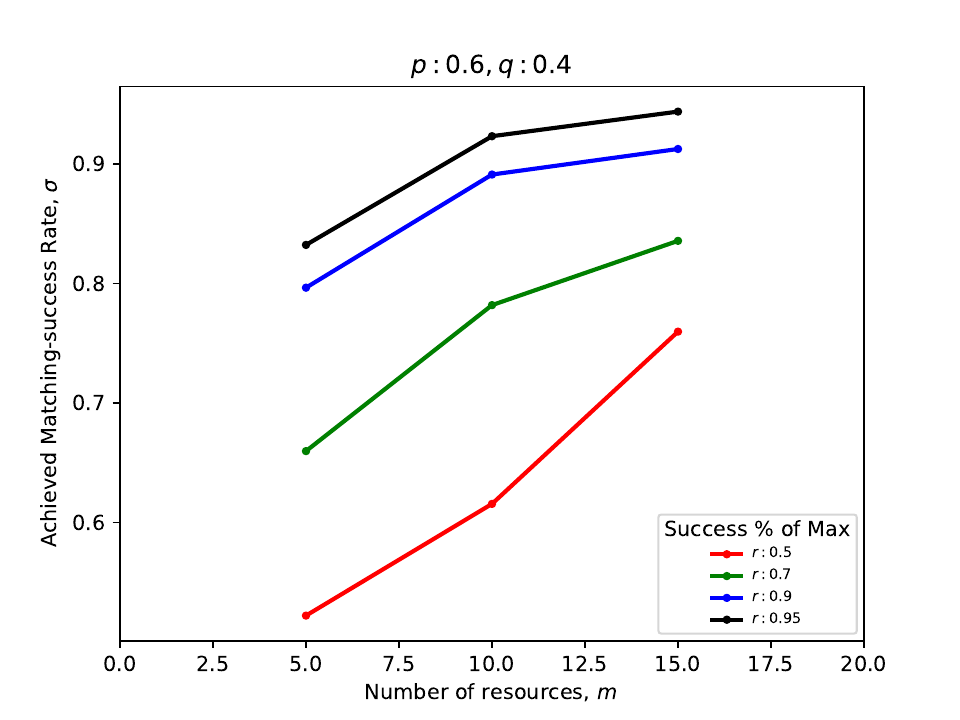} &
    \plt{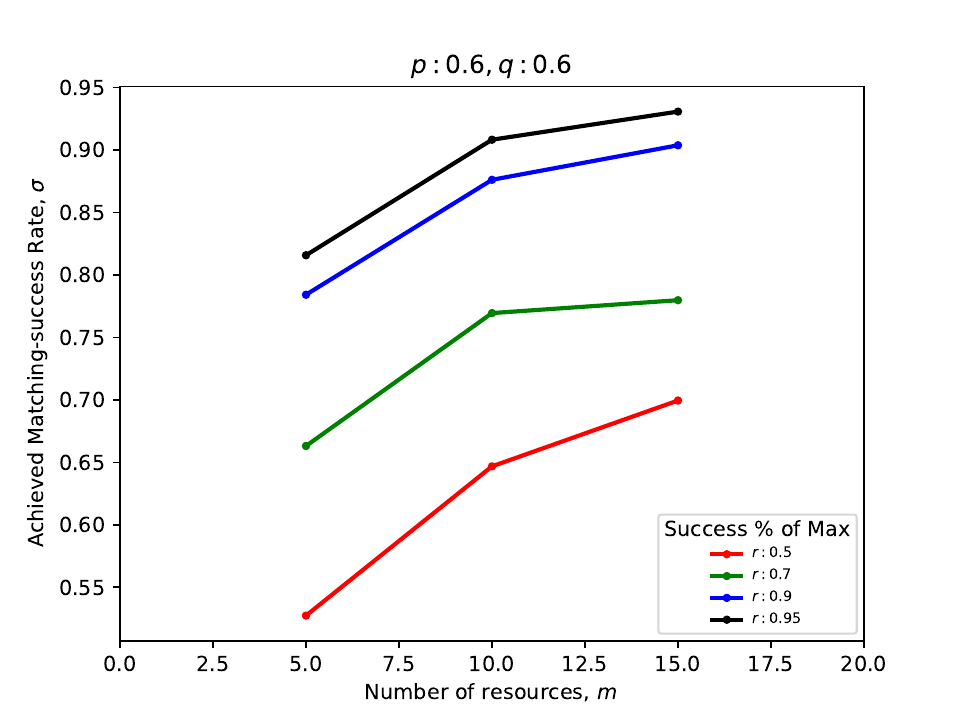} &
    \plt{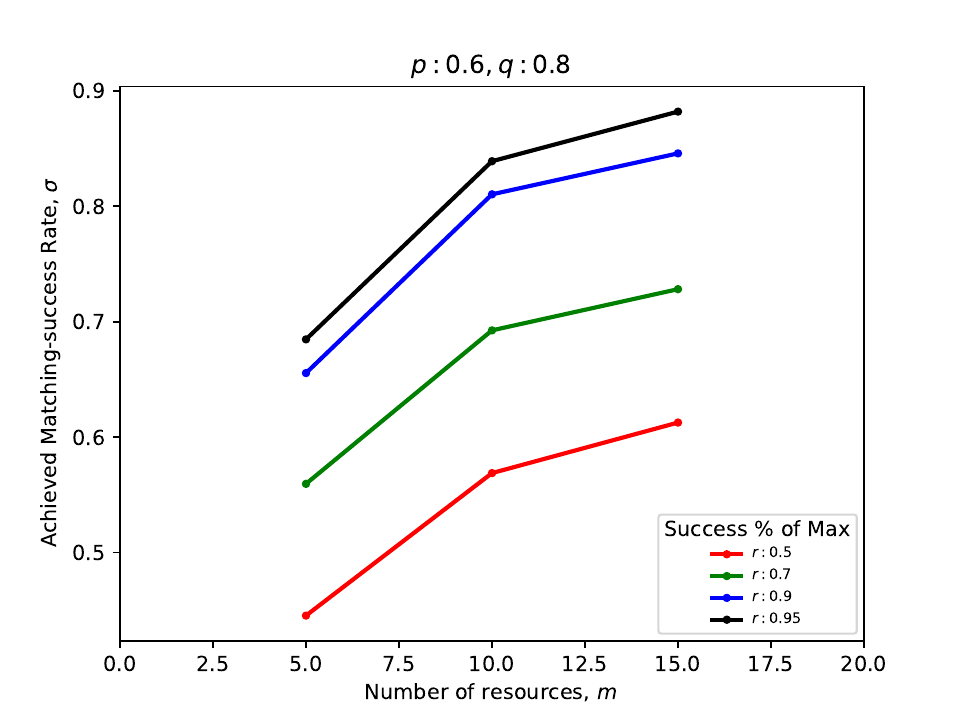} \\
    \plt{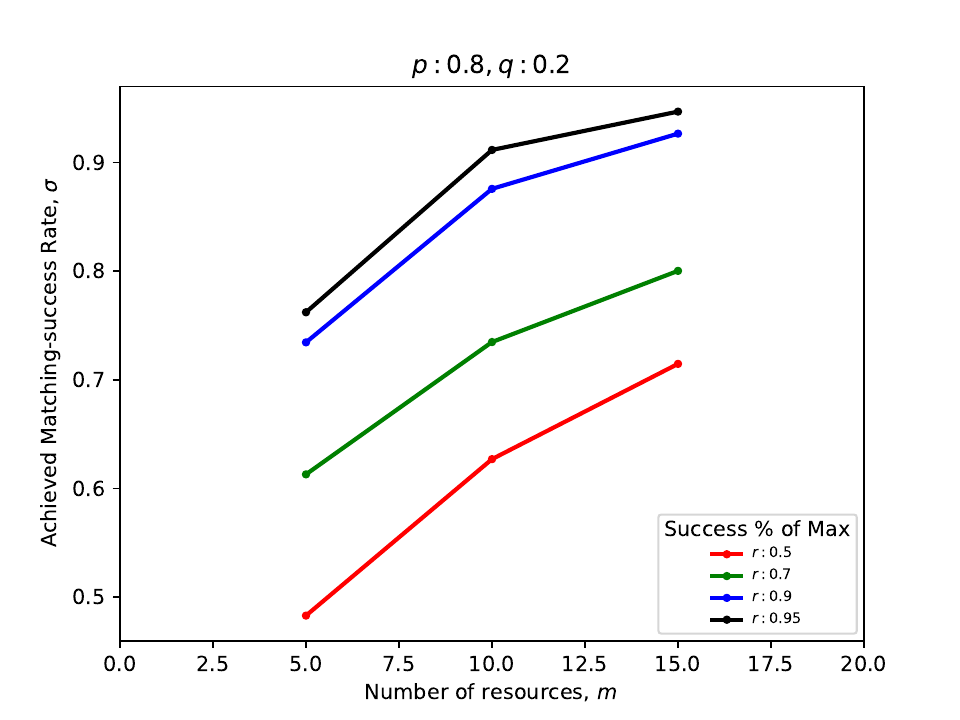} & \plt{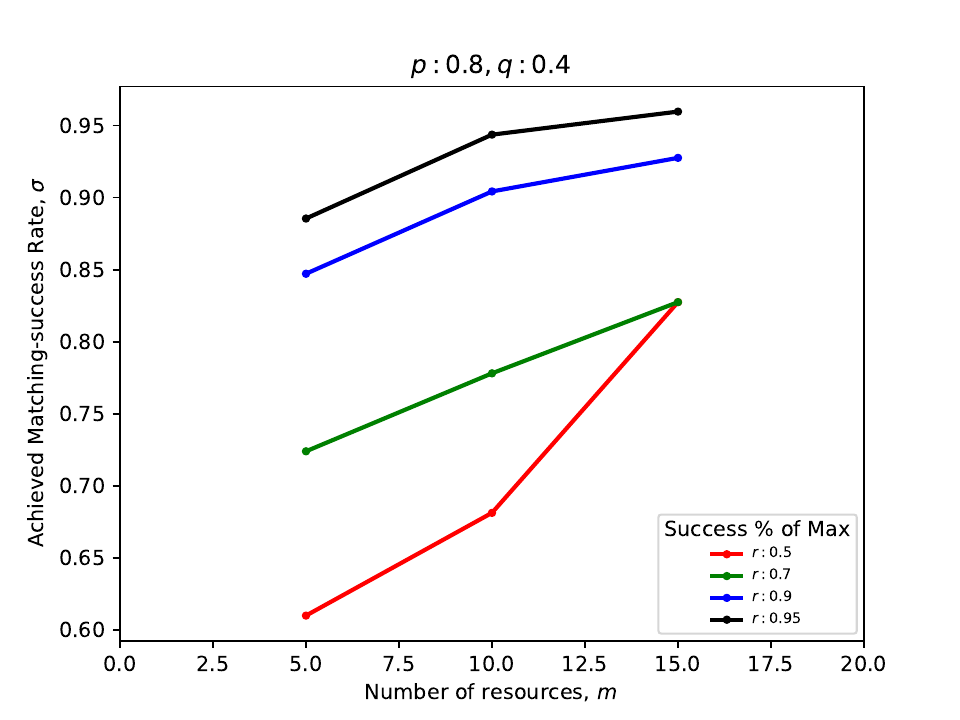} &
    \plt{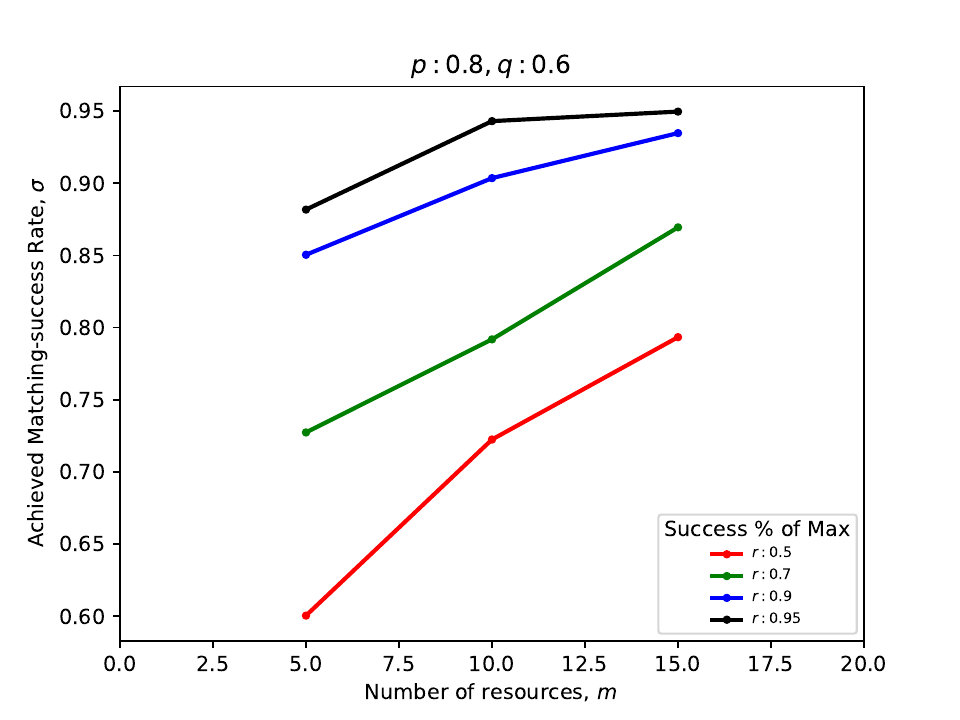} &
    \plt{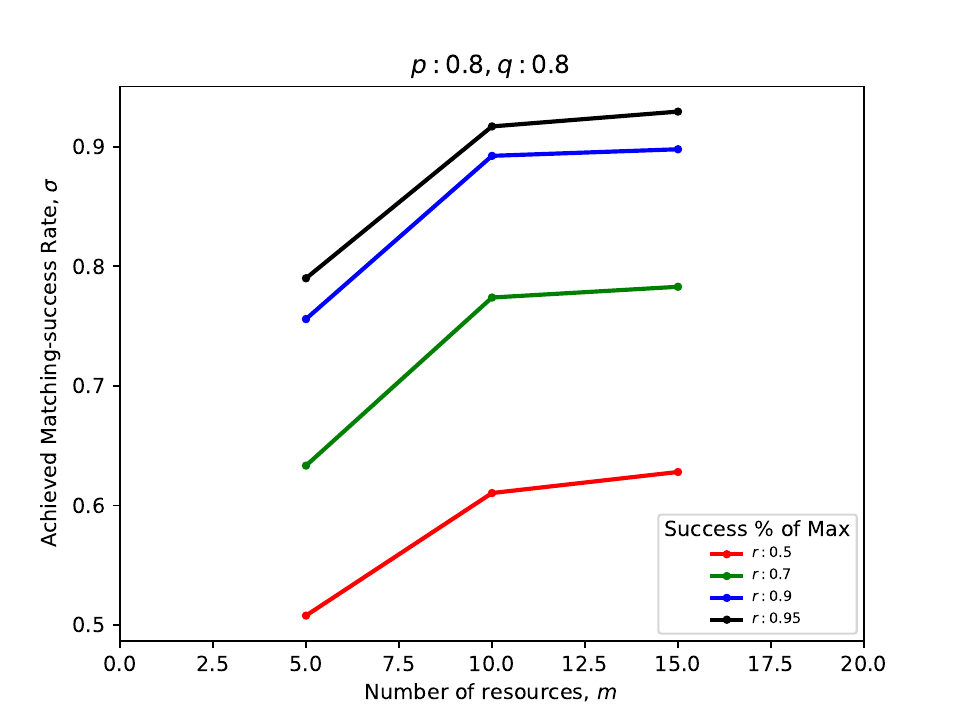} \\
\bottomrule 
\end{tabular*}
\caption{The figures show how achieved matching-success rates $\sigma$ change with number of resources $m$ when activity-graph density $p$ and resource-map graph density $q$ are varied across each figure in the array. Each figure shows several individual plots for different settings of number of resources $m$.}
\label{CV_1}
\end{figure*}

\begin{figure*}[t]\sffamily
\begin{tabular*}{\textwidth}{c@{}c@{}c@{}c@{}}
\toprule
    \plt{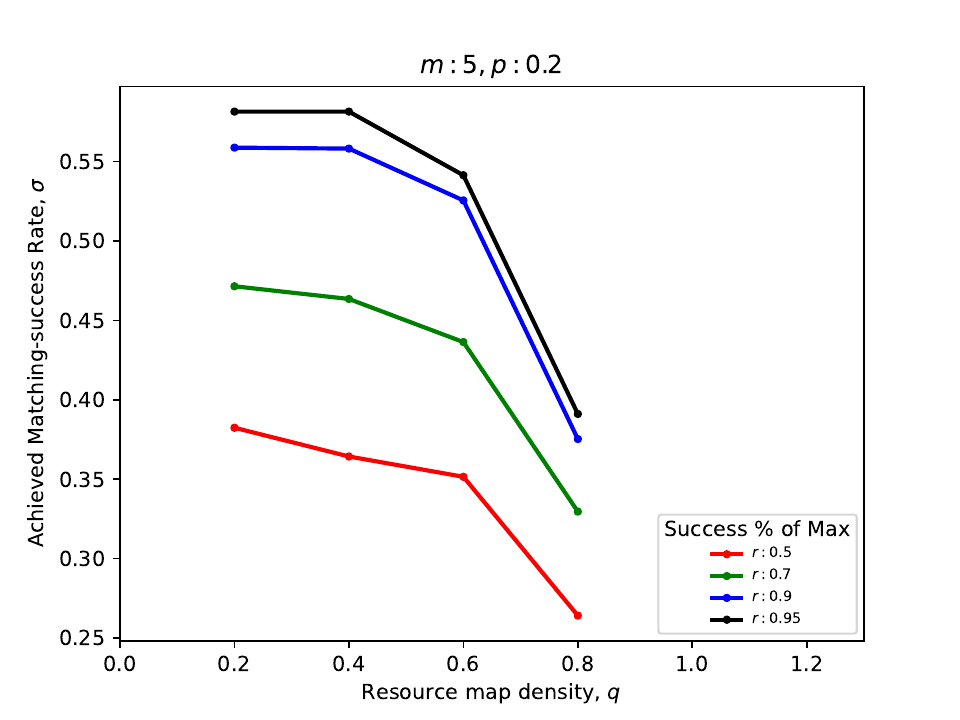} & \plt{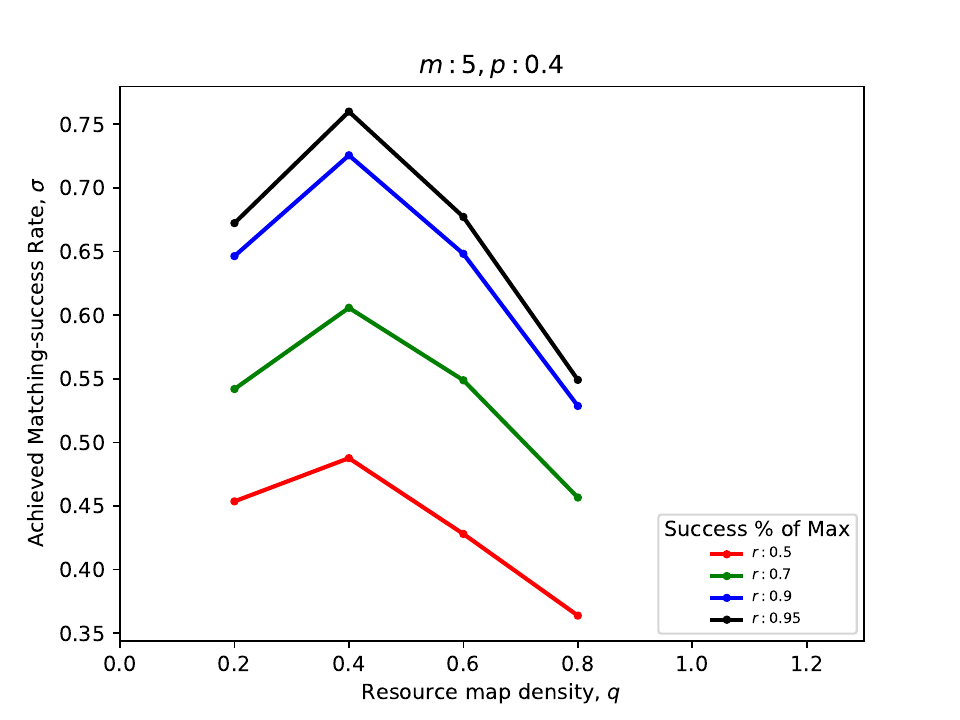} &
    \plt{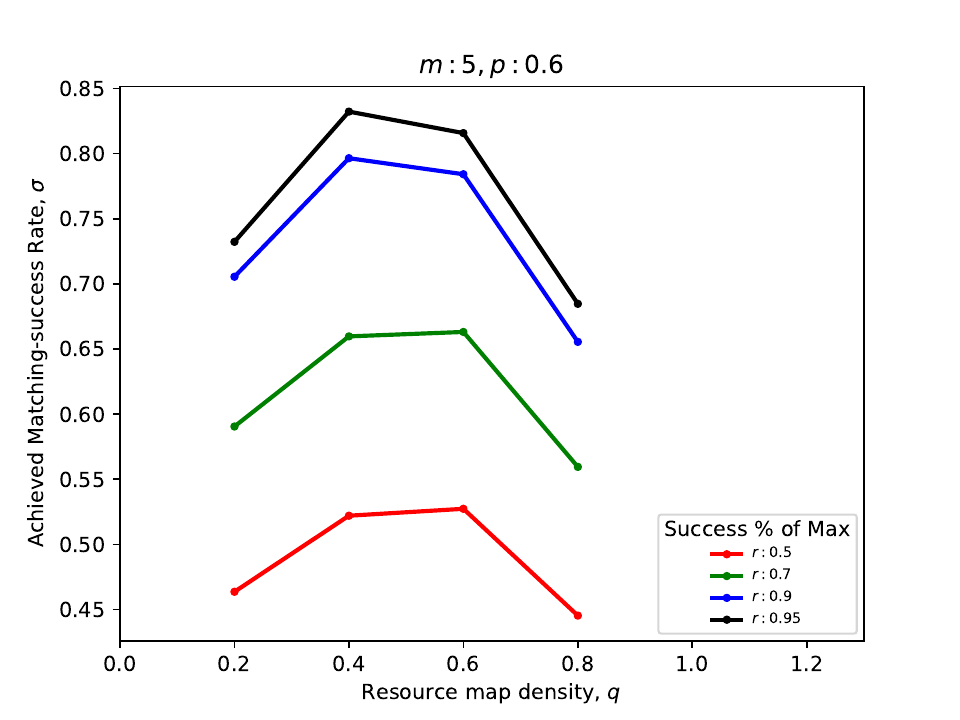} &
    \plt{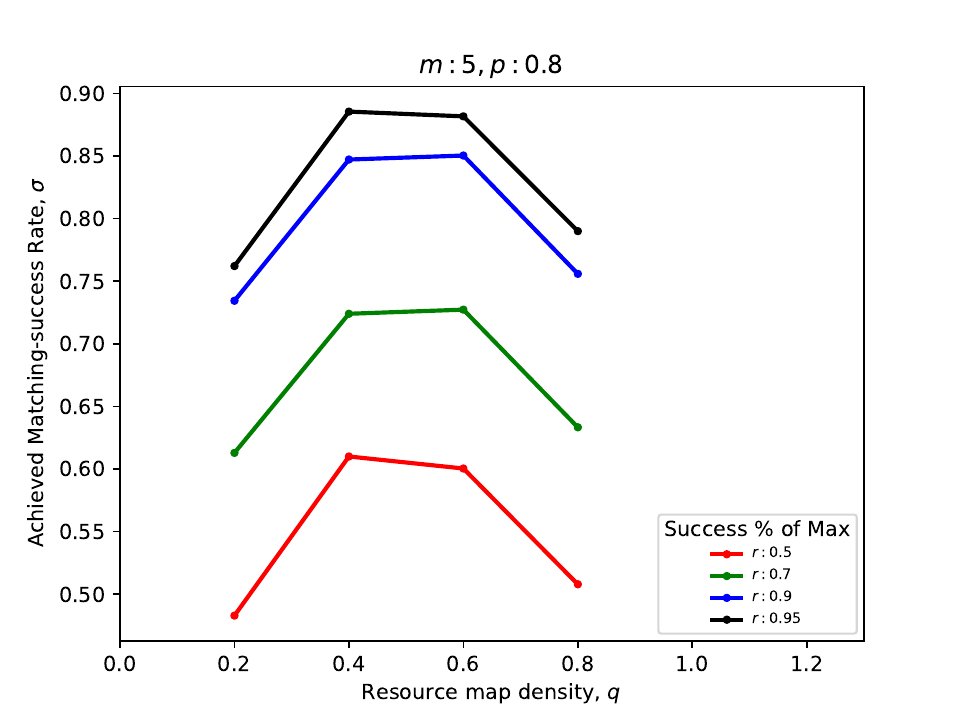} \\
    \plt{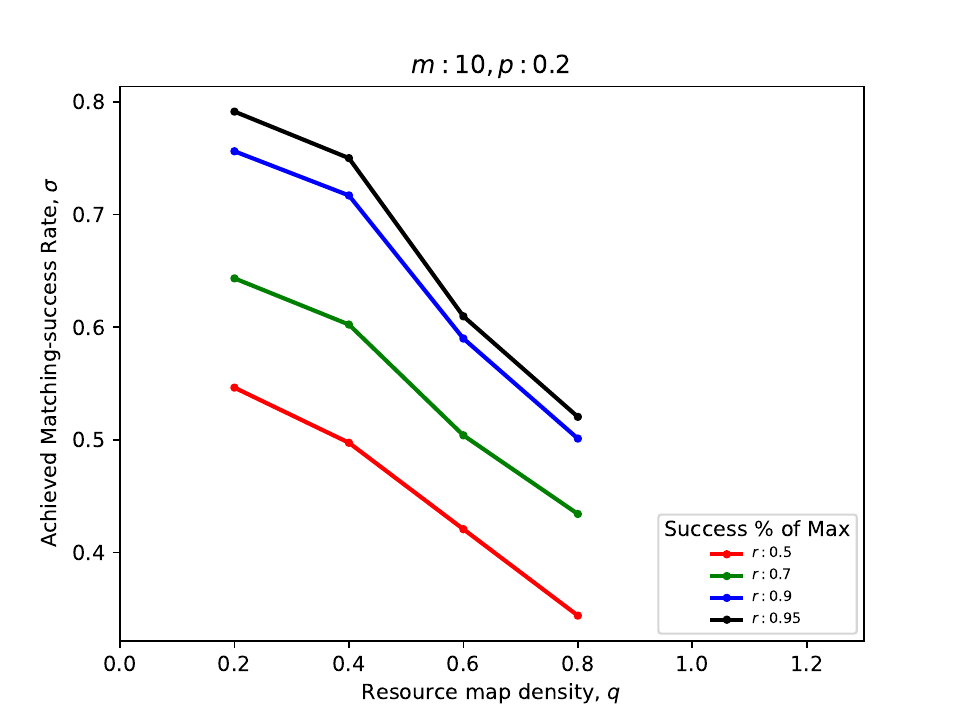} & \plt{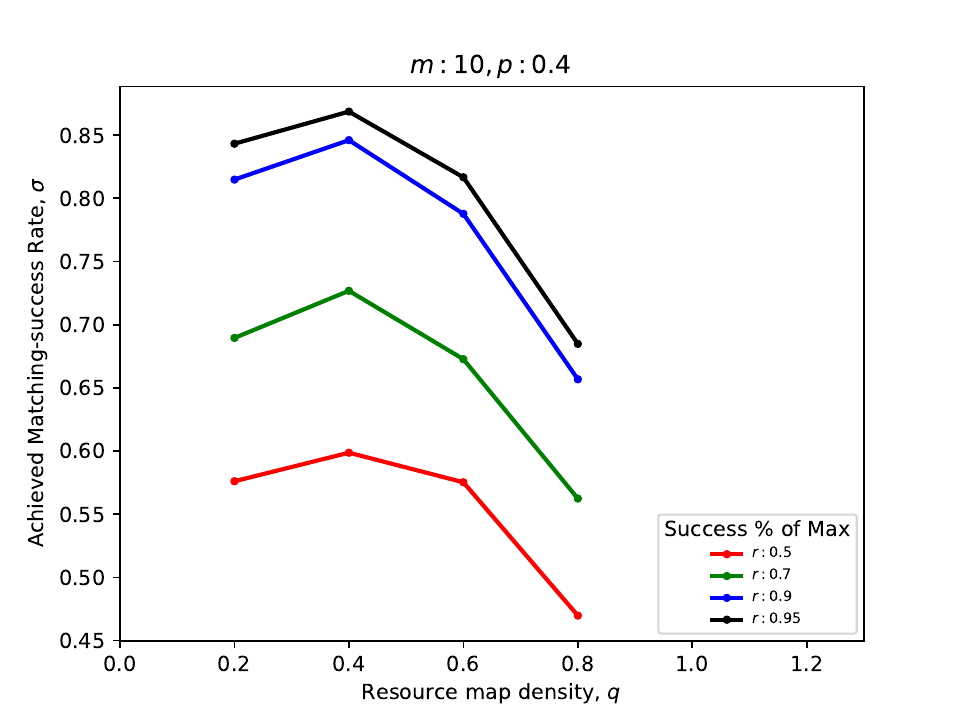} &
    \plt{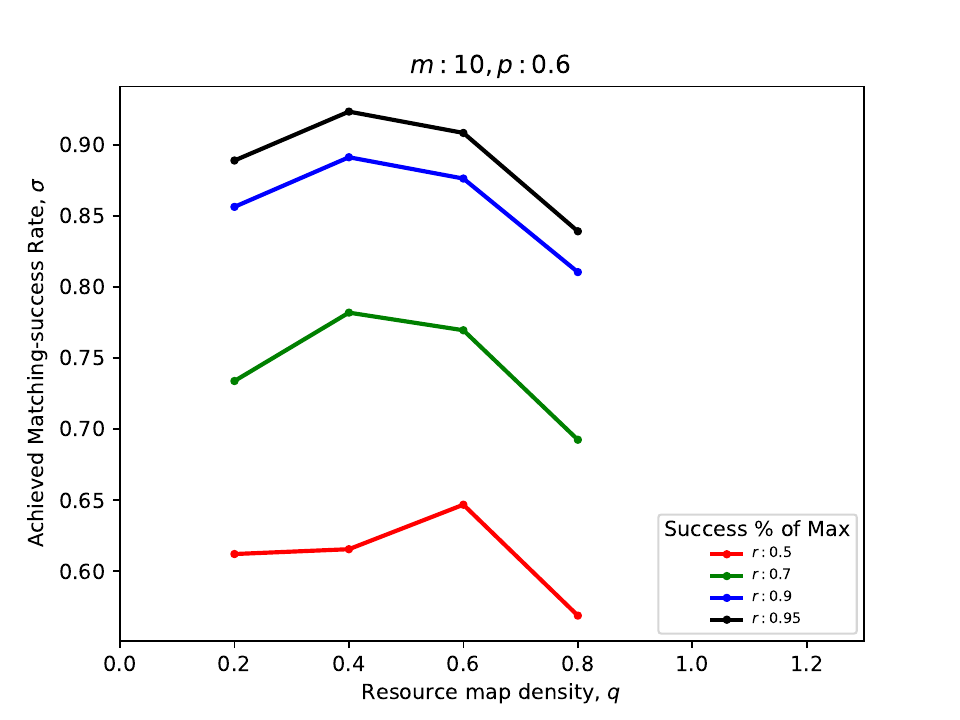} &
    \plt{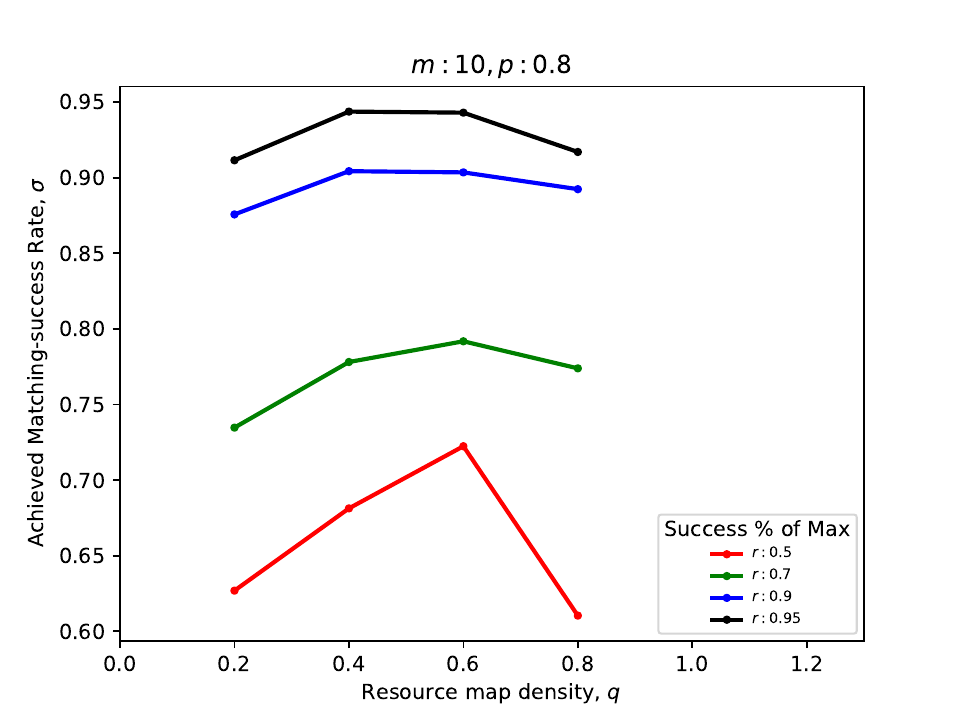} \\
    \plt{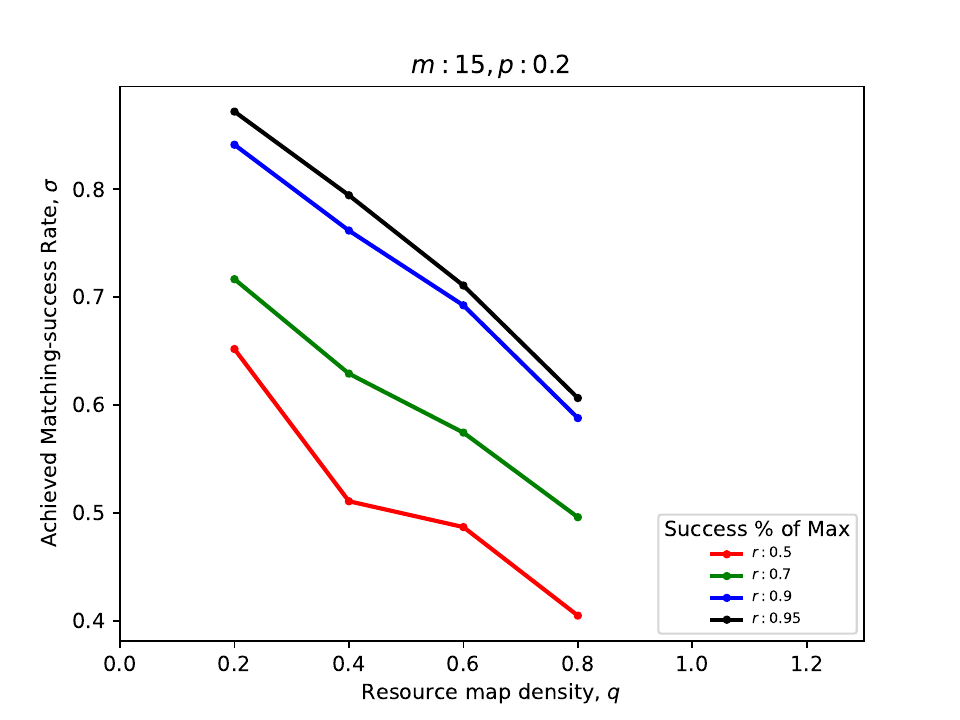} & \plt{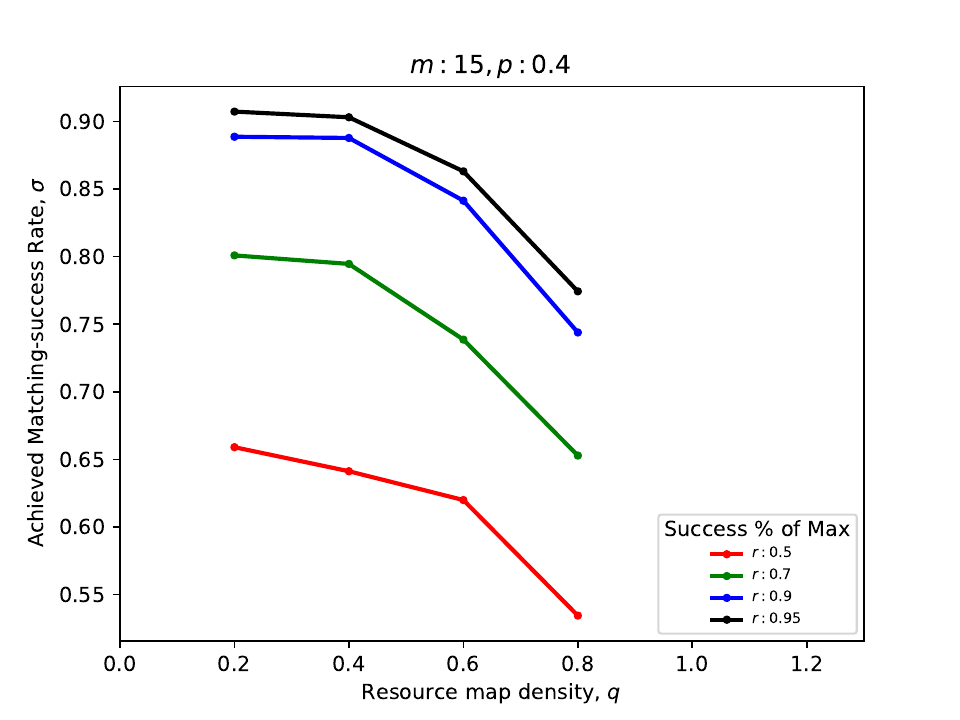} &
    \plt{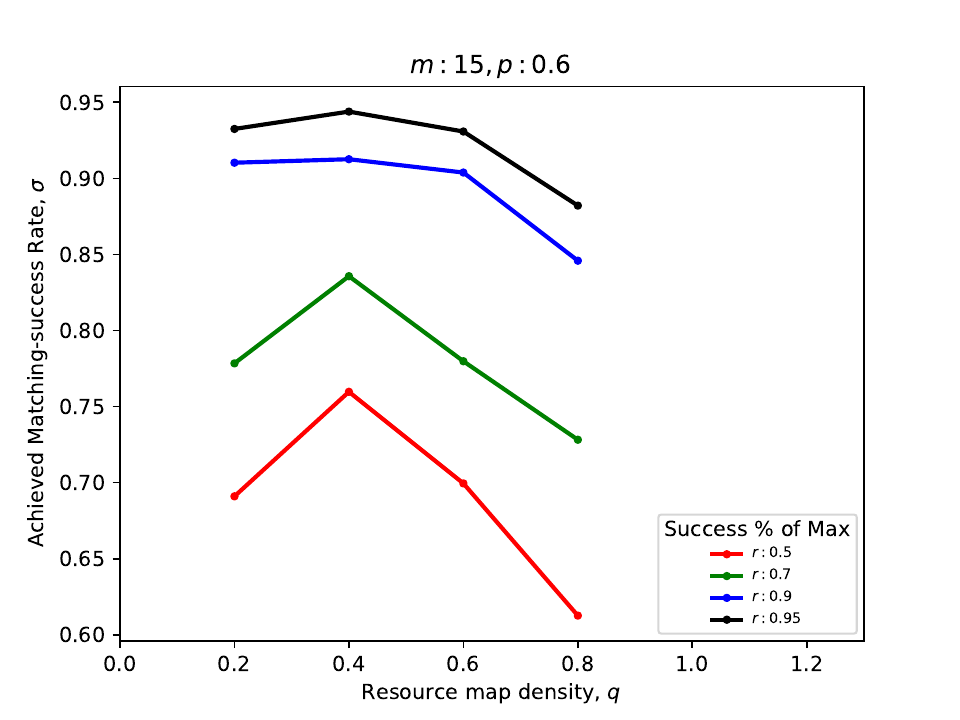} &
    \plt{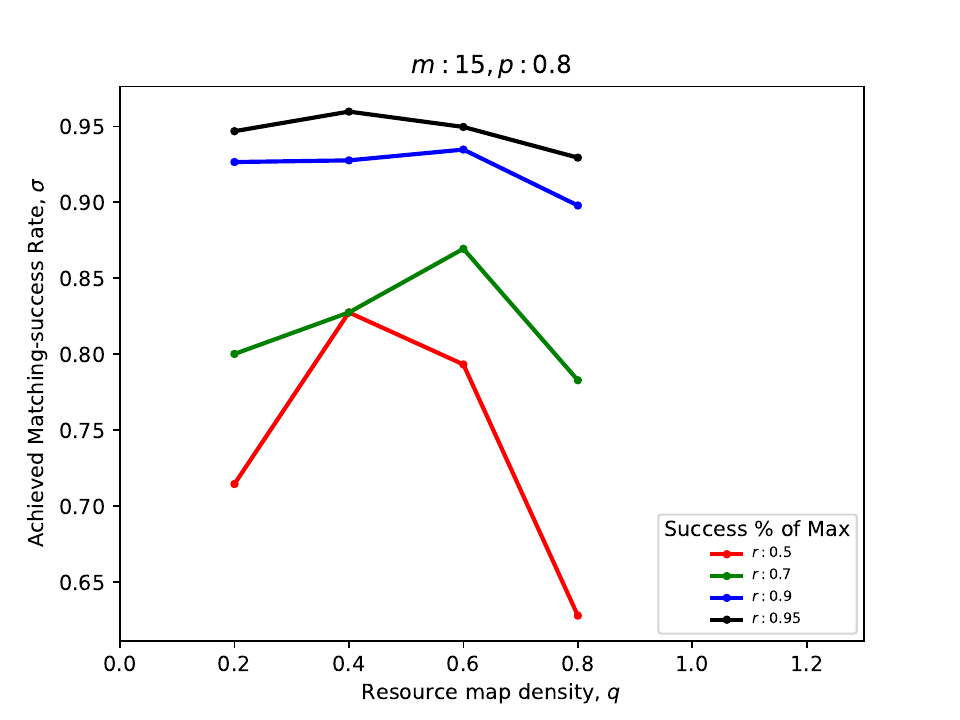} \\
\bottomrule 
\end{tabular*}
\caption{The figures show how achieved matching-success rates $\sigma$ change with resource-map density $q$ when the number of resources $m$ and activity-graph density $p$ are varied across each figure in the array. Each figure shows several individual plots for different settings of resource-map graph densities $q$.}
\label{CV_2}
\end{figure*}

\begin{figure*}[t]\sffamily
\begin{tabular*}{\textwidth}{c@{}c@{}c@{}c@{}}
\toprule
    \plt{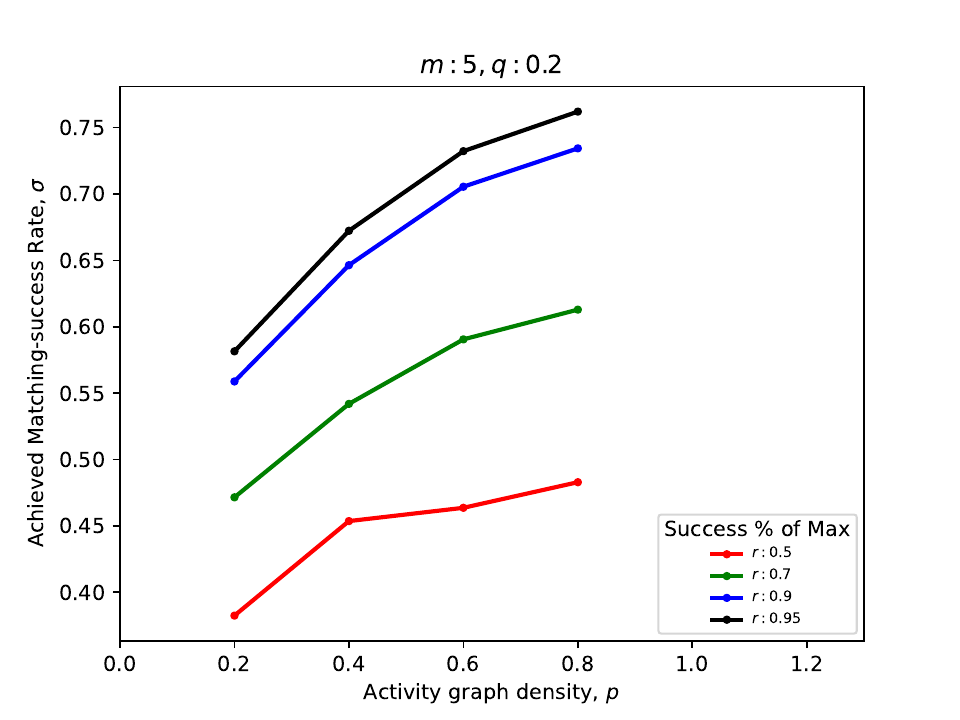} & \plt{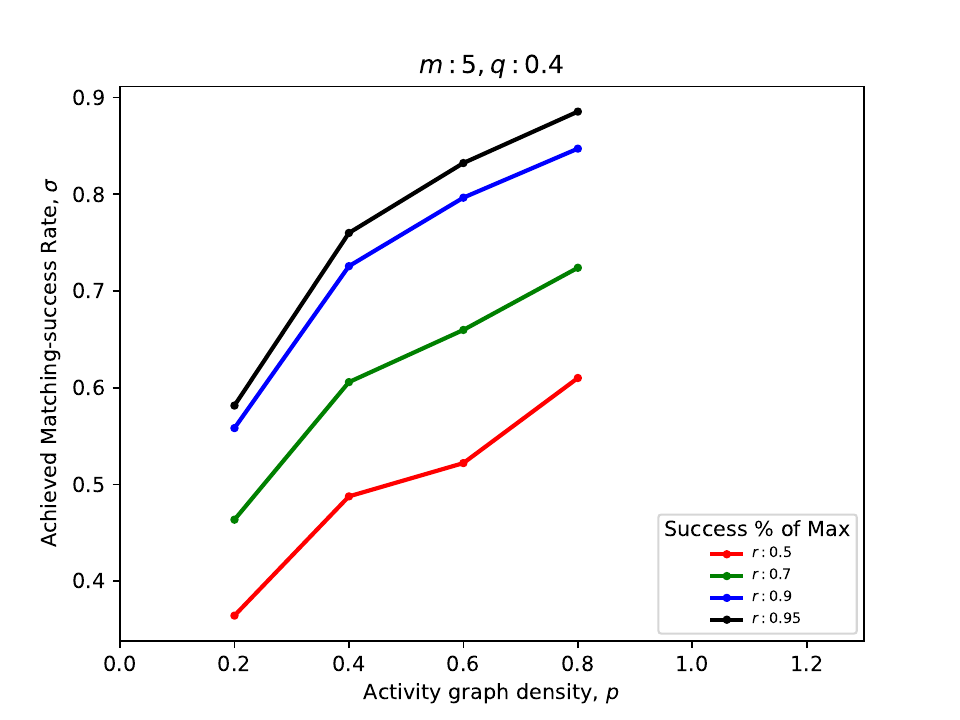} &
    \plt{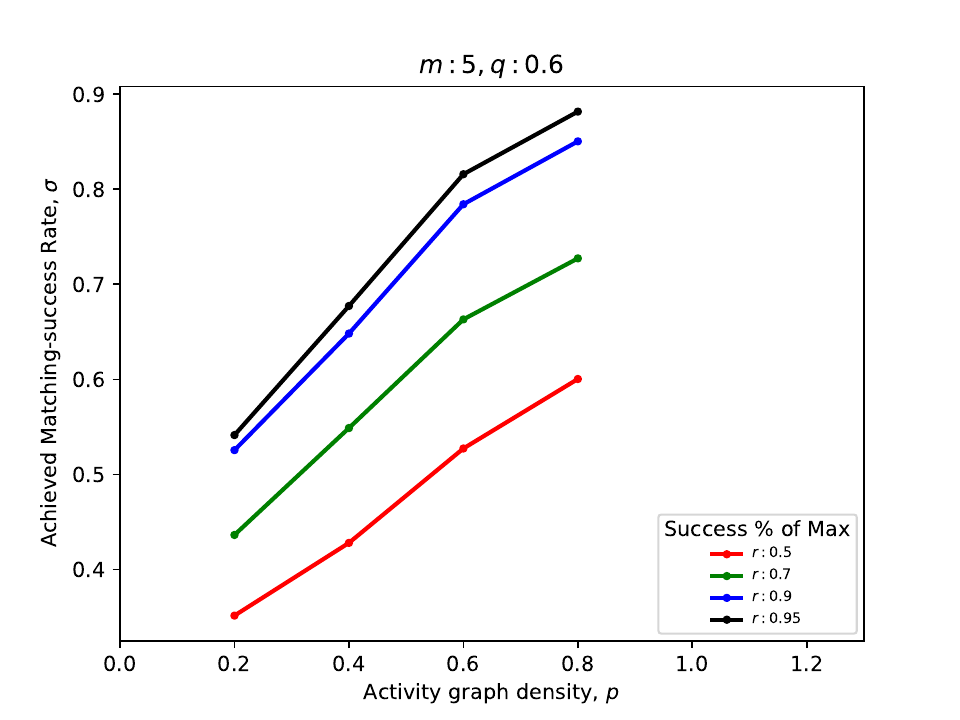} &
    \plt{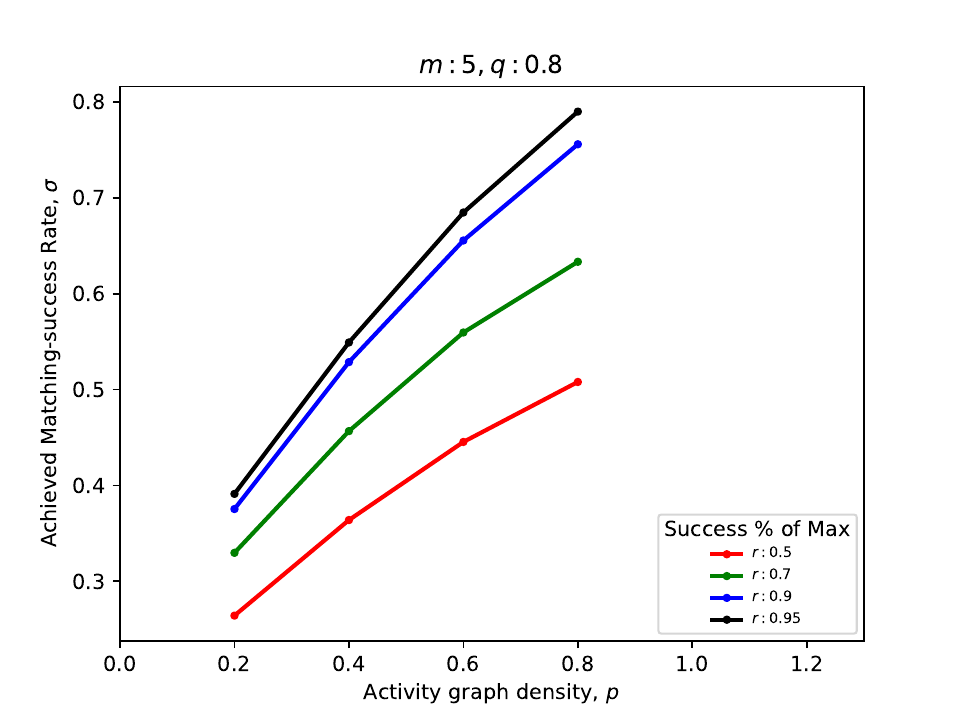} \\
    \plt{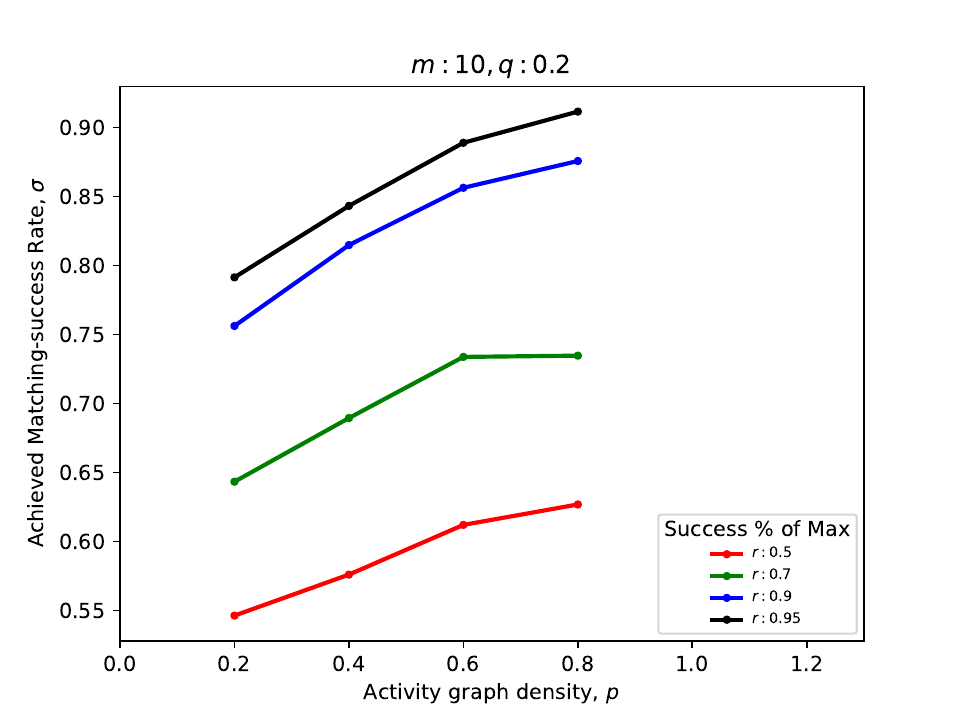} & \plt{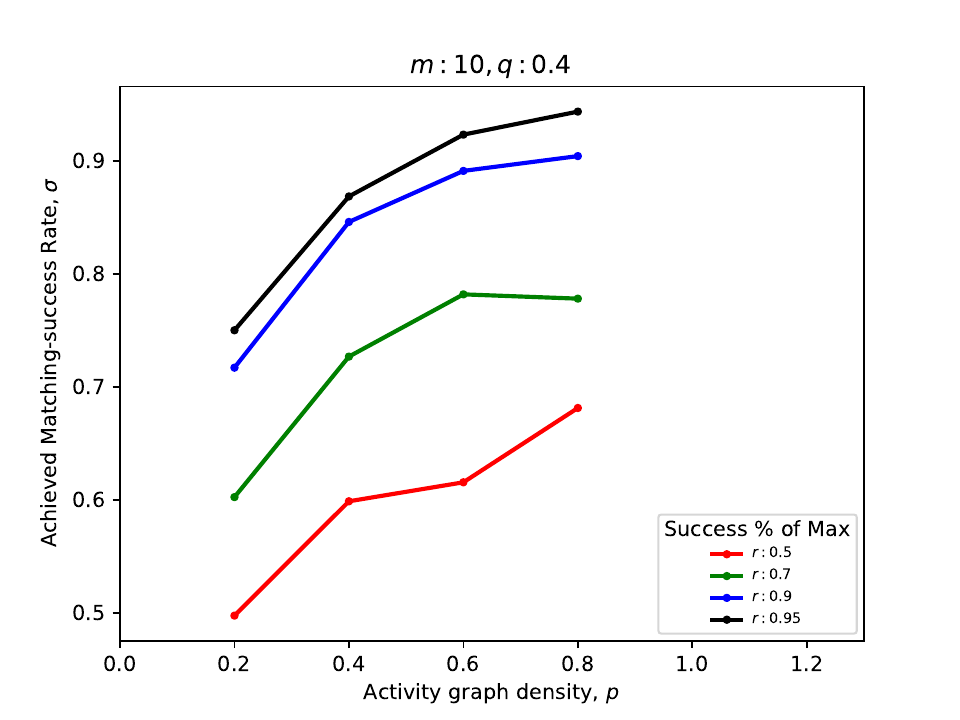} &
    \plt{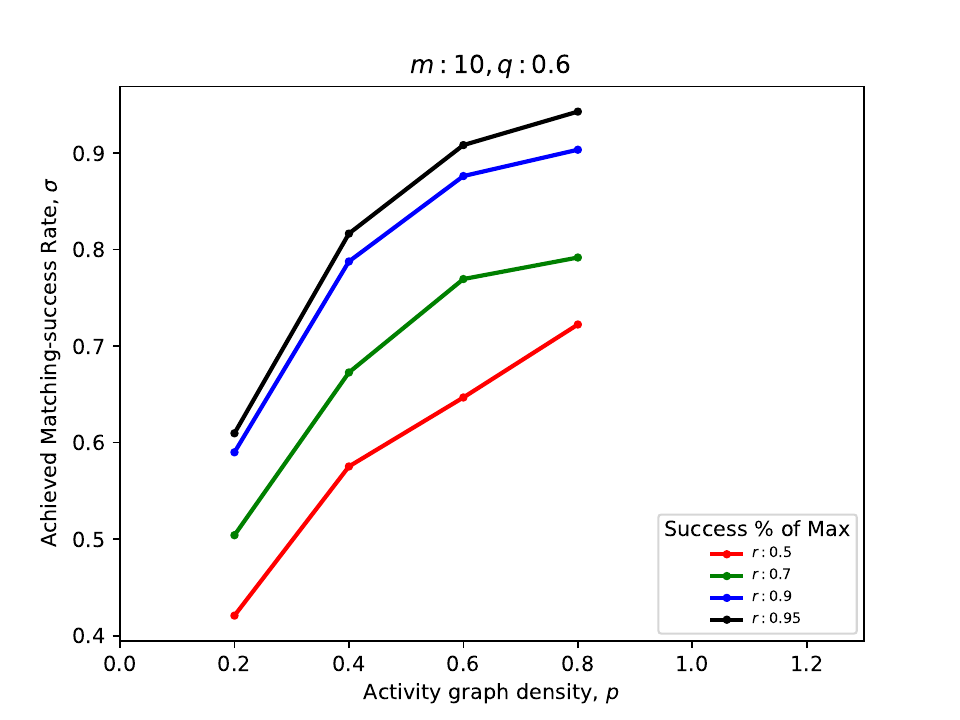} &
    \plt{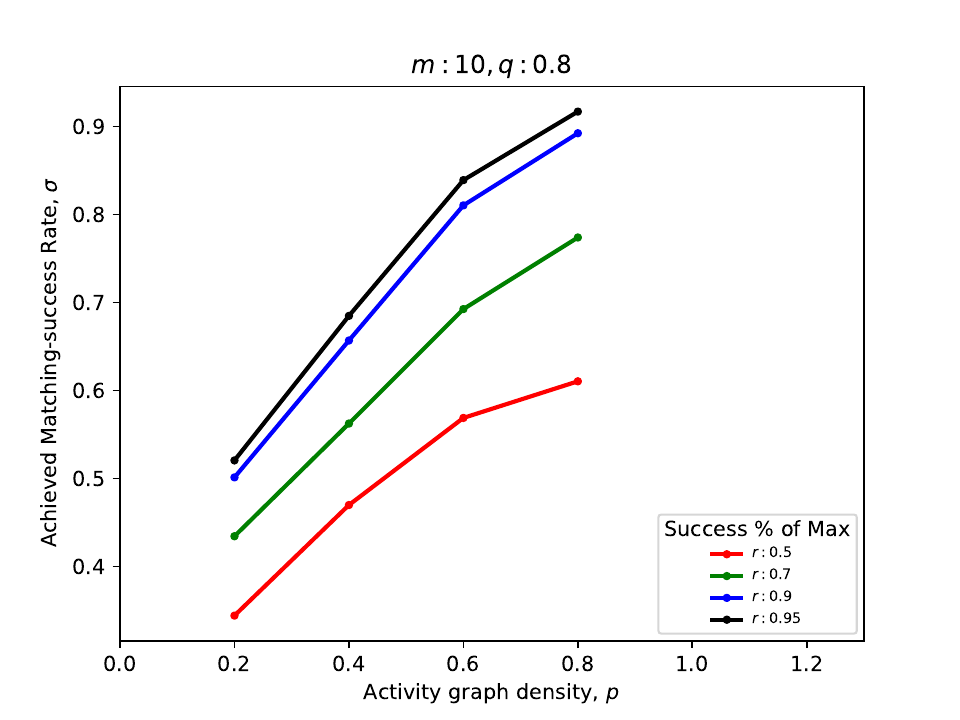} \\
    \plt{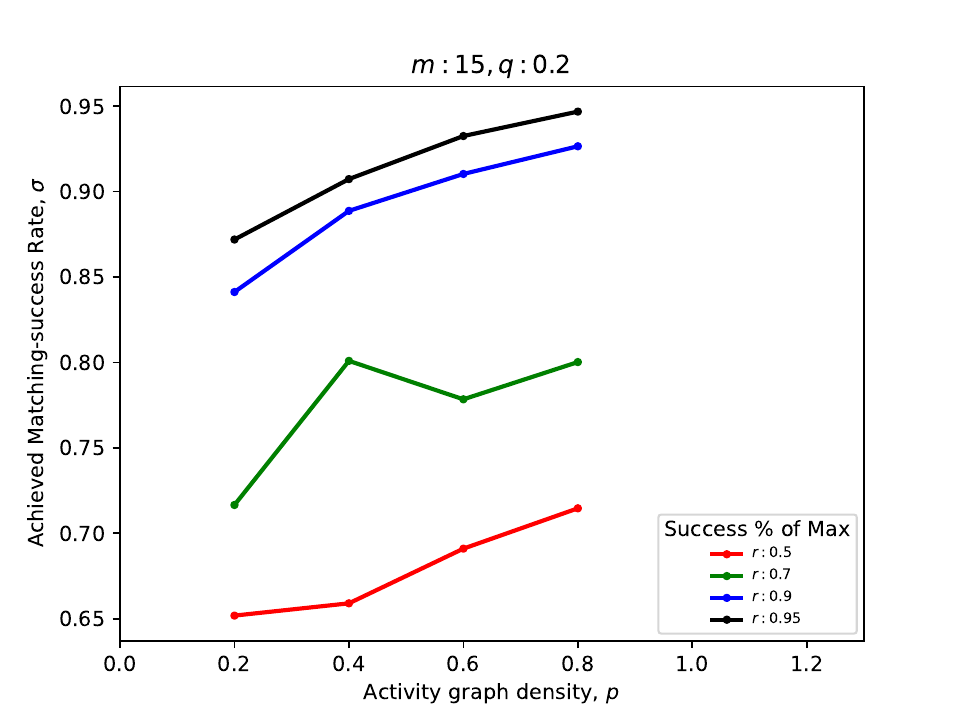} & \plt{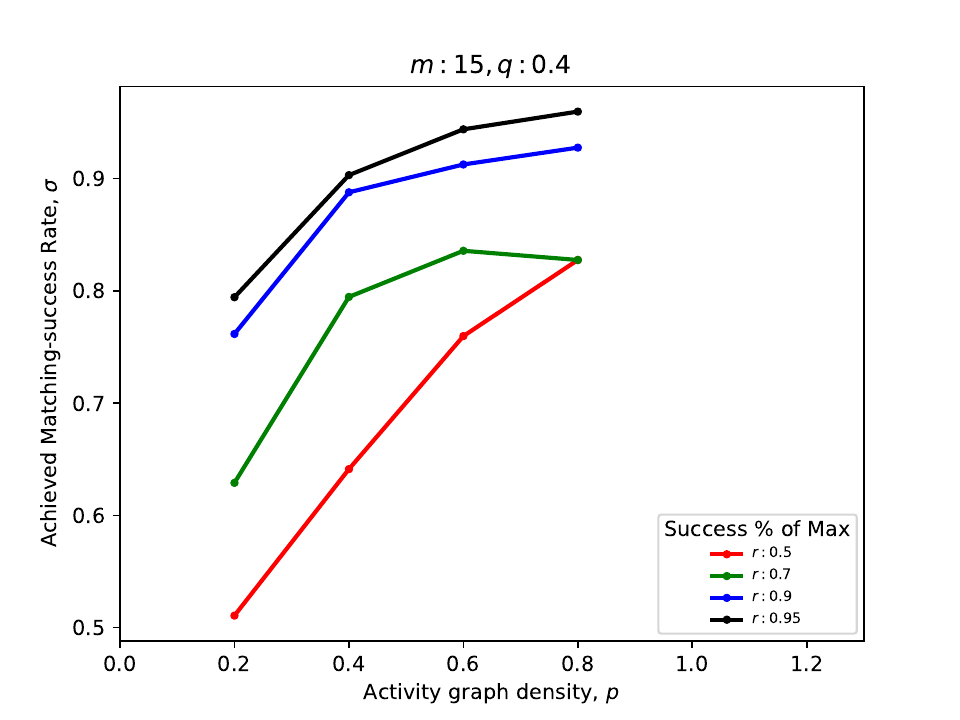} &
    \plt{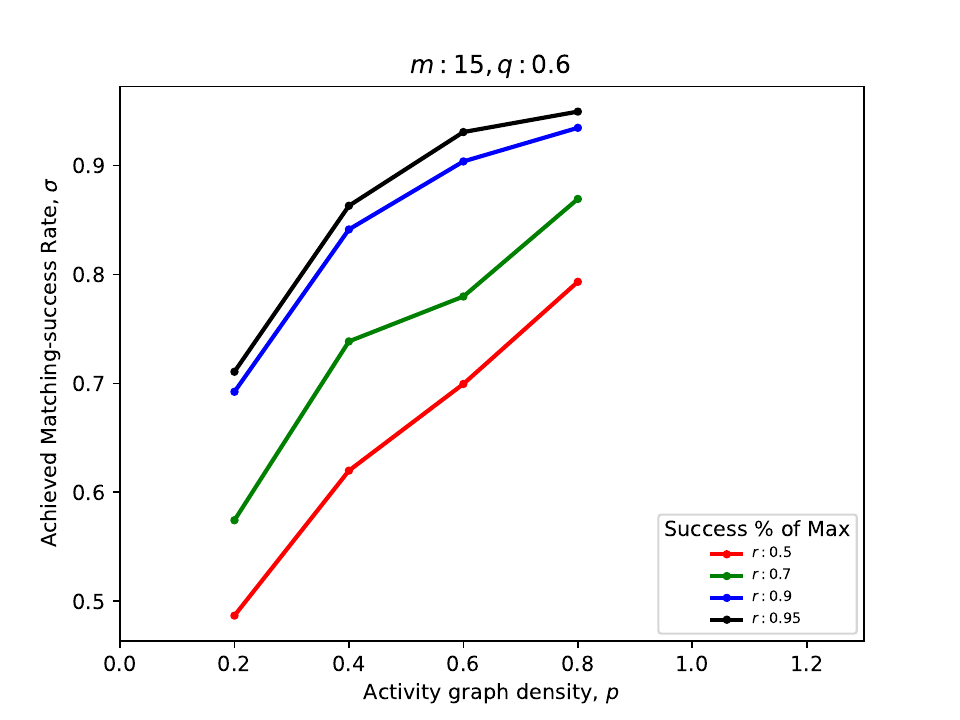} &
    \plt{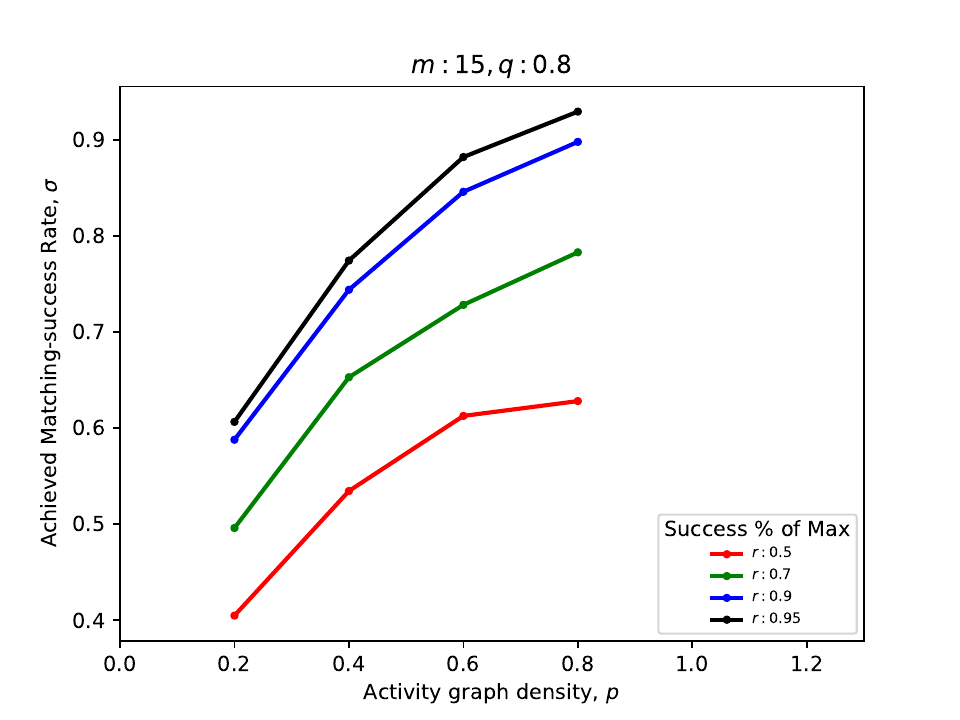} \\
\bottomrule 
\end{tabular*}
\caption{The figures show how achieved matching-success rates $\sigma$ change with activity-graph density $p$ when the number of resources $m$ and resource-map graph density $q$ are varied across each figure in the array. Each figure shows several individual plots for different settings of activity-graph densities $p$.}
\label{CV_3}
\end{figure*}

\section{Process Model with Resource Observation}
\label{sec:procmodel}

The manufacturing of a product (including R\&D, building manufacturing infrastructure, hiring skilled labor, etc) typically follows a process flow chart that is either explicitly defined upfront, or it may emerge during the manufacturing. As an example, see \cite{sherman} for a description of the car manufacturing.

For our purposes we build a process model that consists of the following components: a set of sub-processes (which we call activities) that require certain resources to complete and that can only be started after some other activities have been completed. To be more precise, our process model $$P = (A, R)$$ where $A = \{a_1, \cdots, a_n \}$ is a set of $n$ activities, and  $R = \{R_1, \cdots, R_m \}$ denotes a set of $m$ available resources with $R_i \in \mathcal{N}$ the available amount of the $i$-th resource. Each activity is a tuple 
$$a_j = ( p_j, r_j  ,t^{\min}_j, t^{exp}_j, t^{\max}_j,  )$$ 
with the following interpretation:
\begin{itemize}
    \item 
    $p_j  \subset A$ denotes the set of parent activities for activity $a_j$, i.e. activities that must have been completed before starting activity $a_j$. 
    \item 
    $r_j = (r_j^1, \cdots, r_j^m) $ is the vector of required resource counts for activity $a_j$ with each $r_j^l \le R_l$ (i.e., less than the total number of units available of resource $l$). A value of zero, which is frequently the case, implies that that activity $a_j$ does not require resource $l$
    \item 
    $t^{\min}_j, t^{exp}_j, t^{\max}_j$ are the minimum, average, and maximum time durations (in hours or days) that activity $a_j$ should take to complete. Our model assumes that these three values are the parameters of a $\beta$-distribution of the actual activity duration.   
\end{itemize}

The concept of parent activities directly leads to a dependency graph of activities and the resources required for specific activities directly lead to a mapping of activity to resources required. In order to be properly defined, the dependency graph must of course by cycle-free, i.e. and there must exist at least one activity that has no parent activities. 

When actually executing this process model, the key idea is that an activity starts as soon as all its parent activities have completed, it then waits until all its required resources are available, claims these resources and works to completion for a duration of time drawn from $\beta$ distribution of its time parameters. Thus, activities can execute in parallel. The process model execution is complete once all activities have been finished. To be a bit more precise, we describe the main steps:
\begin{enumerate}
\item 
Start at simulation time $T=0$. 
\item 
While there are still activities that have not yet been completed:
\begin{enumerate}
    \item 
    Check for each incomplete activity $a_j$ whether all its parent activities $p_j$ have completed
    \item 
    If that is the case for $a_j$, then check if the required resources $r_j$ are available.
    \item
    If that is the case for $a_j$, claim these resources, randomly select a duration time $t_j^{actual}$ drawing from a $\beta$ distribution with parameters  $t^{\min}_j, t^{exp}_j, t^{\max}_j$, and set the starting time of $a_j$ to current simulation time $T$. 
    \item 
    Increase simulation time $T$ by $\Delta$ units
    \item
    Check for each active activity $a_j$ whether its duration time has now been completed and mark $a_j$ as completed at time $T$ if this is the case and release used resources
    \item 
    Check for all activities that are awaiting availability of resources, if this is the case now, and mark those activites as active if it is the case
\end{enumerate}
\item Return current simulation time $T$ as completion time together with the start and completion times of each activity and the set of resources used at each point in time during the simulation

\end{enumerate}

Thus, the result of a process model run consists of  (i) the total time to completion, which is the predicted time to market, (ii) the start and completion times of each activity, and (iii) information as to what resources were used at each point in time. These simulation trace-style outputs will be required to build the HMM model. The time durations that are randomly chosen for each activity can of course be captured by a single random seed. These durations enable us to run the process model in a generative fashion, by changing the random seed, which will then allow us to build a statistically sound Hidden Markov Model. We call a set of such runs an \emph{ensemble}.

We have implemented this Process Model in the Python-based version of the Simian Parallel Discrete Event Simulation Package \cite{simian}, which is very well suited for our model as it natively uses processes as a primitive. The entire code-base fits into 700 lines of code, with at least 500 lines devoted to statistical analysis.

\section{Hidden Markov Models to Predict Development Status}
\label{sec:hmm}

For a process model $P=(A,R)$, let $P_i$ be the result of a run of $P$ with $i$ as random seed that determines the duration times $t_j^{actual}$ for each activity $j$. We run an ensemble $S(P) = \{ P_1, P_2, \ldots, P_s \}$ of runs, the result data of which we then use to train a Hidden Markov Model $HMM(P)$.

The components of $HMM(P)$ are the set of hidden states $H$ and a set of observable states, which we call emissions or observation states. The key concept of an HMM are that, starting in some starting state, the HMM picks a next hidden state, which corresponds to advancing time by one unit. This next state is picked with some probability (called transition probability), if a hidden states corresponds to an activity vector that takes many days to be completed, the hidden state has a high probability of picking itself as next state. Each time a hidden state is picked, it emits an observable state, which is a subset of the resources that the corresponding activity vector requires. The HMM eventually reaches a terminal state, which corresponds to all activities having been executed successfully. We describe the definition of the hidden states, observation states, emission probabilities and hidden state transition probabilities in more detail. Figure \ref{fig:hmmexample} gives a small example of a resulting HMM.

\subsection{Hidden States} 
Each hidden state has a subset of activities from $A$ as a label and defining feature with the interpretation that those activities are being executed whenever the HMM is in this hidden state. For example, consider an activity subset $\{a_1, a_2, a_3\} \subset A$. Hidden state $h_{\{a_1, a_2, a_3\}}$ denotes a state with those three activities active. For simplicity we will drop the $h$ notation and simple stick to the labels for hidden states, e.g., $\{a_1, a_2, a_3\} \in H$. In principle, there could exist a hidden state for all possible combinations of activities (e.g., $2^n$ hidden states), but in practice, we only define those hidden states that actually exist at least once in the ensemble $S(P)$. In practice, the number of hidden states is at most quadratic in the number $n$ of activities.

We calculate transition probabilities by counting (and then normalizing) the number of times that a transition from hidden state $h$ (corresponding to some activity vector) to another hidden state $h'$ (corresponding to a different activity vector) happens in the simulation traces from the ensemble runs $S(P)$; in order to do this, we impose a time duration (typically set to one day, but could be set as small as one second) after which the hidden state transition is modeled to occur. The transition probabilities $Prob(h,h')$ of moving from state $h$ to state $h'$ form a triangular matrix if we sort the hidden state state labels according to a partial order as imposed by the process model dependency graph.

\subsection{Emission States}
The emission or observation states $O$ are combinations of resources. To be more precise, for a hidden state $h \in H$, we look at the resource vector $r_j$ of each activity $a_j \in h$; the meta observation state $O'$ corresponding to hidden state $h$, is a $(0,1)$ vector of length $m$, the $l$-th element (for the $l$-th resource is set to one if there exists at least one activity  $a_j \in h$ which requires a non-zero amount of the $l$-th resource, e.g., $r^l_j > 0$. We choose to turn the resource vector into a binary form for simplicity. Thus, the meta observation vector $O'$ is a cumulative set of all resources required by the activities in the corresponding activity vector of hidden state $h$.

In reality, there is uncertainty on whether a resource required for a certain activity can actually be observed, for instance Firm B might be actively hiding a resource, or Firm A's observation skills may just miss it. Thus, let us define $p_j$ as the probability that any resource that is required for activity $a_j$ is actually observed in a single observation round (we define later how we set $p_j$ in our experiments. Now looking at meta observation vector $O'$ for hidden state $h$, we obtain a set of potentially multiple probabilities for each resource in $O'$; if there are  multiple cases, we just pick the maximum probability for the resource, resulting in a vector $O''$ of length $m$ (total number of resources) of probabilities. We call $O''$ the observation probability meta vector associated with hidden state $h$.

Now we are finally ready to define the set of observation states and the corresponding emission probabilities from hidden states: for observation probability meta vector 
$O''$ of hidden state $h$, we form an emission state for each possible combination of resources that occur in $O''$ with non-zero probability. Thus, if $O''$ has $m'' < m$ non-zero probability entries, each element of the $2^{m''}$ sized power set becomes an emission or observation state $o$ that is connected to hidden state $h$ with an emission probability that is calculated as follows. 

First, let $O'''$ be a vector that contains only the non-zero entries of $O''$. Let us explain the emission calculation by way of example: Let $O''' = (p_1, p_2, p_3)$ consists of three non-zero probabilities, w.l.o.g. assume these correspond to three resources $(r_1,r_2,r_3)$.  We know that by construction, all corresponding resources are indeed required for the activity vector of hidden state $h$ to be executed. Thus, the probability of observing all three $Prob(r_1, r_2, r_3) = p_1*p_2*p_3$. The probability of observing only say $r_1$ and $r_2$ is $Prob(r_1, r_2) = p_1*p_2*(1-p_3)$, probability of observing only $r_1$ is $Prob(r_1) = p_1*(1-p_2)*(1-p_3)$, the probability of observing none of the three resources is $Prob(\{\}) = (1-p_1)*(1-p_2)*(1-p_3)$. All those emission probabilities add up to one.

Figure \ref{fig:hmmexample} shows a small example of how the HMM is built. The hidden states correspond to activity vectors. Once all activities have been completed, the HMM jumps into the terminal state, denoted by an empty vector.

\subsection{Using the HMM to Predict Time to Market}
Now that we have defined the transition and emission matrix of our Hidden Markov Model, we briefly remind ourselves what an HMM can be used for in our context of predicting time to market.

Given a sequence of consecutive observations (of resources used), the \emph{Forward Algorithm} calculates the probability that the HMM would have generated this observation sequence. As a significant extension, the \emph{Viterbi Algorithm} gives the most likely sequence of hidden states that would have led to an input observation sequence. Thus, a combination of the Viterbi and forward algorithm on an observation sequence, gives us the most likely sequence of hidden states as well as the probability of having produced that observation sequence. 

In our context of predicting time to market in a process model, the observation sequence of resources gets mapped into the precedence relationship graph of the activities by the HMM. In other words, the HMM tells us in what part of the activity graph, Firm B most likely is given our observations. Let $A_{HMM} \subseteq A$ be the set of activities of the final hidden state $h$, e.g., the activity label of state $h$, as returned by the Viterbi Algorithm. A simple backward search lets us compute the set of activities $A'_{HMM}$, which we recursively define as all activities from $A$ that are a parent activity (or great parent etc) of an activity in  $A_{HMM}$. We mark each activity in $A'_{HMM}$ as 'completed'. Then run the Simian process model on this modified set of activities to determine a prediction of how long it will still take to complete the entire process and thus determine the time to market. As before, we can of course run an entire ensemble of runs to generate a distribution of the likely date of when the product of Firm B will hit the market.

We note in passing that the Baum-Welch Algorithm for HMMs, which is a special case of the expectation maximization algorithm, is sometimes used to update the transition and emission matrices based on input data; however, since we use our Simian-based process model to produce as much training data as we please, we use the simple frequency-based approach to setting transition and emission probabilitites as described the previous paragraphs.

\subsection{Rationale for Studying the Structure of Process Models} 
Depending on the exact nature of the dependency graph as well as the activity to resource mapping, it is clear that very few observations could uniquely determine the hidden state. For instance, if an activity $a_j$ requires resource $r_i$, but no other activity requires resource $r_i$, then it is clear that an observation of resource $r_i$ uniquely determines that $a_j$ must be part of the hidden state label. On the other extreme end, if every activity requires almost every resource, a resource observation will give almost no information as to where we stand in the process model.

More generally, the structure or topology of both the dependency graph (among activities) as well as the resource to activity mapping of the process model, impacts how many observations we need to predict the correct set of activities with our HMM approach with some accuracy. Our numerical simulation study aims to discover basic insights about the relationship of this detection accuracy with the input process model based on random graph structures and mappings controlled with very few parameters, e.g., edge probabilities. While such random generation of process models does not capture any particular process for a specific product or industry in great detail and accuracy, we instead aim to learn more fundamental relationships of the complexity of a process model (as expressed by dependency graph and resource mapping) with observation sequence length and accuracy of prediction.

\section{Simulation Results}
\label{sec:simulation}

\subsection{Experimental Setup}

Our analysis aims to discover basic properties of this combination of PDES-based industrial modeling and HMM-based learning, rather than delve in detail into a specific example, which in most cases are subject to NDA rules. A previous study \cite{skau2024generating} uses a similar approach to ours on a specific example (coupled with intense mathematical analysis). We are more interested in the basic capabilities of our approach. We thus generate a large set of different process models, varying the activity sets, the resource sets, the corresponding dependency graphs among activities and the activity to resource maps. 

We set the following parameters: the total number of activities $n$, the total number of resources $m$, the density of the activity-graph $p$, and the density of the activity to resources mapping graph $q$. By definition, the activities' dependency graph is a directed acyclic graph (DAG); while the activities to resources mapping-graph is a bipartite graph.

For the experiments, we construct a probabilistic ensemble of activity-graphs and mapping-graphs with the specified densities $p$ and $q$. Similar to the Erd\"os-R\'enyi ensemble \cite{newman_evolution_2011}, the probability of an edge in an activities' dependency DAG is set to $p$, and probability of an edge in a mapping bipartite-graph is set to $q$. 

\begin{lstlisting}[
    language=Python,
    caption={Python function used for generating the adjacency matrix $A$ for a random activities' dependency DAG with edge density $p$, and number of activity nodes $n$.},
    label=Listing:DAG
]
def generateRandomDAG(n, p):
    A = []
    for i in range(0, n):
        A.append([0] * (i+1))
        for j in range(0, i):
            v = random.uniform(0.0, 1.0)
            if (v <= p):
                A[i][j] = 1

        # All nodes (i.e., except the root) 
        # have atleast one parent (predecessor) 
        # if there are no other parents (orphan)
        if (i > 0):
            totDeg = 0
            for j in range(0, i):
                totDeg += A[i][j]

            parent = random.randrange(0, i)
            if (totDeg == 0):
                A[i][parent] = 1

    return A
\end{lstlisting}

In order to generate a random activities' dependency DAG efficiently, observe that a DAG's nodes can be ordered in what is called a {\em topological sorting} partial-order. By this, we mean that predecessor nodes in the DAG always come before the successor nodes in the topological (partial) order. Given such an order of the nodes in a DAG, clearly the adjacency matrix will be lower-triangular, with a $1$ appearing in any off-diagonal position with probability given by the desired value $p$. Now, in addition to being a DAG, the graph needs to also be connected in our use case -- this is ensured by checking that a row always has at least one non-zero off-diagonal element; and in the exceptional case, if not, then one such element is assigned at random. Since throughout this procedure the adjacency matrix remains lower-diagonal, we ensure that the result is a random DAG with desired edge density. A Python function used to produce such a random ensemble is given in Listing \ref{Listing:DAG}.

For generating a random activities to resources bipartite-map with edge-density $q$, the edges are randomly placed with probability $q$ in a bipartite graph with $n$ activity nodes on one side and $m$ resource nodes on the other. A Python function used to generate the bipartite-map ensemble is shown in Listing \ref{Listing:BG}.

\begin{lstlisting}[
    language=Python,
    caption={Python function used for generating a random activities to resources bipartite mapping graph with edge density $q$, number of activity nodes $n$, and number of resource nodes $m$.},
    label=Listing:BG
]
def generateRandomBipartite(n, m, q):
    B = []
    for i in range(0, n):
        B.append([0] * m)
        for j in range(0, m):
            v = random.uniform(0.0, 1.0)
            if (v <= q):
                B[i][j] = 1

    return B
\end{lstlisting}

For generating an ensemble of graphs 
for our experiments, we varied $p,q \in \{0.2, 0.4, 0.6, 0.8\}$ and $m \in \{5, 10, 15\}$.
In all these cases, we held the total number of activities, $n$ a constant at $20$.
For each of the $m$ resources, the number required by an activity that depended on that resource was set uniformly at random from an integer set, $R_i \in \{1, 2, \ldots, 20\}$.

The time required to complete an activity was set using the {\em beta-distribution} with mean value selected uniformly at random from the range $t_j^{exp} \in [1, 200]$. Similarly, the extent of the beta-distribution was always in the closed range $[1, 200]$, i.e., $t^{\min}_j = 1, t^{\max}_j = 200$.

 \subsection{Base case results}
For our base case results, we ran Simian PDES sub-simulations for each sample from the above described ensemble $(\mathcal{E}_{DAG}, \mathcal{E}_{BG})$ to estimate the state-transition probabilities for the HMM. Next, the {\em Forward} and {\em Viterbi} algorithms were run on the inferred state-transition graph to estimate matching-success rates on hidden activities from observed resource-usages. Up to a maximum of $1000$ Simian sub-simulation runs were performed for each sample from the ensemble. In total $100$ samples from each ensemble was used for running these Simian sub-simulations. If for any of these $100$ samples after running $1000$ PDES sub-simulations we could not reliably collect all the state-transition probabilities, then we collected additional samples in order to make up for the bad samples.

In Figs. \ref{RI_1}, \ref{RI_2}, \ref{RI_3} we plot the matching-success rates versus the observation-sequence lengths needed to achieve those success rates when sweeping $(p, q)$, $(m, p)$, and $(m, q)$ while holding constant $m \in \{5, 10, 15\}$, $q \in \{0.2, 0.4, 0.6, 0.8\}$, and $p \in \{0.2, 0.4, 0.6, 0.8\}$ respectively. For each of these plots, we also show a transparent, colored $95\%$ confidence-interval band for the average matching-success rates.

For calculating the confidence bands, we first calculate the sample mean of matching-success rate $\mathbb{E}(\sigma)$ and its sample variance $\mathbb{V}(\sigma)$. Next we calculate a measure of the uncertainty in mean using the {\em standard error of mean} (SEM), $\lambda_{sem} := \frac{\sqrt{\mathbb{V}(\sigma)}}{S}$, where $S$ denotes the number of samples collected during the simulations. For a confidence-interval of $x$ percentage, the z-score is given by $z := \sqrt{2}\cdot\erf^{-1}{(\frac{x}{100})}$. Then the confidence band on the mean is given by the range: $[\mathbb{E}(\sigma) \pm z\cdot \lambda_{sem}]$. In our plots, we used $x := 95$.

The most striking insight from these initial figures is that we achieve success ratios of up to more than $.95$ and after 20 days of observations ($x$-axis), we are typically already very close to the highest achieved success rate. Our simulated processes last on average 150 days; thus observations for a duration of only 13 percent of the overall process duration are sufficient to capture most of the time-to-market knowledge that the HMM will produce even after many additional observations.

In Figs. \ref{CL_1}, \ref{CL_2}, \ref{CL_3} we plot the observation-sequence lengths required, $\ell_r$ to achieve $\{\geq 50\%, \geq 70\%, \geq 90\%, \geq 95\%\}$ of the maximum achievable success rates versus $m$, $q$, and $p$ while sweeping $(p, q)$, $(m, p)$, and $(m, q)$ respectively. In Figs. \ref{CV_1}, \ref{CV_2}, \ref{CV_3} we plot the actual achieved success-rates for $\{\geq 50\%, \geq 70\%, \geq 90\%, \geq 95\%\}$ of the maximum achievable success rates versus $m$, $q$, and $p$ while sweeping $(p, q)$, $(m, p)$, and $(m, q)$ respectively.

\subsection{Impact of Number of Resources}
Since the number of activities is being held constant in our study, varying the number of resources alone effectively constitutes varying the average number of resources needed per activity. As this average value gets larger, there are more opportunities for the HMM to capture these dependencies in an implicit manner, and hence the maximum asymptotically achieved matching-success ratios (as the observation-length is increased), saturates at a higher value. This is seen in Figs. \ref{RI_1}, \ref{CL_1}, and \ref{CV_1}. At the same time, the saturation in matching-success ratios sets in much slower for smaller values of $m$ when $n$ is held constant. The trends in required observation-sequence lengths for any given cut-off percentage of asymptotic matching-success rates is also similar -- larger values of $m$ require shorter observation-sequences as seen in Fig. \ref{CL_1}. Achieved success-ratios for various cutoff lengths shows a similar trend in Fig. \ref{CV_1}.

\subsection{Impact of Resource-Activity Mapping Density}
As the resource activity mapping graph gets sparser (or equivalently, $q$ gets smaller), fewer resource usage observations can help pinpoint the implicit activities. Therefore, the maximum asymptotically achieved matching-success ratios (as the observation-length is increased), saturates at a higher value as $q$ gets smaller. This is seen in Figs. \ref{RI_2}, \ref{CL_2}, and \ref{CV_2}. The saturation in matching-success ratios sets in much slower for larger values of $q$. The trends in required observation-sequence lengths for any given cut-off percentage of asymptotic matching-success rates gets interesting as $p$ gets larger -- as seen in Fig. \ref{CL_2}, there seems to be an optimal $q$ value (activity-resource map density) somewhere in the middle of the range $[0.0, 1.0]$, when the required observation-sequence lengths are the lowest. Achieved success-ratios for various cutoff lengths shows a similar trend in Fig. \ref{CV_2}.

\subsection{Impact of Activity Dependency Graph Density}
As the activity dependency graph gets denser (or equivalently, $p$ gets larger), there are more dependencies for each activity on average. This means that there are more correlations between the activities, and hence lesser number of observations would suffice for the HMM to identify hidden activities from resource usage observations. Therefore, the maximum asymptotically achieved matching-success ratios (as the observation-length is increased), saturates at a higher value as $p$ gets larger. This is seen in Figs. \ref{RI_3}, \ref{CL_3}, and \ref{CV_3}. The saturation in matching-success ratios sets in slower (curve is steeper) for smaller values of $p$. The trends in required observation-sequence lengths for any given cut-off percentage of asymptotic matching-success rates is also similar -- larger values of $p$ require shorter observation-sequences as seen in Fig. \ref{CL_3}. Achieved success-ratios for various cutoff lengths shows a similar trend in Fig. \ref{CV_3}.

\subsection{Impact of Shorter Observation Sequences}
We notice in Figs. \ref{CL_1}, \ref{CL_2}, and \ref{CL_3} that in most cases, to achieve $\geq 90\%$ of the asymptotic matching-success rate, it is sufficient to perform less that $50$ observations, as opposed to the default $100$ observations even in the worst case of small values of $m$. When the ratio $(\frac{m}{n})$ is increased, the required observation-sequence length decreases even more, to less than $20$ in most cases. This can be a significant savings in terms of data collection requirements in order to achieve a significant portion of the maximum possible asymptotic matching-success rates.

\subsection{Simulation Run Times}
%
%
%
For collecting $100$ good samples for $m=5$, and sweeping over a total of $16$ different settings for $p,q \in \{0.2, 0.4, 0.6, 0.8\}$, it took us $\approx$ $2$ hours and $40$ minutes on a $2023$, $14"$ Apple M3 Mac book. This amounted to approximately $6$ seconds per sample.
At $m=10$, a total of $100$ good samples took us $4$ hours, which is on average $9$ seconds per sample.
Finally, at $m=15$, all $100$ good samples took us $8$ hours, which is on average $18$ seconds per sample. From this, we observe that it took us more time on average to simulate a sample as the ratio $(\frac{m}{n})$ increased.

\section{Conclusion}
\label{sec:conclusion}
We have shown how to leverage process knowledge to combine the techniques of  Parallel Discrete Event Simulation (PDES)-based process model and Hidden Markov Models (HMMs). The PDES-model is thus used as a generative model to train the HMM. This is particularly useful in a data-sparse situation. Our numerical analysis across a broad range of activity graph densitities and resource-activity map densities has shown that in most cases, the HMM can identify the state of an observed process after observing for about 20 percent of the entire time duration. 
Future work includes further parameter studies, in particular with respect to larger process models. We also conjecture that the HMM, and more precisely the probability associated with the result of the Viterbi algorithm could be used to train a classifier to predict whether the HMM prediction of activities is actually correct.

\section*{Acknowledgment}
This work was funded by the U.S. Department of Energy National Nuclear Security Administration's Office of Defense Nuclear Nonproliferation Research and Development. This article is Los Alamos National Laboratory Publication LA-UR:24-31512.

\bibliographystyle{IEEEtran}
\bibliography{sources}
\vspace{12pt}

\end{document}